%% file: main.tex
\documentclass[twoside,11pt]{article}
\usepackage[preprint,abbrvbib]{jmlr2e}
\usepackage[T1]{fontenc}
\usepackage[nomath]{lmodern}
\usepackage[utf8]{inputenc}
\usepackage{amsmath,mathtools,mathrsfs}
\usepackage{microtype,booktabs,array,lastpage}
\hypersetup{hidelinks,pdftitle={Finite Spectral Depth Controls Gaussian Entrance Rank and Avoidance of the First Gradient-Flow Saddle},pdfauthor={Yanlin Long},pdfsubject={Finite-depth Newton--Schulz dynamics}}
\newtheorem{assumption}[theorem]{Assumption}
\DeclareMathOperator{\diag}{diag}
\DeclareMathOperator{\rank}{rank}
\DeclareMathOperator{\tr}{tr}
\DeclareMathOperator{\Ran}{Ran}
\DeclareMathOperator{\sign}{sign}
\newcommand{\R}{\mathbb R}
\newcommand{\N}{\mathbb N}
\newcommand{\eps}{\varepsilon}
\newcommand{\Lop}{\mathcal L}
\newcommand{\op}{\mathrm{op}}
\newcommand{\Sym}{\mathbb S}
\newcommand{\ip}[2]{\langle #1,#2\rangle}
\newcommand{\norm}[1]{\lVert#1\rVert}
\newcommand{\dd}{\,\mathrm d}
\newcommand{\kloss}{K_{\rm loss}}
\newcommand{\kpath}{K_{\rm bypass}}
\newcommand{\Ogroup}{\mathrm O}
\newcommand{\gm}{\mathcal G}
\newcommand{\TF}{\mathsf T}
\newcommand{\CE}{\mathsf C}
\allowdisplaybreaks[1]
\ShortHeadings{Finite-Depth Newton--Schulz Dynamics}{Long}
\firstpageno{1}
\jmlrheading{}{2026}{1--\pageref{LastPage}}{}{}{}{Yanlin Long}
\begin{document}
\input{sections/introduction}
\input{sections/model}
\input{sections/entrance}
\input{sections/continuation}
\input{sections/inertia}
\input{sections/example}
\input{sections/scope}
\label{body:lastpage}
\clearpage
\appendix
\input{appendices/foundations}

\input{appendices/entrance}

\input{appendices/comparison}

\input{appendices/depth}

\clearpage
\bibliography{references}
\end{document}

%% file: sections/introduction.tex
\title{Finite Spectral Depth Controls Gaussian Entrance Rank\\
 and Avoidance of the First Gradient-Flow Saddle}
\author{\name Yanlin Long \email ynlinlong@gmail.com\\
\addr Independent Researcher\\Guizhou, China}
\editor{}
\maketitle
\begin{abstract}
Matrix factorization provides a simple model for studying the training dynamics of neural networks. Recent studies of gradient flow under small initialization have analyzed how trajectories escape from a neighborhood of the origin and approach their first nonzero stationary point. In this paper, we study continuous-time dynamics that applies a finite number of Newton–Schulz iterations to the normalized matrix gradient (Muon). We show that the number of Newton–Schulz steps can change the rank selected during the early stages of training and allow the resulting trajectory to bypass the first nonzero stationary point approached by gradient flow. Under an additional residual-inertia condition, we prove that bypass occurs for all sufficiently large finite Newton–Schulz depths and provide an explicit sufficient threshold. We also construct an example in which the bypass threshold is four.
\end{abstract}
\section{Introduction}\label{sec:intro}

Muon and its variants have emerged as practical optimizers for large-scale language-model training \citep{liu2025scalable,kimi2025k2}, using Newton–Schulz iterations to approximately orthogonalize matrix-valued updates \citep{jordan2024muon}. Their design is closely related to steepest descent under the spectral norm, where the polar factor provides an idealized update direction \citep{bernstein2024old}. In practice, however, orthogonalization is performed by a fixed, small number of polynomial iterations. This distinction is particularly relevant near rank deficiency: exact polar orthogonalization maps every nonzero singular value to one, whereas finite Newton–Schulz iterations have a bounded amplification factor near zero at each fixed depth. Analyses of idealized spectral updates have identified learning dynamics distinct from gradient descent in concrete matrix models \citep{ma2026preconditioning,rhee2026balanced,li2026facts}, while finite Newton–Schulz updates have been studied through convergence guarantees and their smoothing effect \citep{kim2026convergence,li2026smoothing}. These developments motivate a direct study of how the finite number of iterations shapes the optimization trajectory.

Polynomial iterations for matrix orthogonalization have a long history in numerical linear algebra \citep{kovarik1970methods,bjorck1971orthogonal,higham1986polar,higham1990fast}. More recently, optimization analyses that explicitly retain finitely many Newton–Schulz iterations have investigated convergence to stationary points, the dependence on iteration depth, and the effects of stabilization and momentum \citep{kim2026convergence,li2026smoothing,peng2026practical,li2026lastiterate}. These results provide a useful foundation for studying the dynamical consequences of a fixed-depth spectral transformation.

Gradient-based training of factorized models from small initialization can exhibit incremental learning, with different components becoming appreciable at successive stages \citep{saxe2014exact,jacot2021saddle,li2021greedy,berthier2023incremental,jin2023incremental}. In matrix factorization, this behavior can take the form of dominant directions being learned first, followed by weaker directions that increase the effective rank of the predictor. The trajectory may consequently pass near intermediate low-rank stationary points. Related analyses of homogeneous networks describe directional convergence near the origin and saddle points, as well as the subsequent trajectory toward the first saddle after escape \citep{kumar2024early,kumar2025beyond}. A complementary approach studies training dynamics in high-dimensional limits, extracting effective structure from the evolving parameters or deriving dynamical mean-field descriptions of gradient flow \citep{martin2024impact,martin2026quadratic}. Related work on feature learning and the distinction from lazy training provides another perspective on how different directions become accessible during optimization \citep{chizat2019lazy,dandi2024giant}.

Since Newton–Schulz iterations amplify small singular values of the normalized gradient, a natural question is whether a finite increase in the number of iterations can make weaker directions emerge early enough to change which stationary points the trajectory approaches. Answering this question requires tracking both singular-value amplification and the coupled evolution of different directions, as the gradient and its singular subspaces evolve with the factorized parameters.

We consider a two-layer neural network with quadratic activation and all second-layer weights fixed to one. This model admits an equivalent representation as a positive-semidefinite matrix factorization. Such quadratic networks provide a tractable setting for analyzing nonlinear training dynamics, and have also been studied through gradient flow and spectral gradient methods \citep{martin2024impact,martin2026quadratic,braun2026spectral}. We study continuous-time optimization dynamics in which \(K\) cubic Newton–Schulz iterations are applied to the normalized factor gradient, and refer to \(K\) as the spectral depth.

For the analysis, we use a two-scale Gaussian initialization: one hidden row is initialized at scale \(\varepsilon\), while the remaining rows have scale \(\varepsilon^b\), where \(b>1\). The gradient normalization includes an additive stabilization proportional to \(\varepsilon\). This choice of initialization scales allows us to identify the leading directional dynamics first and then derive the coupled equations governing the weaker components in the remaining independent directions. We compare trajectories on the same task and with the same underlying Gaussian seed as \(\varepsilon\) tends to zero, while keeping each spectral depth fixed. Removing the row-dependent initialization scales would instead lead to an isotropic Gaussian initialization, for which the competing directions may exhibit different and potentially richer dynamics. The analysis of this regime remains open.

Another question concerns the extent to which small-initialization asymptotics can determine the subsequent training trajectory. Under suitable structural conditions, gradient flow admits a precise description in which the initialization scale can be used to reverse-engineer the full training dynamics, leading also to a characterization of time-dependent implicit regularization \citep{berthier2023incremental}. For finite-depth Newton–Schulz dynamics, the nonlinear matrix iteration prevents this reconstruction argument from carrying over directly. Although the dynamics can be analyzed in a small neighborhood of the origin, the resulting local description does not by itself determine the trajectory after leaving that neighborhood. This creates two distinct dynamical scales: the entrance dynamics near the origin and the subsequent evolution away from it. Establishing how the entrance dynamics are transmitted to the later trajectory is one of the main results of this paper.

Following the framework of \citet{li2021greedy}, which characterizes the first nonzero stationary point selected by gradient flow under suitable assumptions, we study the evolution from small initialization through the first stationary point encountered after escape. This allows us to compare the stationary points approached by gradient descent and by finite-depth Newton–Schulz dynamics. Under an additional structural assumption, we establish a sufficient large-depth guarantee: for all sufficiently large spectral depths, the selected Newton–Schulz continuation approaches a first stationary point whose loss is strictly smaller than that of the first stationary point selected by gradient descent. For smaller depths, our results do not provide a general criterion determining whether the trajectory visits or bypasses the gradient-descent stationary point. Moreover, without the additional assumption, the first stationary points reached by finite-depth Newton–Schulz dynamics and gradient descent may have the same loss. Such equal-loss behavior may reveal further dynamical structures, which we leave for future work.

The rest of this paper is organized as follows. Section 2 introduces the model and pathwise comparison criteria. Sections 3–5 establish the Gaussian entrance law, continuation criteria, and eventual depth guarantee. Section 6 presents an exact empirical task, and Section 7 discusses the scope of the results and open questions. Detailed proofs are given in the appendices.

Notations. We write \(\mathbb S^r\) for the real symmetric \(r\times r\) matrices and \(\mathbb S^r_+\) for the positive-semidefinite cone. The matrix inner product is \(\langle A,B\rangle=\operatorname{tr}(A^\top B)\). Subscripts \(\mathrm F\) and \(\mathrm{op}\) denote the Frobenius and operator norms, respectively; vector norms are Euclidean. For a symmetric matrix \(A\), \(n_+(A)\) counts its strictly positive eigenvalues with multiplicity. We write \(O(m)=\{Q\in\mathbb R^{m\times m}:Q^\top Q=I_m\}\).

%% file: sections/model.tex
\section{Main Results}\label{sec:setting}
In this section, we introduce the empirical quadratic model and finite-depth Newton–Schulz dynamics, and formulate the pathwise comparison with gradient flow from small initialization.

\subsection{Empirical quadratic learning}
Fix $m\ge2$ and $r\ge3$, and data $\{(x_l,y_l)\}_{l=1}^n$ with $x_l\in\R^r$ spanning $\R^r$ and $y_l\in\R$. Put $q=\min\{m,r\}$. For $W\in\R^{m\times r}$, set
\begin{equation}\label{eq:model}
 M=W^\top W,\qquad
 \Phi(M)=\frac1{2n}\sum_{l=1}^n(y_l-x_l^\top Mx_l)^2,\qquad
 f(W)=\Phi(W^\top W).
\end{equation}
This is the predictor of the quadratic network $x\mapsto\|Wx\|_2^2$ with fixed unit readouts. Define the task matrix, measurement operator, and residual by
\begin{equation}\label{eq:empirical}
 C=\frac1n\sum_l y_lx_lx_l^\top,\qquad
 \Lop(H)=\frac1n\sum_l(x_l^\top Hx_l)x_lx_l^\top,\qquad
 R(M)=C-\Lop(M).
\end{equation}
The gradient is
\[
 G(W):=-\nabla f(W)=2WR(W^\top W).
\]

\subsection{Finite spectral depth and two-scale initialization}
We use the cubic Newton--Schulz polynomial iteration \citep{bjorck1971orthogonal,higham1986polar}
\begin{equation}\label{eq:cubic-definition}
 P_0(X)=X,\qquad
 P_{k+1}(X)=\tfrac32P_k(X)-\tfrac12P_k(X)P_k(X)^\top P_k(X).
\end{equation}
On the unit Frobenius ball, $P_k$ preserves singular vectors and applies
the scalar polynomial $p_k$ to each singular value, where
\begin{equation}\label{eq:scalar}
 p_0(s)=s,\qquad p_{k+1}(s)=\tfrac32p_k(s)-\tfrac12p_k(s)^3,
 \qquad a_K=p_K'(0)=(3/2)^K.
\end{equation}
Fix $b>1$, $c_\eta>0$, and finite $K\in\N_0=\{0,1,2,\ldots\}$. For a matrix $U\in\R^{m\times r}$ with independent standard normal entries, initialize
\begin{equation}\label{eq:init}
 W_K^{\eps,b}(0;U)=D_{\eps,b}U,\qquad
 D_{\eps,b}=\diag(\eps,\eps^b,\ldots,\eps^b),\qquad
 \eta_\eps=c_\eta\eps.
\end{equation}
The actual trajectory, in physical time $t$, solves
\begin{equation}\label{eq:flow}
 \frac{\dd W}{\dd t}=P_K\!\left(\frac{G(W)}{\norm{G(W)}_F+\eta_\eps}\right).
\end{equation}

The additive stabilization in the denominator follows the original
Muon implementation, which uses a fixed numerical constant of
$10^{-7}$ in gradient normalization \citep{jordan2024muon}.
The choice $\eta_\eps=c_\eta\eps$ introduces an additional
structural assumption: the stabilization term is taken to be
of the same order as the small-initialization scale.
Different scaling choices may lead to different balances
between the gradient norm and the stabilization term, and
hence different rescaled dynamics.
For simplicity, we adopt this linear scaling throughout the paper.

Throughout the paper, we fix a finite spectral depth $K$
before taking the limit $\eps\downarrow0$, with the same
Gaussian seed $U$.

Since $p_K$ is odd, there is a polynomial $q_K$ satisfying
\[
 p_K(s)=s q_K(s^2),\qquad q_K(0)=a_K.
\]
For $G\in\R^{m\times r}$ and $\eta\ge0$, define
\begin{equation}\label{eq:H}
 H_{K,\eta}(G)=
 \begin{cases}
 (\norm G_F+\eta)P_K\!\left(G/(\norm G_F+\eta)\right),
 &\norm G_F+\eta>0,\\
 0,&G=0,\ \eta=0.
 \end{cases}
\end{equation}
We use $\tau$ for the time variable in
$\dd W/\dd\tau=H_{K,\eta}(G(W))$.

For the actual predictor, write
\[
 M_K^{\eps,b}(t;U)
 =W_K^{\eps,b}(t;U)^\top W_K^{\eps,b}(t;U).
\]

\subsection{Continuation and the gradient-flow comparison}
Multiplying the physical vector field by its positive denominator gives
a convenient time parameter for continuation. The following bounds keep
that change of time and all later finite-interval comparisons on the
same ordered path.
\begin{proposition}[Continuation of finite-depth dynamics]\label{main:global}
\label{lem:compact-sublevels}\label{lem:descent}\label{lem:continuation}\label{lem:rank-preservation}
For a spanning empirical task, the following assertions hold in arbitrary
positive factor dimensions. Fix finite $K$ and $\eta\ge0$.
Every solution of
\[
 \frac{\dd W}{\dd\tau}=H_{K,\eta}(G(W))
\]
is uniquely defined for all $\tau\ge0$ and converges to one stationary
factor $W_s$. Every finite time preserves the initial rank.
Every PSD sublevel of $\Phi$, and every factor sublevel of $f$, is compact.
Along the solution,
\begin{equation}\label{eq:descent}
 -\frac{\dd f}{\dd\tau}\ge\norm{\nabla f}_F^2,\qquad
 \norm{\frac{\dd W}{\dd\tau}}_F\le a_K\norm{\nabla f}_F,\qquad
 -\frac{\dd f}{\dd\tau}\ge
 a_K^{-1}\norm{\nabla f}_F\norm{\frac{\dd W}{\dd\tau}}_F.
\end{equation}
For $\eta>0$, the physical-time equation with denominator
$\norm{G(W)}_F+\eta$ has the same ordered path, is defined for all
$t\ge0$, converges to $W_s$, and satisfies
\[
 \rank W(t)=\rank W(0)\qquad(0\le t<\infty).
\]
The change of time is
$\dd t/\dd\tau=\norm{G(W)}_F+\eta$.
More generally, a bounded $C^1$ curve satisfying \eqref{eq:descent}
for a real-analytic loss on a neighborhood of its closure converges
to a stationary point. These conclusions also apply to one-row and
projected empirical problems used below.
\end{proposition}

At depth zero, $H_{0,\eta}(G)=G$. The corresponding predictor evolves
by ordinary factor gradient flow. Its first small-initialization stage
selects a rank-one trajectory whose stationary endpoint will remain
fixed throughout the optimizer comparison.
We further assume that the two largest eigenvalues of $C$
are distinct.

\begin{assumption}[Task spectrum]\label{ass:task}
The task matrix $C$ is nonsingular. Its signed eigenvalues and an
orthonormal eigenbasis satisfy
\[
 Cv_l=\lambda_lv_l,\qquad
 \lambda_1>\lambda_2\ge\cdots\ge\lambda_r,\qquad\lambda_2>0.
\]
\end{assumption}

Spectral separation conditions are common in analyses of
incremental learning in matrix factorization
\citep{jiang2023algorithmic}.
Here, strict gap between the two largest
eigenvalues, allowing multiplicities among the remaining ones.
Removing the leading spectral gap may give rise to other
interesting dynamical structures, which we leave for future work.
\begin{proposition}[First gradient-flow stage]\label{main:gd}
\label{prop:glrl}\label{cor:gf-gaussian}
Under Assumption~\ref{ass:task}, there is a task-dependent nonconstant
rank-one predictor trajectory $\mathcal M^{\rm G}(s)$ and a nonzero
rank-one stationary predictor
\begin{equation}\label{eq:gf-endpoint}
 M_1=\lim_{s\to\infty}\mathcal M^{\rm G}(s),\qquad M_1R(M_1)=0.
\end{equation}
The trajectory is selected by the leading positive eigendirection of
$C$ in the small-initialization limit of factor gradient flow.
For each fixed $b>1$, the actual depth-zero paths satisfy
\begin{equation}\label{eq:gd-approach}
 \lim_{\eps\downarrow0}\inf_{t\ge0}
 \norm{M_0^{\eps,b}(t;U)-M_1}_F=0
 \quad\text{for Gaussian-almost every }U.
\end{equation}
\end{proposition}
The explicit time-shifted selection law for a general positive
semidefinite initial family is given in Appendix~\ref{app:comparison}.
The endpoint $M_1$ is determined by the empirical flow after its leading
direction has been selected. Set $R_1=R(M_1)$ and $\phi_1=\Phi(M_1)$.

\subsection{All-time distance and loss crossing}
A visit is measured over the whole physical-time path. The second
observable asks whether training reaches a level below the comparison
loss, providing a quantitative loss improvement alongside separation.
\begin{definition}[First nonzero stationary-point bypass]
\label{def:bypass}
For the predictor $M_1$ of Proposition~\ref{main:gd} and a positive loss gap $\Delta$, define
\begin{align}
d_1(K,b,\eps,U)
&=\inf_{t\ge0}
\frac{\norm{M_K^{\eps,b}(t;U)-M_1}_F}
{\norm{M_1}_F},
\label{eq:distance}\\
T_\Delta(K,b,\eps,U)
&=\inf\left\{t\ge0:
\Phi(M_K^{\eps,b}(t;U))
\le\phi_1-\Delta\right\}.
\label{eq:crossing}
\end{align}
An empty hitting set has time $+\infty$.

A finite depth $K$ gives almost-sure small-initialization
bypass of $M_1$ if, for almost every fixed Gaussian seed $U$,
\[
 \liminf_{\eps\downarrow0}d_1(K,b,\eps,U)>0,
 \qquad
 \limsup_{\eps\downarrow0}\inf_{t\ge0}
 \Phi(M_K^{\eps,b}(t;U))<\phi_1.
\]
The second condition is equivalent to eventual finite crossing of a common loss level strictly below $\phi_1$; this level and the initialization cutoff may depend on the fixed seed.

Define $\kpath(b)$ as the minimum such depth, with
$\min\varnothing=+\infty$.
\end{definition}

The infimum over time is taken separately at each positive $\eps$.
The ensuing small-initialization limit therefore tests the entire path,
including intervals whose physical or continuation times diverge with
initialization. Later results give a deterministic distance bound and a
deterministic loss gap, with only the admissible initialization cutoff
depending on the fixed seed.

%% file: sections/entrance.tex
\section{Gaussian Entrance at Finite Spectral Depth}\label{sec:entrance}
An independent weak component can be suppressed during one stage and
amplified during a later stage. The entrance rank records how many such
components reach a fixed rescaled size before physical time becomes
independent of initialization. This section first gives the finite
recursion computing that rank, then identifies its limit in the actual
Gaussian dynamics.

\subsection{Rescaled shapes and normal growth rates}
For $\sigma=t/\eps$ and $v=\log t$, define
\[
 Y^\eps(\sigma)=\frac{W_K^{\eps,b}(\eps\sigma;U)}\eps,
 \qquad Q^\eps(t)=\frac{W_K^{\eps,b}(t;U)}t.
\]
The exact relation is $Q^\eps(\eps\sigma)=Y^\eps(\sigma)/\sigma$.
The limiting vector field for the second rescaling is
\begin{equation}\label{eq:limiting-field}
 F_K(Q)=P_K(QC/\norm{QC}_F)-Q,\qquad Q\ne0.
\end{equation}
Nonsingularity of $C$ makes this expression well defined.
Lemma~\ref{lem:rescaling} gives the exact rescaled empirical equation and
its local $C^1$ error of order $t^2+\eps/t$. The two contributions are the
empirical correction and denominator stabilization, respectively.

A matrix satisfying
$F_K(Q_{I,K})=0$ and
$\Ran(Q_{I,K}^\top)=\operatorname{span}\{v_l:l\in I\}$
is denoted by $Q_{I,K}$ whenever it exists, and
$s_{I,K}=\norm{Q_{I,K}C}_F$. We identify factors in
$\{OQ_{I,K}:O\in\Ogroup(m)\}$ by their common Gram matrix.
For the leading $d$ task directions, write
$Q_{d,K}=Q_{\{1,\ldots,d\},K}$ and
$s_{d,K}=s_{\{1,\ldots,d\},K}$.

For $K\ge1$, an existing positive support has unique coefficients
$q_l>0$ determined by
\begin{equation}\label{eq:zero-system}
\begin{aligned}
 Q_{I,K}^\top Q_{I,K}&=\sum_{l\in I}q_l^2v_lv_l^\top,\\
 q_l&=p_K(\lambda_lq_l/s_{I,K}),\quad l\in I,
 &s_{I,K}^2&=\sum_{l\in I}\lambda_l^2q_l^2.
\end{aligned}
\end{equation}
The rank-one shape is $Q_{1,K}=e_1v_1^\top$ with
$s_{1,K}=\lambda_1$, including $K=0$. A new independent direction $v_l$
has growth rate
\begin{equation}\label{eq:normal-rate-definition}
 g_l^{(d)}=\frac{a_K\lambda_l}{s_{d,K}}-1,\qquad l>d.
\end{equation}
A positive next rate constructs the next leading support by
Lemma~\ref{lem:add-zero}. Negative task eigenvalues give rates below
$-1$. The local derivative in the already occupied rank manifold is
contracting after factors with equal Gram matrices are identified.

For a symmetric matrix $A$, let $n_+(A)$ count its strictly positive
eigenvalues with multiplicity. The positive-direction capacity is
\begin{equation}\label{eq:capacity}
 p=\min\{m,n_+(C)\},\qquad 2\le p\le q=\min\{m,r\}.
\end{equation}
The entrance may have rank below $p$. The actual positive Gaussian
initialization has rank $q$ almost surely; its rank is preserved at every
finite time. These two ranks enter different parts of the analysis.

\begin{assumption}[Separated capacity prefix]\label{ass:prefix}
The ordered eigenvalues in Assumption~\ref{ass:task} satisfy
\begin{equation}\label{eq:capacity-gaps}
 \lambda_k>\lambda_{k+1}
 \qquad(1\le k\le\min\{p,r-1\}).
\end{equation}
\end{assumption}

The gaps in this assumption distinguish the leading exterior column at
each retained order. Eigenvalues beyond the tested prefix may repeat.
When $p=n_+(C)<r$, the last tested boundary is already strict because
$C$ is nonsingular and that boundary separates positive and negative
eigenvalues.

\subsection{Executable clocks and exceptional exponents}\label{def:clocks}
The clock $x_d$ measures a fraction of the logarithmic interval from
$t=O(\eps)$ to a fixed physical scale. Its increment is the remaining
suppression exponent divided by the new normal growth rate.
\begin{definition}[Candidate entrance-rank computation]
\label{def:entrance-rank-computation}
Fix $K$ and $b>1$, with capacity $p$ as in \eqref{eq:capacity}.

Start at rank one with $x_1=0$ and
$s_{1,K}=\lambda_1$.
Define the first candidate by
\begin{equation}\label{eq:first-clock}
x_2=
\begin{cases}
\displaystyle
\frac{b-1}{a_K\lambda_2/\lambda_1-1},
& a_K\lambda_2/\lambda_1>1,\\[6pt]
+\infty,
& a_K\lambda_2/\lambda_1\le1.
\end{cases}
\end{equation}
At each subsequently reached rank $2\le d<p$, compute the
normalization from \eqref{eq:zero-system} and the next
growth rate from \eqref{eq:normal-rate-definition}.
Define
\begin{equation}\label{eq:clock}
x_{d+1}
=
\begin{cases}
\displaystyle
x_d+
\frac{(b-1+x_d)(1-\lambda_{d+1}/\lambda_d)}
     {g_{d+1}^{(d)}},
& g_{d+1}^{(d)}>0,\\[8pt]
+\infty,
& g_{d+1}^{(d)}\le0.
\end{cases}
\end{equation}
Proceed to rank $d+1$ whenever $x_{d+1}<1$.
Stop at capacity $p$ or at the first candidate $x_{d+1}\ge1$.
Denote the final rank by $d_K(b)$.
\end{definition}
To separate the exponent from this procedure, first build the longest
leading support allowed by positive addition rates, without applying
the test $x_d<1$. Start with $Q_{1,K}$ and, while $d<p$, add the
next positive task eigenvector exactly when $g_{d+1}^{(d)}>0$.
The construction stops at capacity $p$. On this algebraic sequence set
$c_{1,K}=0$ and
\begin{equation}\label{eq:tau-clock}
 \begin{aligned}
 c_{2,K}&=\frac1{a_K\lambda_2/\lambda_1-1},\\
 c_{d+1,K}&=c_{d,K}+
 \frac{(1+c_{d,K})(1-\lambda_{d+1}/\lambda_d)}{g_{d+1}^{(d)}}
 \qquad(2\le d<p).
 \end{aligned}
\end{equation}
A displayed quotient is evaluated only when its addition rate is
positive. At the first nonpositive rate, that next coefficient and all
later coefficients are set to $+\infty$. Put
\begin{equation}\label{eq:exceptional-exponents}
 \mathcal B_{C,K}=\{1+1/c_{d,K}:2\le d\le p,\ 0<c_{d,K}<\infty\},
 \qquad \mathcal B_C=\bigcup_{K\in\N_0}\mathcal B_{C,K}.
\end{equation}
The finite candidate equations determine $\mathcal B_{C,K}$ from the
task spectrum, depth, and capacity. Proposition~\ref{prop:clock-domain}
proves existence of each evaluated support and the identity
$x_d=(b-1)c_{d,K}$. In particular $|\mathcal B_{C,K}|\le p-1$, and
$\mathcal B_C$ is countable. The resulting staircase is
\begin{equation}\label{eq:staircase}
 d_K(b)=\max\{d\le p:(b-1)c_{d,K}<1\}.
\end{equation}
Here a positive number times $+\infty$ is $+\infty$.
The first rank increase occurs exactly when $b<a_K\lambda_2/\lambda_1$.
\subsection{The actual first-radius limit}
For fixed $K,b$ and $\rho>0$, define
\begin{equation}\label{eq:first-radius}
 T_\rho^\eps=\inf\{t\ge0:\norm{M_K^{\eps,b}(t;U)}_F=\rho\},
 \qquad\inf\varnothing=+\infty.
\end{equation}
The first-radius entrance rank is the rank of the predictor limit at
this first hit, whenever the hit and limit exist. The theorem identifies
both and fixes the continuation selected at the boundary $t=0$.
\begin{theorem}[Gaussian first-radius entrance]\label{thm:passage}
Fix Assumptions~\ref{ass:task} and~\ref{ass:prefix}, $m\ge2$, $r\ge3$,
and $c_\eta>0$. For every fixed finite $K$ and every
$b>1$ with $b\notin\mathcal B_{C,K}$, put $d=d_K(b)$.
There is a local rank-$d$ solution $\overline W_{K,d}$ of
\[
 \dot W=P_K(G(W)/\norm{G(W)}_F),\qquad W(t)/t\longrightarrow Q_{d,K},
\]
unique in this fixed-rank boundary class up to a constant left orthogonal
transformation. For some fixed $t_*>0$ it satisfies
\begin{equation}\label{eq:entrance-expansion}
 \overline W_{K,d}(t)=tQ_{d,K}+O(t^3),\qquad
 \dot{\overline W}_{K,d}(t)=Q_{d,K}+O(t^2).
\end{equation}
There is a measurable Gaussian-full set $\gm_{K,b}$ such that for every
$U\in\gm_{K,b}$,
\begin{equation}\label{eq:full-limit}
 \lim_{\eps\downarrow0}\inf_{O\in\Ogroup(m)}
 \norm{OW_K^{\eps,b}(t;U)-\overline W_{K,d}(t)}_F=0
 \quad\text{locally uniformly on }(0,t_*).
\end{equation}
For each sufficiently small fixed $\rho>0$, let $t_\rho$ be the first
time when $\norm{\overline W_{K,d}(t)^\top\overline W_{K,d}(t)}_F=\rho$.
For every $U\in\gm_{K,b}$, the actual first hits exist for all
sufficiently small $\eps$, their times converge to $t_\rho$, and
\begin{equation}\label{eq:entrance-main}
 M_K^{\eps,b}(T_\rho^\eps;U)\longrightarrow
 \overline W_{K,d}(t_\rho)^\top\overline W_{K,d}(t_\rho).
\end{equation}
This predictor is independent of $U$ and has rank $d_K(b)$.
For every fixed $b\notin\mathcal B_C$, these assertions hold on one
common Gaussian-full event for all finite $K$.
\end{theorem}
The time interval and admissible radii are determined by the task and
selected shape. Constants may depend on $K$, and the small-initialization
cutoffs may additionally depend on the seed. The countable exceptional
set implies the stated result for Lebesgue-almost every $b>1$ at every
finite depth.

\begin{proposition}[Depth monotonicity of entrance rank]\label{prop:rank-monotone}
Under Assumptions~\ref{ass:task} and~\ref{ass:prefix}, the algebraically
computed rank $d_K(b)$ is nondecreasing in the integer depth $K$.
For the actual first-radius ranks, this comparison holds at the depths
covered by Theorem~\ref{thm:passage}.
\end{proposition}
The normalization of an existing support of rank at least two increases
with depth, but its ratio to $a_K$ decreases. The latter comparison
increases the unused growth rates and decreases the finite clock
coefficients. Appendix~\ref{app:entrance} proves these scalar properties
and the proposition.

\subsection{How Gaussian independent scales determine the clocks}
The initial-layer equations separate the leading row from the weak rows
on every fixed finite $\sigma$ interval. Eliminating the component of a
weak row parallel to the leading row gives a linear right-multiplication
equation. Conditional on the leading Gaussian row, its independent
coefficients have a nonsingular Gaussian density. The relevant minors
are nonzero simultaneously on one full-probability event. One then fixes
a seed in that event and chooses a sufficiently large finite
$\sigma_0(U)$; this time stays fixed while $\eps\downarrow0$.

In logarithmic initial-layer time
\begin{equation}\label{eq:mu-time}
 \mu=\log\sigma=\log(t/\eps),
\end{equation}
the leading independent rates are
$g_l^{(1)}=a_K\lambda_l/\lambda_1-1$, with an integrable coefficient
error. At $t=\eps\sigma_0(U)$, the leading independent $h$-volume has
logarithm $h(b-1)\log\eps+O_U(1)$. This is the quantitative information
passed from the fixed initial layer to the longer residence intervals.

For a later rank-$d$ stage, project the unused task columns away from
the occupied hidden span and express the result as $Z$. The exact
projected equation is homogeneous in $Z$ and retains both the empirical
correction and the denominator term $\eps/t$. Exterior products measure
independence: the consecutive volume ratio is the component of the next
column orthogonal to the preceding columns. Thus a large column already
contained in their span contributes no new independent direction.

The first waiting fraction already illustrates the balance. On the
rank-one stage, a weak independent amplitude begins at order
$\eps^{b-1}$ and grows at logarithmic rate $g_2^{(1)}$. Reaching a fixed
positive size therefore requires a logarithmic elapsed time
$-(b-1)\log\eps/g_2^{(1)}+O_U(1)$. Returning to physical time gives
\[
 \log t_2^\eps
 =\left(1-\frac{b-1}{g_2^{(1)}}\right)\log\eps+O_U(1).
\]
This crossing remains on a vanishing physical scale exactly when
$x_2<1$. If the rate is nonpositive, the first candidate is infinite;
if the finite clock reaches one, the crossing is at the boundary of the
scale on which the homogeneous approximation determines the answer.
The exceptional set records precisely these finite boundary values.

For two completed additions, consider an unused task label $l>2$.
At the second-rank entry, its remaining exponent is
\[
 (b-1)-g_l^{(1)}x_2
 =(b-1+x_2)\left(1-\frac{\lambda_l}{\lambda_2}\right).
\]
The equality follows from $(b-1)=g_2^{(1)}x_2$ and
$g_l^{(1)}+1=(\lambda_l/\lambda_2)(g_2^{(1)}+1)$.
Projection away from the newly occupied direction changes the
multiplicative coefficient by bounded positive factors, so it preserves
this exponent. At a later rank $d$, the same calculation with the
transferred exponents gives
$(b-1+x_d)(1-\lambda_l/\lambda_d)$. Dividing the exponent for
$l=d+1$ by its new rate gives the increment in \eqref{eq:clock}.
Thus each clock uses the growth rates of the occupied stage together
with the suppression inherited from earlier stages.

The residence interval can have logarithmic length proportional to
$\log(1/\eps)$. Integrating its volume equation therefore requires a
bound on the time integral of the coefficient error. The actual estimates
in Lemma~\ref{lem:integral} give this bound uniformly in residence length.
For the two explicit rescaling terms, their accumulated contribution
from $t=\eps\sigma_0$ to a fixed small $t_*$ is bounded by
\[
 \int_{\eps\sigma_0}^{t_*}
       \left(t^2+\frac{\eps}{t}\right)\frac{\dd t}{t}
 =\frac{t_*^2-\eps^2\sigma_0^2}{2}
       +\frac1{\sigma_0}-\frac{\eps}{t_*}
 \le\frac{t_*^2}{2}+\frac1{\sigma_0}.
\]
The stable tangent equation and the increasing normal component give
corresponding integrated bounds for the remaining coefficient terms.
The resulting control is multiplicative on each independent volume:
it preserves a positive lower bound relative to its incoming value,
even when that value is a high power of $\eps$.
These estimates identify each completed first crossing as
\[
 \log t_{d+1}^\eps=(1-x_{d+1})\log\eps+O_U(1).
\]
After that crossing, projection away from the new direction and relative
transport on a fixed connection segment preserve every remaining
independent exponent. This produces the numerator
$(b-1+x_d)(1-\lambda_{d+1}/\lambda_d)$ in \eqref{eq:clock}.

At the final stage, the unused component tends to zero on each compact
positive-time interval. A stable Gram equation with a zero-time boundary
then selects the rank-$d_K(b)$ empirical entrance. Its expansion in
\eqref{eq:entrance-expansion}, a speed bound before a fixed early time,
and transverse crossing at the selected section identify the actual
\emph{first} hit. The definitions and estimates for these steps appear
in Appendix~\ref{app:entrance}.

The selected entrance solves the empirical equation at positive time.
In its rescaled Gram equation, the stable tangent correction has order
$t^2$, so lifting back to a factor gives the $O(t^3)$ correction in
\eqref{eq:entrance-expansion}. The loss operator therefore enters the
selected continuation even though the finite clock recursion uses only
the task spectrum. This separates two computations: the spectrum and
initialization exponent determine the entrance rank, and the empirical
operator determines the continuation from its boundary shape.

The argument also distinguishes convergence at a prescribed positive
time from convergence at the first section. The uniform speed bound
excludes an earlier actual hit on an initial interval. On the remaining
compact interval before the selected hit, the limiting predictor stays
strictly below the chosen radius. Uniform convergence preserves that
margin. Transversality then places the actual first hit in every small
neighborhood of the selected hitting time. This gives the time and
predictor conclusions of Theorem~\ref{thm:passage} on the same seed event.

%% file: sections/continuation.tex
\section{Selected Continuation and All-Time Bypass}\label{sec:continuation}
\label{sec:main-continuation}
The entrance theorem supplies a deterministic predictor at a small
section. Continuing from that selected entrance gives a finite comparison
segment. A strict loss decrease on such a segment subsequently controls
all later actual times. At equal terminal loss, control instead comes
from the finite length of trajectories near an analytic stationary point.

\subsection{The selected empirical continuation}
Write $\varphi_K(\tau,Z)$ for the solution starting at $Z$ of
\begin{equation}\label{eq:continuation-flow}
 \frac{\dd W}{\dd\tau}=H_{K,0}(G(W)).
\end{equation}
For a depth and exponent covered by Theorem~\ref{thm:passage}, choose a
factor $Z$ at a sufficiently small positive time on
$\overline W_{K,d_K(b)}$ and define
\[
 M_{s,K}=\lim_{\tau\to\infty}
       \varphi_K(\tau,Z)^\top\varphi_K(\tau,Z),\qquad
 \ell_{K,d_K(b)}=\Phi(M_{s,K}).
\]
Proposition~\ref{main:global} gives existence. Uniqueness of continuation,
its positive time change, and left orthogonal equivariance make these
objects independent of the chosen positive entrance point and hidden
frame. The starting point is fixed by the selected entrance. The
notation $\ell_{K,d}$ also applies when that entrance has a specified
rank $d$, with its boundary Gram matrix fixed.

\begin{assumption}[A nonglobal comparison predictor]\label{ass:comparison}
The first gradient-flow endpoint satisfies
\begin{equation}\label{eq:bottleneck}
 \lambda_{\max}(R_1)>0.
\end{equation}
\end{assumption}

Under this condition every factor of $M_1$ has negative curvature in an
unused hidden direction, as proved in
Corollary~\ref{cor:comparison-saddle}. It also implies that the minimum
PSD loss is strictly below $\phi_1$. These properties concern the
comparison and loss-transfer results of this section and
Section~\ref{sec:inertia}.

\begin{theorem}[Strict selected loss implies actual bypass]\label{thm:separation}
Under the pointwise hypotheses of Theorem~\ref{thm:passage} and
Assumption~\ref{ass:comparison}, suppose
$\ell_{K,d_K(b)}<\phi_1$.
There are deterministic constants $d_*,\Delta_*>0$ and a Gaussian-full
event such that, for every fixed seed $U$ in that event,
\[
 d_1(K,b,\eps,U)\ge d_*,\qquad
 T_{\Delta_*}(K,b,\eps,U)<\infty
\]
for every sufficiently small $\eps>0$, with the cutoff allowed to
depend on $U$.
\end{theorem}
The entry-set formulation in Appendix~\ref{app:comparison} gives the
same selected terminal loss at every sufficiently small section. It also
retains a section-independent terminal-loss test and its minimum
successful depth.

\subsection{A finite comparison segment and the entire tail}
Before the first-radius hit, the predictor norm is small, giving a
positive distance from $M_1$. Starting at that section, choose a finite
segment of the selected continuation ending below $\phi_1$.
Every factor of $M_1$ is stationary. Uniqueness prevents this selected
segment from meeting such a factor, since its terminal loss is lower.
Compactness of the segment gives a positive minimum distance.
Finite-time continuous dependence transfers both this distance and the
terminal loss gap to the actual small-initialization path.

For all subsequent times, actual loss monotonicity preserves the gap.
At $M_1$, the empirical quadratic identity is
\[
 \Phi(M)-\phi_1
 =-\langle R_1,M-M_1\rangle
  +\tfrac12\langle M-M_1,\Lop(M-M_1)\rangle.
\]
Its nonnegative quadratic term converts the preserved loss decrease
into a permanent predictor-distance bound. In particular, whenever
$\Phi(M)\le\phi_1-\Delta$ for a fixed $\Delta>0$,
\[
 \Delta\le\langle R_1,M-M_1\rangle
 \le\norm{R_1}_F\norm{M-M_1}_F.
\]
After the crossing, the same inequality applies at every later actual
time. The small-initialization comparison is used only to reach this
fixed loss level; the algebraic loss identity then supplies the tail
bound without an initialization-dependent comparison interval.
Combining the early interval,
the fixed comparison segment, and this tail proves the all-time result.
The comparison time is chosen before taking the initialization limit.

\subsection{Equal loss, selected capture, and full width}
A selected endpoint can have loss $\phi_1$ while differing from $M_1$.
The following corollary distinguishes that case from selection of $M_1$
itself. It also records the actual full-width loss statement with its
own empirical hypotheses.

\begin{corollary}[Selected endpoints and actual paths]\label{thm:conditional-classification}
The following assertions have separate hypotheses.
\begin{enumerate}
\item Under the pointwise hypotheses of Theorem~\ref{thm:passage}, if
$M_{s,K}=M_1$, then for Gaussian-almost every fixed seed,
\[
 \lim_{\eps\downarrow0}d_1(K,b,\eps,U)=0.
\]
\item Under the pointwise hypotheses of Theorem~\ref{thm:passage} and
Assumption~\ref{ass:comparison}, suppose
$\ell_{K,d_K(b)}=\phi_1$ and $M_{s,K}\ne M_1$.
There is a deterministic $d_*>0$ and a Gaussian-full event on which,
for each seed, $d_1(K,b,\eps,U)\ge d_*$ for all sufficiently small
$\eps>0$. If also $m\ge r$, this depth gives bypass.
\item Under the empirical model, $m\ge r$, and
Assumption~\ref{ass:comparison}, let
$\Phi_*:=\min_{M\succeq0}\Phi(M)$. Then $\Phi_*<\phi_1$.
On one Gaussian-full event, for every $\eps>0$ and every finite $K$,
\[
 \lim_{t\to\infty}\Phi(M_K^{\eps,b}(t;U))=\Phi_*.
\]
Thus the strict loss-crossing part of Definition~\ref{def:bypass}
holds at every finite depth. The crossing times may depend on $\eps$.
\item Under the pointwise hypotheses of Theorem~\ref{thm:passage},
$m\ge r$, and Assumption~\ref{ass:comparison}, at every fixed depth
with $\ell_{K,d_K(b)}\le\phi_1$,
\[
 K\text{ gives almost-sure small-initialization bypass}
 \quad\Longleftrightarrow\quad M_{s,K}\ne M_1.
\]
\end{enumerate}
\end{corollary}
Part~(iii) requires the empirical model and full width at positive
initialization. Its probability-one event is the full-column-rank event
for $U$. Parts~(i), (ii), and (iv) use the entrance event at the fixed
admissible depth; the loss premise in part~(iv) is imposed at that depth.

For part~(ii), choose a small factor ball around the selected terminal
factor whose predictors stay separated from $M_1$. The analytic gradient
inequality and the angle bound in \eqref{eq:descent} control the length
of actual trajectories inside a small loss interval around $\phi_1$.
The portions above and below $\phi_1$ have separate integrable length
bounds. A finite selected segment brings the actual factor inside the
ball. It then either stays there or crosses the lower loss boundary,
after which the permanent loss-gap argument applies. All radii, loss
widths, and distance constants are fixed before choosing the actual seed
cutoff. Appendix~\ref{app:comparison} gives the quantitative margins.

The distinction between parts~(ii) and~(iii) is useful for the order of
limits. In part~(ii), the selected endpoint and its neighborhood are
fixed before $\eps\downarrow0$. In part~(iii), the positive-initialization
path is followed to its stationary limit; full finite-time rank forces
the terminal residual to be nonpositive by Lemma~\ref{lem:inertia}.
The latter argument gives a loss decrease at a finite time for every
positive initialization, but that time can diverge as initialization
vanishes. The corollary combines these two conclusions only in the
parts whose hypotheses include both entrance selection and full width.

%% file: sections/inertia.tex
\section{Residual Inertia and an Eventual Depth Guarantee}\label{sec:inertia}
A large entrance rank can rule out a high-loss stationary endpoint.
The relevant restriction uses the rank at finite time, which the flow
preserves, and the positive eigenvalues of the residual at the limit.
This distinction allows the selected predictor to lose rank as it
converges.

\subsection{Preserved rank and terminal residual inertia}
\begin{lemma}[Rank--inertia bound]\label{lem:inertia}
Let $B(\tau)$ be measurable and symmetric with $I\preceq B(\tau)\preceq aI$ for fixed $a\ge1$, and let a locally absolutely continuous forward trajectory satisfy
\[
 W'=2WR(W^\top W)B(\tau)\quad\text{for almost every }\tau.
\]
Assume the coefficients are locally bounded and the trajectory converges to $W_s$. If $\rank W(0)=d$, then
\begin{equation}\label{eq:rank-inertia}
 d+n_+(R(W_s^\top W_s))\le r.
\end{equation}
In particular this applies to $W'=H_{K,\eta}(G(W))$ for every fixed finite $K$ and $\eta\ge0$.
\end{lemma}

The proof uses the actual right-multiplication equation. A positive
quadratic cone associated with the terminal residual is forward invariant
once the residual is sufficiently close to its limit. Uniform positivity
of the multiplier forces exponential growth of every row combination
in that cone. A dimension argument supplies such a combination whenever
the inequality in \eqref{eq:rank-inertia} fails, contradicting convergence.
The derivative estimate keeps the order of all matrix products.

For the empirical loss, a separate stationary identity gives
$\rank M+n_+(R(M))\le n_+(C)$ whenever $M\succeq0$ and $MR(M)=0$.
The proof uses strict positivity of $\Lop(M)$ on $\Ran M$ and is given
in Lemma~\ref{lem:static-inertia}. The dynamic bound above instead uses
the rank before the terminal limit. For example, a full-column-rank
trajectory has a terminal residual with no positive eigenvalues and
therefore attains a PSD-optimal loss.

\subsection{A critical stationary loss band}
Under Assumption~\ref{ass:comparison}, define
\begin{equation}\label{eq:criticalset}
\begin{aligned}
 \CE_{\ge\phi_1}^{(q)}
 &=\{M\succeq0:\rank M\le q,\ MR(M)=0,
                   \ \phi_1\le\Phi(M)\le\Phi(0)\},\\
 p_*^{(q)}&=\min_{M\in\CE_{\ge\phi_1}^{(q)}}n_+(R(M)).
\end{aligned}
\end{equation}
This set is nonempty and compact, and
$1\le p_*^{(q)}\le n_+(R_1)\le n_+(C)-1$
(Lemma~\ref{lem:positive-residual}). Its lower bound follows because
$M_1$ admits a lower-loss PSD perturbation, whereas a stationary PSD
predictor with nonpositive residual is globally optimal.

Every selected continuation starts below $\Phi(0)$ and converges to a
stationary predictor of rank at most $d_K(b)$. If its terminal loss were
at least $\phi_1$, that endpoint would belong to the displayed band.
Lemma~\ref{lem:inertia} would then require
$d_K(b)+p_*^{(q)}\le r$. Consequently the strict reverse inequality
supplies the selected-loss premise of Theorem~\ref{thm:separation}.
This argument is formalized in Corollary~\ref{cor:rank-loss}.

\subsection{Capacity saturation and the depth tail}
The next bound makes every clock complete before the fixed physical
scale. It depends on the input spectrum, hidden width, and initialization
exponent. The sharper selected loss remains a property of the full
empirical continuation.
\begin{proposition}[Saturation at an explicit finite depth]\label{prop:capacity-tail}
Under Assumptions~\ref{ass:task} and~\ref{ass:prefix}, define
\begin{equation}\label{eq:capacity-depth}
 K_{\rm cap}(b)
 =1+\left\lfloor\log_{3/2}
 \frac{[1+2(p-1)\max\{1,b-1\}]
       (\sum_{i=1}^p\lambda_i^2)^{1/2}}{\lambda_p}
 \right\rfloor .
\end{equation}
For every fixed $b>1$ and every finite integer $K\ge K_{\rm cap}(b)$,
all candidate supports through rank $p$ exist, every finite candidate
clock is strictly below one, and Theorem~\ref{thm:passage} applies with
$d_K(b)=p$. For each fixed $b,c_\eta$, the Gaussian events can be
chosen simultaneously for all these integer depths.
Moreover, for each $B>1$, the set $\mathcal B_C\cap(1,B]$ is finite,
with cardinality at most $(p-1)K_{\rm cap}(B)$.
\end{proposition}
\begin{theorem}[Eventual avoidance under a residual-inertia condition]\label{thm:eventual-depth}
Fix a task and width satisfying Assumptions~\ref{ass:task},
\ref{ass:prefix}, and~\ref{ass:comparison}. Suppose, in addition, that
\begin{equation}\label{eq:task-inertia-condition}
 p+p_*^{(q)}>r,
 \qquad p=\min\{m,n_+(C)\},\quad q=\min\{m,r\}.
\end{equation}
For every fixed $b>1$ and $c_\eta>0$, every finite integer
$K\ge K_{\rm cap}(b)$ has
\[
 d_K(b)=p,\qquad \ell_{K,p}<\phi_1,
\]
and gives almost-sure small-initialization bypass in
Definition~\ref{def:bypass}.
More precisely, for every finite $K\ge K_{\rm cap}(b)$ there are
constants $d_{*,K},\Delta_{*,K}>0$. There is one Gaussian-full
event such that, for every $U$ in it and every such $K$, the following inequalities hold for every sufficiently small $\eps>0$,
with a cutoff allowed to depend on $K$ and $U$:
\begin{equation}\label{eq:eventual-actual}
 d_1(K,b,\eps,U)\ge d_{*,K},\qquad
 T_{\Delta_{*,K}}(K,b,\eps,U)<\infty
.
\end{equation}
Here $K_{\rm cap}(b)$ is the explicit sufficient cutoff in
\eqref{eq:capacity-depth}; the constants and initialization cutoffs
are allowed to vary with the fixed finite depth.
\end{theorem}

For each finite depth in the asserted tail, the proof first selects its
rank-$p$ entrance, applies the residual-inertia bound to its continuation,
and transfers the resulting strict loss decrease to the actual path.
A countable intersection then gives the common Gaussian-full event.
This preserves the depth-dependent constants and the seed-dependent
initialization cutoffs in the theorem.

%% file: sections/example.tex
\section{Numerical Experiments}\label{sec:example}
We compare gradient descent, finite-depth Newton--Schulz dynamics, and
polar updates on the same empirical quadratic task. The experiment
follows the approach to the first gradient-flow stationary point and
measures the closest point reached along each training trajectory.

We use $m=r=3$ and $24$ equally weighted samples. The first eight have
\[
 x_l=3^{1/4}(\pm1,\pm1,\pm\sqrt{3/20}),\qquad y_l=3\sqrt3,
\]
with all independent sign choices. The remaining samples consist of
four copies of each signed pair
\[
 \begin{aligned}
 x_l&=(0,\pm\sqrt3,0),& y_l&=-1,\\
 x_l&=(0,0,\pm(1173/400)^{1/4}),&
 y_l&=11\sqrt{3/391}.
 \end{aligned}
\]
For these data, $C=\diag(3,2,1)$. The first gradient-flow stationary
predictor in Proposition~\ref{main:gd} is $M_1=\diag(3,0,0)$, with
$\Phi(M_1)=377/1173$; the minimum PSD loss is $1/6$.

We initialize with $b=5/4$ and $c_\eta=1$. For each
$\eps\in\{10^{-3},10^{-5},\ldots,10^{-15}\}$, we compare GD,
$K=1,\ldots,6$, and an SVD-computed polar update. We draw $32$
independent standard Gaussian matrices $U$ and reuse this bank across
all methods and initialization scales, giving $1792$ trajectories.
GD is integrated as ordinary gradient flow and displayed using the
positive time change in Proposition~\ref{main:global}, so the left
panel uses the normalized clock of \eqref{eq:flow}. The polar branch
uses the polar factor of the full factor gradient at each numerical
step. All methods update the complete $3\times3$ factor.

\begin{figure}[tbp]
\centering
\includegraphics[width=\linewidth]{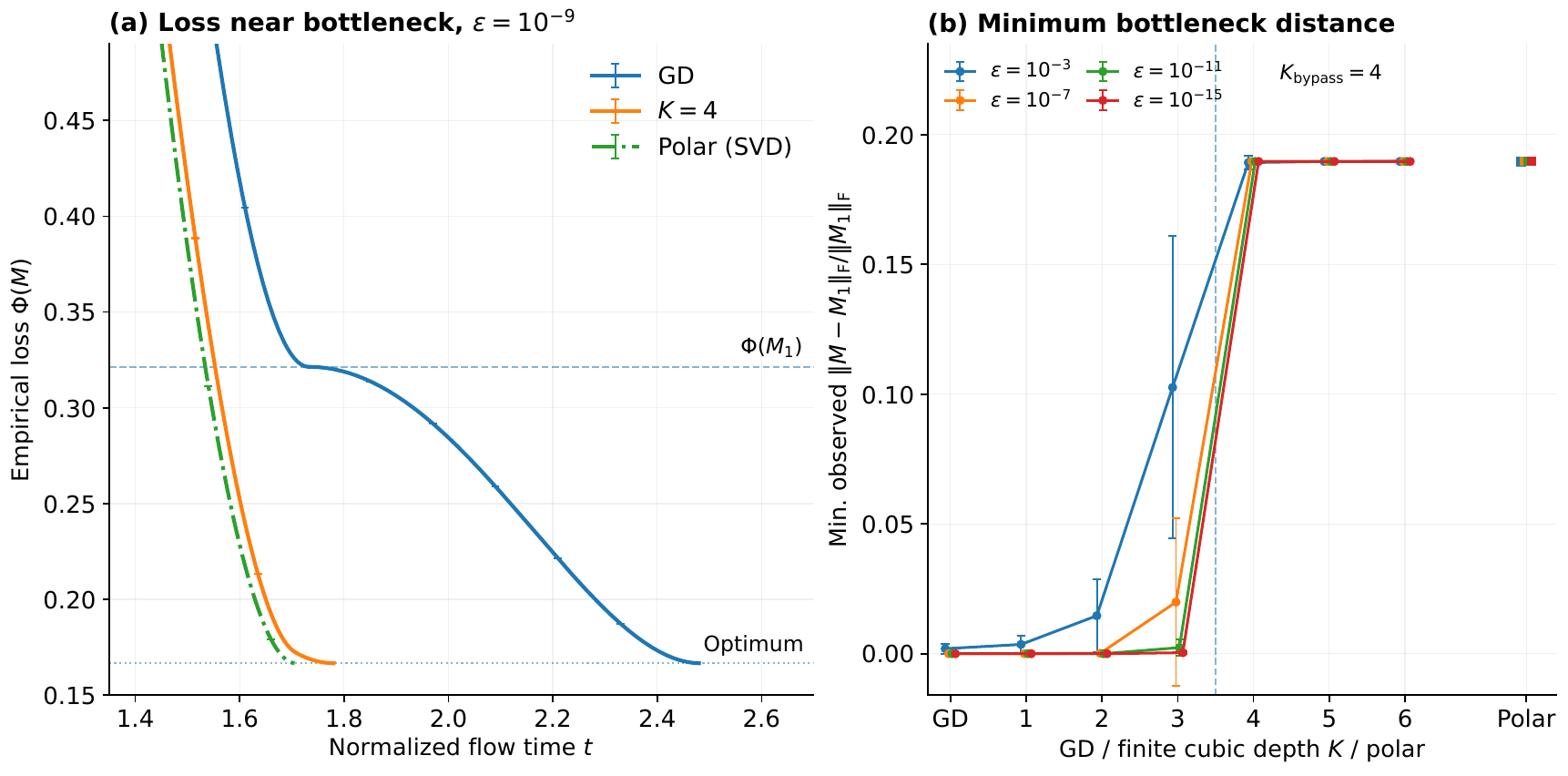}
\caption{Finite-initialization trajectories on the $24$-sample task.
(a) Loss at $\eps=10^{-9}$ for GD, depth four, and the SVD polar
reference. GD is shown in the normalized path time described in the
text. (b) Minimum observed relative distance to $M_1$ before the
numerical stopping condition, for four initialization scales.
Points and curves show means over $32$ Gaussian draws; error bars
show one sample standard deviation. Small horizontal offsets separate
overlapping error bars. Lines between integer depths guide the eye;
the polar reference is displayed separately.}
\label{fig:round34-intro}
\end{figure}

Theorem~\ref{thm:passage} gives entrance rank one at $K=1$, rank two
at $K=2,3$, and rank three at $K\ge4$ for this task. The rank-two
selected continuation lies in $\operatorname{span}(e_1,e_2)$ and
converges to $M_1$, so Corollary~\ref{thm:conditional-classification}
gives approach to $M_1$ by the corresponding small-initialization
paths. At rank three, Lemma~\ref{lem:inertia} yields PSD-optimal
selected loss, and Theorem~\ref{thm:separation} gives bypass.
The numerical comparison in Figure~\ref{fig:round34-intro} follows
this classification: the distance decreases with initialization scale
for GD and depths one to three, while depths four to six remain
separated from $M_1$. In the left panel, GD passes through a shoulder
near $\Phi(M_1)$; depth four and the polar reference descend through
that level without the same shoulder. All runs reach the common
near-optimal loss, consistent with the full-width conclusion of
Corollary~\ref{thm:conditional-classification} for GD and finite depth.

The GD and finite-depth equations are integrated with DOP853 in
continuation time while accumulating the normalized time. Polar is
integrated by explicit SVD updates with step-size control and
backtracking. Integration stops within $10^{-11}$ of the optimal loss
for GD and finite depth, and within $10^{-9}$ for polar. We check
smaller steps and tighter tolerances on repeated runs. The source,
Gaussian draws, per-run measurements, and numerical diagnostics are
provided with the experiment code.

%% file: sections/scope.tex
\section{Conclusion}\label{sec:discussion}
In this paper, we have studied how finitely many iterations of a nonlinear matrix transformation can reshape optimization dynamics. In a quadratic two-layer network, we have characterized the evolution from small-initialization escape near the origin to the first nonzero stationary stage. Our results provide a precise description of how the number of Newton–Schulz iterations controls early rank selection and regulates the subsequent optimization path. Beyond this particular model, the analysis offers a framework combining spectral dynamics with small-initialization techniques, which may help understand certain training phenomena arising in large-scale neural networks, including language models.

An immediate open question is whether similar dynamical descriptions can be established for more general models and under weaker assumptions. Of particular interest are networks with nonlinear activations such as ReLU, high-dimensional training regimes, and deeper architectures, where the interaction between spectral updates and feature learning may lead to richer dynamics.

%% file: appendices/foundations.tex
\section{Foundations and Shared Structural Tools}\label{app:foundations}\label{app:proofs}\label{sec:tools}
The empirical identities provide compact sublevels and positive
multipliers. These yield continuation, finite-time comparison, and rank
preservation. The last subsection proves the rank--inertia bound used
both for full-width optimality and for the sufficient depth theorem.

\subsection{Empirical identities, update bounds, and changes of time}
The next three statements give the algebraic and analytic properties of
the update before any spectral-prefix or comparison assumption is used.
\begin{lemma}[Empirical loss identities]\label{lem:empirical-identities}
The empirical loss satisfies
\begin{equation}\label{eq:loss-expand}
 \Phi(M)=\Phi(0)-\langle C,M\rangle+\tfrac12\langle M,\Lop(M)\rangle,
 \qquad \langle H,\Lop(H)\rangle=\frac1n\sum_l(x_l^\top Hx_l)^2\ge0.
\end{equation}
Moreover, $\Phi$ is convex, $\Lop$ is self-adjoint, $\Lop(M)\succeq0$ for $M\succeq0$, $\nabla\Phi(M)=-R(M)$, and $G(W)=2WR(W^\top W)$.
\end{lemma}
\begin{proof}(of Lemma~\ref{lem:empirical-identities}).
\label{proof:lem:empirical-identities}
Expand each summand as $y_l^2-2y_lx_l^\top Mx_l+(x_l^\top Mx_l)^2$. Since $x_l^\top Mx_l=\langle x_lx_l^\top,M\rangle$, the linear term is $-\langle C,M\rangle$ and the quadratic term is $\langle M,\Lop(M)\rangle/2$. For symmetric $A,B$,
\[
 \langle A,\Lop(B)\rangle
 =\frac1n\sum_l(x_l^\top Ax_l)(x_l^\top Bx_l)
 =\langle\Lop(A),B\rangle.
\]
This proves self-adjointness and the second identity. If $M\succeq0$, every coefficient of $x_lx_l^\top$ in $\Lop(M)$ is nonnegative. Differentiating the expansion now gives $\nabla\Phi(M)=\Lop(M)-C$. Finally, for a factor increment $E$,
\[
 Df(W)[E]=\langle-R(W^\top W),E^\top W+W^\top E\rangle
 =\langle-2WR(W^\top W),E\rangle,
\]
which gives the stated negative gradient. In particular $\Phi$ is convex.
\end{proof}

\begin{lemma}[Bounds and continuity of the update map]\label{lem:update-map}
Fix finite $K$. On $[0,1]$ the scalar polynomials satisfy
\[
 s\le p_K(s)\le a_Ks,\qquad 0\le p_K(s)\le1,\qquad
 1\le q_K(x)\le a_K.
\]
Moreover $p_K'(s)>0$ for $0<s<1$. For $K\ge1$, $q_K'(x)<0$ for $0<x<1$, and $q_K$ maps $[0,1]$ bijectively onto $[1,a_K]$ in decreasing order. For every matrix $X$,
\[
 P_K(X)=Xq_K(X^\top X)=q_K(XX^\top)X.
\]
The maps $H_{K,\eta}$, $\eta\ge0$, are globally Lipschitz in $G$ with a constant independent of $\eta$. They satisfy
\begin{align}
 H_{K,\eta}(G)&=GB^R,& I&\preceq B^R\preceq a_KI,\label{eq:multiplier}\\
 \ip G{H_{K,\eta}(G)}&\ge\norm G_F^2,&
 \norm{H_{K,\eta}(G)}_F&\le a_K\norm G_F,\label{eq:angle-bounds}\\
 \sup_G\norm{H_{K,\eta}(G)-H_{K,0}(G)}_F&\le C_K\eta.\label{eq:eta-bound}
\end{align}
Here the identity matrix has the appropriate size, and
\[
 B^R=q_K\!\left(\frac{G^\top G}{(\norm G_F+\eta)^2}\right)\in\Sym^r.
\]
The constant $C_K<\infty$ depends only on the fixed depth and matrix dimensions. At $G=0,\eta=0$, set $B^R=a_KI_r$.
\end{lemma}
\begin{proof}(of Lemma~\ref{lem:update-map}).
\label{proof:lem:update-map}
The polynomial identity in the statement follows by expanding $p_K$ into odd powers and multiplying the corresponding matrix monomials. Equivalently, a singular value decomposition applies the scalar polynomial to each singular value.

For $0\le s\le1$, the scalar map $s\mapsto(3s-s^3)/2$ takes $[0,1]$ into itself, is at least $s$, and is at most $3s/2$. Induction gives
\[
 s\le p_K(s)\le a_Ks,\qquad 0\le p_K(s)\le1.
\]
Thus $1\le q_K(x)\le a_K$ on $[0,1]$. The scalar cubic map has positive derivative on $(0,1)$, so composition gives $p_K'(s)>0$ there. Since $q_1(x)=3/2-x/2$, its derivative is negative. The recurrence
\[
 q_{K+1}(x)=q_K(x)\left\{\frac32-\frac12p_K(\sqrt{x})^2\right\}
\]
has a positive second factor; differentiating it shows that both summands in its derivative are negative whenever $q_K'(x)<0$, $0<x<1$. Induction proves strict decrease. The endpoint identities $q_K(0)=a_K$ and $q_K(1)=p_K(1)=1$ give the stated bijection.

The normalized singular values of $G$ belong to this interval. Applying the polynomial identity above in singular coordinates proves \eqref{eq:multiplier}. It also gives
\[
 \langle G,H_{K,\eta}(G)\rangle
 =\tr(G^\top G B^R)\ge\tr(G^\top G),\qquad
 \norm{GB^R}_F\le a_K\norm G_F,
\]
so \eqref{eq:angle-bounds} follows.

For the Lipschitz bound, $P_K$ and its first derivative are bounded on the unit Frobenius ball. At $G\ne0$, direct differentiation gives
\begin{equation}\label{eq:H-derivative}
\begin{aligned}
 DH_{K,\eta}(G)[E]
 ={}&DP_K\!\left(\frac{G}{\norm G_F+\eta}\right)[E]\\
 &+\frac{\langle G,E\rangle}{\norm G_F}
 \left\{P_K\!\left(\frac{G}{\norm G_F+\eta}\right)
 -DP_K\!\left(\frac{G}{\norm G_F+\eta}\right)
       \left[\frac{G}{\norm G_F+\eta}\right]\right\}.
\end{aligned}
\end{equation}
The normalized argument has norm at most one, and
$|\langle G,E\rangle|/\norm G_F\le\norm E_F$. Thus the derivative is bounded by $C_K\norm E_F$, independently of $\eta$. Also \eqref{eq:angle-bounds} implies continuity of $H_{K,\eta}$ at zero. Integrating along a line segment, split at zero when necessary, proves a common global Lipschitz bound.

For $G\ne0$, the parameter difference is
\begin{align*}
 H_{K,\eta}(G)-H_{K,0}(G)
 ={}&\eta P_K\!\left(\frac G{\norm G_F+\eta}\right)\\
 &+\norm G_F\left\{P_K\!\left(\frac G{\norm G_F+\eta}\right)
 -P_K\!\left(\frac G{\norm G_F}\right)\right\},\\
 \left\|\frac G{\norm G_F+\eta}-\frac G{\norm G_F}\right\|_F
 ={}&\frac\eta{\norm G_F+\eta}.
\end{align*}
The first term is $O_K(\eta)$; the second is bounded by
$C_K\norm G_F\eta/(\norm G_F+\eta)\le C_K\eta$.
Both maps vanish at $G=0$, proving \eqref{eq:eta-bound}.
\end{proof}

\begin{lemma}[Positive change of time]\label{lem:time-change}
Fix $\eta>0$. If $W(t)$ solves the physical-time equation with denominator $\norm{G(W)}_F+\eta$, define
\[
 \tau(t)=\int_0^t\frac{\dd s}{\norm{G(W(s))}_F+\eta}.
\]
Then $\tau(t)$ is strictly increasing, and
\begin{equation}\label{eq:time-change}
 \frac{\dd t}{\dd\tau}=\norm{G(W(t(\tau)))}_F+\eta,
 \qquad \frac{\dd}{\dd\tau}W(t(\tau))=H_{K,\eta}(G(W(t(\tau)))).
\end{equation}
On bounded forward solutions this change of time maps the whole forward half-line onto itself and preserves finite hitting times and the infimum of distance over the path. At $K=0$, $H_{0,\eta}(G)=G$ for every $\eta\ge0$.
\end{lemma}
\begin{proof}(of Lemma~\ref{lem:time-change}).
\label{proof:lem:time-change}
The integrand is positive and continuous, so $\tau(t)$ is strictly increasing and has an inverse on its image. The chain rule gives
\[
 \frac{\dd W(t(\tau))}{\dd\tau}
 =\frac{\dd W}{\dd t}\frac{\dd t}{\dd\tau}
 =(\norm{G(W(t(\tau)))}_F+\eta)
 P_K\!\left(\frac{G(W(t(\tau)))}{\norm{G(W(t(\tau)))}_F+\eta}\right).
\]

This proves \eqref{eq:time-change}. On a bounded trajectory $G(W)$ is bounded, so $\eta\le\dd t/\dd\tau\le\sup_\tau\norm{G(W(\tau))}_F+\eta$. Integrating these inequalities in either direction shows that the time variables cover the same forward half-line and that finite times correspond to finite times. The ordered set of matrices visited is the same, proving the hitting-time and distance claims. Finally $P_0(X)=X$ gives $H_{0,\eta}(G)=G$ directly.
\end{proof}

The two rescalings introduced in Section~\ref{sec:entrance} retain the
empirical operator and denominator at their respective scales. The error
bound below is local in a region with a positive limiting denominator.
\begin{lemma}[Exact rescaling and local error]\label{lem:rescaling}
The initial-layer and shape equations are
\begin{equation}\label{eq:initial-layer}
 \begin{aligned}
 \frac{\dd Y^\eps}{\dd\sigma}
 &=P_K\!\left(\frac{2Y^\eps[C-\eps^2\Lop((Y^\eps)^\top Y^\eps)]}
 {2\norm{Y^\eps[C-\eps^2\Lop((Y^\eps)^\top Y^\eps)]}_F+c_\eta}\right),\\
 Y^\eps(0)&=\diag(1,\eps^{b-1},\ldots,\eps^{b-1})U,
 \end{aligned}
\end{equation}
and
\begin{equation}\label{eq:actual-rescaled}
 \frac{\dd Q^\eps}{\dd v}
 =P_K\!\left(\frac{Q^\eps[C-t^2\Lop((Q^\eps)^\top Q^\eps)]}
 {\norm{Q^\eps[C-t^2\Lop((Q^\eps)^\top Q^\eps)]}_F+c_\eta\eps/(2t)}\right)-Q^\eps,
 \qquad t=e^v.
\end{equation}
On a fixed bounded $Q$ region with $\inf_Q\norm{QC}_F>0$,
\begin{equation}\label{eq:rescaling-error}
 \norm{\text{right side of \eqref{eq:actual-rescaled}}-F_K}_{C^1}
 =O(t^2+\eps/t)
\end{equation}
for all sufficiently small $t>0$ and $0\le\eps/t\le1$.
The implied bound and the allowed upper bound on $t$ depend only on this
region and the fixed model parameters.
\end{lemma}
\begin{proof}(of Lemma~\ref{lem:rescaling}).
\label{proof:lem:rescaling}
The chain rule gives
\[
 \frac{\dd Q}{\dd v}
 =t\frac{\dd}{\dd t}(W/t)=\dot W-Q.
\]
The exact gradient substitution is
\[
 G(tQ)=2tQ\{C-t^2\Lop(Q^\top Q)\},\qquad
 \frac{G(tQ)}{\norm{G(tQ)}_F+c_\eta\eps}
 =\frac{Q[C-t^2\Lop(Q^\top Q)]}{\norm{Q[C-t^2\Lop(Q^\top Q)]}_F+c_\eta\eps/(2t)}.
\]
Together with $Q'=\dot W-Q$, this proves \eqref{eq:actual-rescaled}.

For the estimate, choose $\kappa>0$ such that $\norm{QC}_F\ge\kappa$ on the stated region, and use the induced Frobenius operator norm of $\Lop$. On $\norm Q_F\le B$,
\[
 \norm{Q\Lop(Q^\top Q)}_F\le\norm\Lop B^3,
 \quad
 D[Q\Lop(Q^\top Q)][E]
 =E\Lop(Q^\top Q)+Q\Lop(E^\top Q+Q^\top E).
\]
The derivative has norm at most $3\norm\Lop B^2\norm E_F$. Therefore $Q[C-t^2\Lop(Q^\top Q)]-QC$ and its first derivative are $O(t^2)$ uniformly on the region. For all sufficiently small $t$, $\norm{Q[C-t^2\Lop(Q^\top Q)]}_F\ge\kappa/2$.

On matrices $A$ in a fixed bounded set with $\norm A_F\ge\kappa/2$ and scalars $0\le z\le c_\eta/2$, the function $A/(\norm A_F+z)$ is smooth in $(A,z)$ and has bounded derivatives through order two. Its derivative in a matrix direction $E$ is
\[
 \frac{E}{\norm A_F+z}
 -\frac{A\langle A,E\rangle}{\norm A_F(\norm A_F+z)^2}.
\]
The denominator lower bound controls this expression and its next derivatives. Applying the mean value estimate to $A=Q[C-t^2\Lop(Q^\top Q)]$, $z=c_\eta\eps/(2t)$ and to $A=QC$, $z=0$ gives a $C^1$ difference of order $t^2+\eps/t$. Finally $P_K$ has bounded derivatives through order two on the relevant bounded set. Composition proves \eqref{eq:rescaling-error}. The same argument applies to every fixed bounded range of $\eps/t$, with a bound depending on that range.

For the initial scale, the chain rule gives
\[
 \frac{\dd Y^\eps}{\dd\sigma}
 =\frac1\eps\frac{\dd W(\eps\sigma)}{\dd\sigma}
 =\dot W(\eps\sigma).
\]
Substituting $W=\eps Y$ gives
$G(\eps Y)=2\eps Y[C-\eps^2\Lop(Y^\top Y)]$.
Cancellation of the positive factor $\eps$ from the stabilized
normalization, together with
$\eps^{-1}D_{\eps,b}=\diag(1,\eps^{b-1},\ldots,\eps^{b-1})$,
proves \eqref{eq:initial-layer}.
\end{proof}

\subsection{Global continuation and finite-time comparison}
Compact empirical sublevels bound the factors. The update inequalities
then supply both a descent angle and a uniform finite-interval
perturbation estimate in the regularization parameter.
\begin{proof}(of Proposition~\ref{main:global}).
\label{proof:main:global}
The spanning assumption gives $n^{-1}\sum_lx_lx_l^\top\succ0$. For $M\succeq0$, all $x_l^\top Mx_l$ are nonnegative. The prediction-vector triangle inequality and Cauchy--Schwarz give
\begin{equation}\label{eq:compact}
 \lambda_{\min}\!\left(\frac1n\sum_lx_lx_l^\top\right)\tr M
 \le\frac1n\sum_lx_l^\top Mx_l
 \le\left(\frac1n\sum_ly_l^2\right)^{1/2}+\sqrt{2\Phi(M)}.
\end{equation}
Since $\norm M_F\le\tr M$ and $\norm W_F^2=\tr(W^\top W)$, closed PSD and factor sublevels are compact.

The identity $G=-\nabla f$, the chain rule, and \eqref{eq:angle-bounds} yield
\[
 -f'=\langle G,H_{K,\eta}(G)\rangle\ge\norm G_F^2,
 \qquad \norm{W'}_F\le a_K\norm G_F.
\]
Their combination proves all three inequalities in \eqref{eq:descent}. The vector field is locally Lipschitz by Lemma~\ref{lem:update-map}. Each solution remains in its compact initial sublevel, so the continuation theorem of \citet[Corollary~2.15]{teschl2012ordinary} gives global existence.

For any bounded $C^1$ curve satisfying \eqref{eq:descent}, the angle inequality is
\[
 \frac{\dd}{\dd\tau}f(W(\tau))
 \le-a_K^{-1}\norm{\nabla f(W(\tau))}_F\norm{W'(\tau)}_F.
\]
A zero loss derivative forces the gradient and then the velocity to vanish. The angle and weak-decrease hypotheses of \citet[Theorem~2.2]{absil2005convergence} therefore hold. Analyticity on a neighborhood of the compact closure and boundedness give convergence. The limit is stationary: otherwise continuity bounds the gradient norm away from zero on the tail, and $-f'\ge\norm{\nabla f}_F^2$ contradicts the lower bound on the loss. This proves both the empirical assertion and the stated bounded-curve assertion.

For a nonstationary solution, local uniqueness implies $G(W(\tau))\ne0$ at each finite time. Thus the coefficient in
\[
 W'=2WR(W^\top W)B^R
\]
 is continuous and bounded on finite intervals, including when $\eta=0$. Multiplication on the left by a fixed row vector gives a solution of the same linear row equation. Forward and backward uniqueness imply that this row solution vanishes at a finite time exactly when it vanishes initially. The left nullspace of $W$ is consequently unchanged, proving finite-time rank preservation. A stationary solution has constant rank directly.

For $\eta>0$, Lemma~\ref{lem:time-change} identifies the physical-time solution with this continuation. On its compact sublevel, $\dd t/\dd\tau$ is bounded above and is at least $\eta$. Both forward time half-lines are covered. The same factor limit and finite-time rank therefore hold in physical time. At $K=0$ the continuation is factor gradient flow.
\end{proof}

\begin{corollary}[Finite-time comparison]\label{cor:flow-comparison}
Let $W_\eta$ and $\bar W$ be solutions of
\[
 W_\eta'=H_{K,\eta}(G(W_\eta)),\qquad
 \bar W'=H_{K,0}(G(\bar W)).
\]
Fix finite $K$ and $\eta\ge0$. The two initial values may differ. Suppose both solutions stay in a fixed bounded factor region for $0\le\tau\le S$, where $S<\infty$. There is a common Lipschitz constant $L$ for the composed vector fields on a ball containing this region, and
\begin{equation}\label{eq:gronwall}
 \sup_{0\le\tau\le S}\norm{W_\eta(\tau)-\bar W(\tau)}_F
 \le e^{LS}\bigl(\norm{W_\eta(0)-\bar W(0)}_F+C_K\eta S\bigr).
\end{equation}
\end{corollary}
\begin{proof}(of Corollary~\ref{cor:flow-comparison}).
\label{proof:cor:flow-comparison}
The polynomial map $G$ is Lipschitz on a bounded ball. Lemma~\ref{lem:update-map} gives a common Lipschitz constant $L$ for its compositions, uniformly in $\eta$. Subtracting the integral equations and adding and subtracting $H_{K,\eta}(G(\bar W))$ yields
\[
 \norm{W_\eta(\tau)-\bar W(\tau)}_F
 \le\norm{W_\eta(0)-\bar W(0)}_F
 +\int_0^\tau\bigl(L\norm{W_\eta(u)-\bar W(u)}_F+C_K\eta\bigr)\dd u.
\]
The nonhomogeneous form of \citet[Lemma~2.7]{teschl2012ordinary} gives
\begin{align*}
 \norm{W_\eta(\tau)-\bar W(\tau)}_F
 &\le e^{L\tau}\norm{W_\eta(0)-\bar W(0)}_F
 +C_K\eta\int_0^\tau e^{L(\tau-u)}\dd u\\
 &\le e^{LS}\bigl(\norm{W_\eta(0)-\bar W(0)}_F+C_K\eta S\bigr).
\end{align*}
Taking the supremum proves \eqref{eq:gronwall}.
\end{proof}

\begin{lemma}[Linearization at a stationary factor]\label{lem:stationary-linearization}
Fix $\eta>0$ and finite $K$. The physical-time vector field vanishes exactly at the stationary factors of $f$. At any such factor $W_s$, its derivative is
\[
 -\frac{a_K}{\eta}\nabla^2f(W_s).
\]
Thus its linearized eigenvalue signs agree with those of negative gradient flow, including zero eigenvalues.
\end{lemma}
\begin{proof}(of Lemma~\ref{lem:stationary-linearization}).
\label{proof:lem:stationary-linearization}
For stationarity, the normalized singular values lie in $[0,1)$ and $p_K(s)>0$ for $s>0$ in this interval. Consequently the update vanishes exactly when $G=0$. At a stationary factor, the derivative of $G/(\norm G_F+\eta)$ is $DG/\eta$, and $DP_K(0)=a_KI$. Since $DG(W_s)=-\nabla^2f(W_s)$, the asserted derivative follows. Multiplication by the positive scalar $a_K/\eta$ preserves signs, including zero eigenvalues.
\end{proof}

\begin{lemma}[Fixed-denominator expansion]\label{lem:fixed-denominator}
For fixed finite $K$ and fixed $\eta>0$,
\[
 P_K\!\left(\frac{G}{\norm G_F+\eta}\right)
 =\frac{a_K}{\eta}G+O_K\!\left(\frac{\norm G_F^2}{\eta^2}\right)
 \qquad(\norm G_F/\eta\to0).
\]
\end{lemma}
\begin{proof}(of Lemma~\ref{lem:fixed-denominator}).
\label{proof:lem:fixed-denominator}
\label{app:fixed-eta}
Each cubic composition is odd and has derivative $a_KI$ at zero. Therefore $P_K(X)=a_KX+O_K(\norm X_F^3)$ on a fixed neighborhood of zero. Also
\[
 \norm{\frac{G}{\norm G_F+\eta}-\frac G\eta}_F
 =\frac{\norm G_F^2}{\eta(\norm G_F+\eta)}\le\frac{\norm G_F^2}{\eta^2}.
\]
The nonlinear polynomial remainder is $O_K((\norm G_F/\eta)^3)$, which is bounded by $O_K((\norm G_F/\eta)^2)$ as $\norm G_F/\eta\to0$. Combining the two estimates proves the claim.
\end{proof}

\subsection{Cited analytic tools and the normalized leading-direction limit}
Local solution, uniqueness, dependence, and continuation results are used
in the forms of \citet[Theorems~2.2, 2.9, 2.11, 2.13 and
Corollary~2.15]{teschl2012ordinary}. At each application the vector field
is locally Lipschitz on the indicated region, and compact confinement
extends the solution as needed. The integral comparison is
\citet[Lemma~2.7]{teschl2012ordinary}; the nonhomogeneous form follows by
including the accumulated forcing in the additive term and interchanging
the integrals. For measurable coefficients the integrating-factor
argument applies to absolutely continuous primitives.

The weighted function spaces used below are Banach spaces after
multiplication by their displayed weight. Each application of
\citet[Theorem~2.1]{teschl2012ordinary} verifies invariance of a closed
ball and a strict contraction estimate. Nonzero determinant polynomials
have null zero sets by \citet{mityagin2020zero}; an identity submatrix
witnesses their nontriviality, and Gaussian density transfers nullity to
probability zero.

The next lemma states the normalization of the local exit result needed
in the empirical passage argument. Its proof retains the asymptotic,
uniqueness, and compact-interval comparison steps around the cited
shifted-limit theorem.
\begin{lemma}[Limit in a simple leading direction]\label{lem:local-exit-background}

Let $g:\R^N\to\R^N$ be $C^2$, with $g(0)=0$, and let $\psi(t,z)$ be the flow of $z'=g(z)$, defined for every $(t,z)\in\R\times\R^N$. Suppose $Dg(0)$ has an algebraically simple real eigenvalue $\alpha_*>0$ with real part strictly larger than every other eigenvalue. Let $r_*$ and $l_*$ be corresponding right and left eigenvectors with $l_*^\top r_*=1$.
There is a common nonconstant solution with $e^{-\alpha_* t}z_*(t)\to r_*$ as $t\to-\infty$, given by
\[
 z_*(t)=\lim_{\alpha\downarrow0}
 \psi\!\left(t+\alpha_*^{-1}\log(1/\alpha),\alpha r_*\right).
\]
For every family $z_\alpha\to0$ with
\[
 \liminf_{\alpha\downarrow0}
 \frac{l_*^\top z_\alpha}{\norm{z_\alpha}}>0,
\]
and every fixed $t\in\R$,
\[
 \psi\!\left(t+\alpha_*^{-1}\log\frac1{l_*^\top z_\alpha},z_\alpha\right)
 \longrightarrow z_*(t).
\]
\end{lemma}
\begin{proof}(of Lemma~\ref{lem:local-exit-background}).
\label{proof:lem:local-exit-background}
Theorem~E.7 of \citet{li2021greedy} gives the common shifted limit under
the stated positive left-projection condition. It allows non-diagonalizable
remaining spectral blocks. To relate its normalization to the one used
here, choose $C\ge\max\{1,\|r_*\|\}$ sufficiently large that
\[
 \|e^{sDg(0)}\|\le Ce^{\alpha_*s}\quad(s\ge0),\qquad
 \|g(z)-Dg(0)z\|\le C\|z\|^2
\]
in a fixed ball about zero for the second estimate. The first estimate
follows from the simple leading eigenvalue and the strict gap in real
parts, including all nonleading Jordan blocks. Variation of constants gives
\[
 \|\psi(s,\alpha r_*)-\alpha e^{\alpha_*s}r_*\|
 \le C^2\int_0^s e^{\alpha_*(s-u)}\|\psi(u,\alpha r_*)\|^2\,\mathrm du.
\]
Under the temporary bound $\|\psi(u,\alpha r_*)\|\le
2C\alpha e^{\alpha_*u}$, the right side is at most
\[
 \frac{4C^4}{\alpha_*}\alpha^2e^{\alpha_*s}
       (e^{\alpha_*s}-1).
\]
If $\alpha e^{\alpha_*s}$ is sufficiently small, depending only on the
fixed field and eigenvectors, this strictly improves the temporary bound
and keeps the solution inside the chosen ball. A first-exit argument
therefore proves the bound throughout such an interval. Substitute
$s=t+\alpha_*^{-1}\log(1/\alpha)$ and let $\alpha\downarrow0$.
For every sufficiently negative fixed $t$, Theorem~E.7 yields
\[
 \|z_*(t)-e^{\alpha_*t}r_*\|
 \le \frac{4C^4}{\alpha_*}e^{2\alpha_*t}.
\]
This proves the normalized leading asymptotic and nonconstancy.
The limit solves the differential equation: convergence at a fixed time,
together with finite-time continuous dependence, identifies all other
limits with the solution from that limiting value. Completeness gives its
extension to every real time. For any compact interval, apply Theorem~E.7
at its left endpoint and then finite-time continuous dependence on the
whole interval. This upgrades the convergence to uniform convergence on
each compact time interval.

The normalized asymptotic identifies the solution
uniquely. If $z$ is another solution with
$e^{-\alpha_*s}z(s)\to r_*$ as $s\to-\infty$, then
$l_*^\top z(s)=e^{\alpha_*s}(1+o(1))$ and
$l_*^\top z(s)/\|z(s)\|\to1/\|r_*\|>0$.
Apply Theorem~E.7 to the initial family $z(s)$ as $s\to-\infty$.
For each fixed $t$, its shifted flow is
\[
 \psi\!\left(t+\alpha_*^{-1}\log\frac1{l_*^\top z(s)},z(s)\right)
 =z\!\left(s+t+\alpha_*^{-1}\log\frac1{l_*^\top z(s)}\right)
 =z(t+o(1)).
\]
Its limit is both $z_*(t)$ by the theorem and $z(t)$ by continuity.
Hence the two solutions agree. All applications must still verify the
smooth completion, spectral gap and positive left projection required
by the cited theorem.
\end{proof}

\subsection{Rank--inertia and PSD optimality}
Finite-time row relations are preserved by a right-linear equation.
The terminal residual then determines a quadratic cone in which a row
would have to grow. The following proof applies with a measurable
positive multiplier and retains its noncommuting products.
\begin{proof}(of Lemma~\ref{lem:inertia}).
\label{proof:lem:inertia}
For any fixed coefficient vector $c\in\R^m$, the row $c^\top W$ is
locally absolutely continuous and satisfies the same right-linear
integral equation as every row of $W$, with coefficient
$2R(W^\top W)B$. On a finite interval this coefficient is measurable
and bounded. Forward and backward Gronwall inequalities imply that a
row combination vanishes at one endpoint exactly when it vanishes
throughout that interval. Thus the linear relations among the rows
are preserved, and the row space has dimension $d$ at every finite time.

Write $D=R(W_s^\top W_s)$. If $n_+(D)=0$, the conclusion is $d\le r$.
Otherwise, let $P_0$ be the orthogonal projector onto $\ker D$, and let
$\sigma>0$ be the smallest nonzero absolute eigenvalue of $D$. Set
\[
 \alpha=\min\left\{1,\frac{\sigma^2}{16a^2\norm D_{\op}}\right\},
 \qquad J=D-\alpha P_0.
\]
The matrix $J$ is nonsingular and has $n_+(D)$ positive eigenvalues.
Moreover $\alpha\le\norm D_{\op}$, so $\norm J_{\op}=\norm D_{\op}$.
Use the quadratic form $\mathcal Q(z)=zJz^\top$ on row vectors, and
apply it to $c^\top W$ with $c$ fixed. On the cone
$\mathcal Q(c^\top W)\ge0$, orthogonality of the kernel and its
complement gives
\[
 \alpha\norm{c^\top WP_0}^2
 \le c^\top WDW^\top c
 \le\norm D_{\op}\norm{c^\top W(I-P_0)}^2,
 \qquad
 \norm{c^\top W(I-P_0)}\le\sigma^{-1}\norm{c^\top WD}.
\]
Consequently,
\[
 \norm{c^\top W}^2
 \le\frac{1+\norm D_{\op}/\alpha}{\sigma^2}
       \norm{c^\top WD}^2.
\]
Differentiation using the actual residual and $DP_0=P_0D=0$ yields
\begin{equation}\label{eq:cone-derivative}
\begin{aligned}
 \frac{\dd}{\dd\tau}\mathcal Q(c^\top W)
 ={}&4c^\top WDBDW^\top c
       -4\alpha c^\top WDBP_0W^\top c\\
    &+4c^\top W\bigl(R(W^\top W)-D\bigr)BJW^\top c.
\end{aligned}
\end{equation}
The first term is at least $4\norm{c^\top WD}^2$. The absolute value
of the second is at most
\[
 4\alpha a\norm{c^\top WD}\norm{c^\top WP_0}
 \le\frac{4a\sqrt{\alpha\norm D_{\op}}}{\sigma}
       \norm{c^\top WD}^2
 \le\norm{c^\top WD}^2.
\]
The last term is bounded in absolute value by
\[
 4a\norm{R(W^\top W)-D}_{\op}\norm D_{\op}
 \frac{1+\norm D_{\op}/\alpha}{\sigma^2}\norm{c^\top WD}^2.
\]
Its coefficient tends to zero. After a sufficiently late time, chosen
independently of $c$, the cone therefore satisfies almost everywhere
\begin{equation}\label{eq:cone-growth}
 \frac{\dd}{\dd\tau}\mathcal Q(c^\top W)
 \ge2\norm{c^\top WD}^2\ge2\sigma\mathcal Q(c^\top W).
\end{equation}
For the second inequality, diagonalization gives $D^2\succeq\sigma D$:
each positive eigenvalue is at least $\sigma$, and each nonpositive
eigenvalue gives a nonpositive right side. Also
$c^\top WDW^\top c\ge\mathcal Q(c^\top W)$.

On an interval where this quadratic form is positive, its logarithmic
derivative is at least $2\sigma$ almost everywhere. Integration before
a possible first zero preserves positivity and excludes that zero by
continuity. The same inequality then forces exponential growth for all
later times. If $d>r-n_+(D)$, the row space at a sufficiently late finite
time contains a vector on which $J$ is positive. Indeed, its projection
onto the negative eigenspace of $J$ cannot be injective; a nonzero vector
in the kernel belongs to the positive eigenspace. Represent this vector
as $c^\top W$ and keep $c$ fixed thereafter. The resulting exponential
growth contradicts boundedness of the convergent trajectory. Hence
$d+n_+(D)\le r$.
\end{proof}

\begin{lemma}[Stationary rank and residual inertia]\label{lem:static-inertia}
For a spanning empirical task in any positive factor dimensions,
every $M\succeq0$ with $MR(M)=0$ satisfies
\begin{equation}\label{eq:static-inertia}
 \rank M+n_+(R(M))\le n_+(C).
\end{equation}
In particular, under Assumption~\ref{ass:comparison},
$1\le n_+(R_1)\le n_+(C)-1$.
\end{lemma}
\begin{proof}(of Lemma~\ref{lem:static-inertia}).
\label{proof:lem:static-inertia}
The empirical formula gives $\Lop(M)\succeq0$. For any nonzero
$u\in\Ran M$, the spanning assumption supplies a sample with
$x_l^\top u\ne0$. For this sample $x_l^\top Mx_l>0$: otherwise
positive semidefiniteness would imply $Mx_l=0$, making $x_l$
orthogonal to $\Ran M$. Consequently
\[
 u^\top\Lop(M)u
 =\frac1n\sum_l(x_l^\top Mx_l)(x_l^\top u)^2>0.
\]
Symmetry and $MR(M)=0$ give $R(M)M=0$, so $\Ran M\subset\ker R(M)$.
For $u\in\Ran M$ and $v$ in the positive eigenspace of $R(M)$,
\[
 (u+v)^\top C(u+v)
 =v^\top R(M)v+(u+v)^\top\Lop(M)(u+v).
\]
This is positive whenever $(u,v)\ne(0,0)$, by the preceding positivity
when $v=0$ and by the first term when $v\ne0$. Thus $C$ is positive
definite on the orthogonal direct sum of these two spaces. Projection
of that sum onto the positive eigenspace of $C$ is injective; a nonzero
vector in its kernel would lie in the nonpositive eigenspace of $C$.
The dimension bound is precisely
$\rank M+n_+(R(M))\le n_+(C)$.
At $M=M_1$, Proposition~\ref{main:gd} gives rank one and stationarity,
and Assumption~\ref{ass:comparison} gives $n_+(R_1)\ge1$.
\end{proof}

\begin{lemma}[PSD optimality conditions]\label{lem:psd-optimality}
If $M_s\succeq0$ satisfies $M_sR(M_s)=0$ and $R(M_s)\preceq0$,
then $\Phi(M)\ge\Phi(M_s)$ for every $M\succeq0$.
\end{lemma}
\begin{proof}(of Lemma~\ref{lem:psd-optimality}).
\label{proof:lem:psd-optimality}
For any $M\succeq0$, convexity and $\nabla\Phi(M_s)=-R(M_s)$ give
\[
 \Phi(M)\ge\Phi(M_s)+\ip{-R(M_s)}{M-M_s}\ge\Phi(M_s).
\]
The second inequality uses $-R(M_s)\succeq0$ and
$\ip{R(M_s)}{M_s}=0$.
\end{proof}

\begin{corollary}[Full-column-rank convergence to a PSD optimum]\label{cor:full-rank-optimum}
Fix finite $K$ and $\eta\ge0$. If $W(0)$ has column rank $r$, the solution
of $W'=H_{K,\eta}(G(W))$ converges to a factor $W_s$ whose predictor
$W_s^\top W_s$ minimizes $\Phi$ over $\Sym^r_+$. For every fixed
$\eps>0$, the same conclusion holds for the actual positive-regularization
trajectory whenever its initial factor has column rank $r$.
\end{corollary}
\begin{proof}(of Corollary~\ref{cor:full-rank-optimum}).
\label{proof:cor:full-rank-optimum}
Proposition~\ref{main:global} gives a stationary limit $W_s$. The initial column rank $r$ and Lemma~\ref{lem:inertia}
give $R(W_s^\top W_s)\preceq0$. Stationarity gives $(W_s^\top W_s)R(W_s^\top W_s)=0$, so
Lemma~\ref{lem:psd-optimality} applies. For each actual trajectory with
fixed $\eps>0$, the time change \eqref{eq:time-change} covers the complete
forward path, which gives the same conclusion in physical time.
\end{proof}

%% file: appendices/entrance.tex
\section{Gaussian Entrance and Actual Passages}\label{app:entrance}
The proof first establishes independent Gaussian coefficients on the
fixed initial scale. It then propagates their relative volumes through
each actual passage, including the empirical and regularization terms.
The final residence selects the boundary-value entrance of
Theorem~\ref{thm:passage}.

\subsection{Initial rows and independent Gaussian components}\label{sec:five}\label{sec:main-rescaling}
The row expansion is taken on a fixed finite initial-layer interval.
Its leading row determines the coefficients of a linear equation for the
weaker rows; eliminating their parallel components preserves a
nonsingular conditional Gaussian law.
Write the rows of $U$ as $u^\top,u_2^\top,\ldots,u_m^\top$, and the rows
of $Y^\eps(\sigma)$ as $y_a^\eps(\sigma)^\top$.
Define $w(\sigma),h_a(\sigma)\in\R^r$ by
\begin{equation}\label{eq:row-equations}
 \begin{aligned}
 \frac{\dd w(\sigma)^\top}{\dd\sigma}&=w(\sigma)^\top A(\sigma),
 &w(0)&=u,\\
 \frac{\dd h_a(\sigma)^\top}{\dd\sigma}&=h_a(\sigma)^\top A(\sigma),
 &h_a(0)&=u_a\quad(2\le a\le m),
 \end{aligned}
\end{equation}
where
\begin{equation}\label{eq:A-initial}
 A(\sigma)=\frac{2C}{2\norm{w(\sigma)^\top C}+c_\eta}
 q_K\!\left(\frac{4Cw(\sigma)w(\sigma)^\top C}
 {(2\norm{w(\sigma)^\top C}+c_\eta)^2}\right).
\end{equation}
In the first row, the same equation reads
\begin{equation}\label{eq:dominant-row}
 \frac{\dd w(\sigma)^\top}{\dd\sigma}
 =p_K\!\left(\frac{2\norm{w(\sigma)^\top C}}
 {2\norm{w(\sigma)^\top C}+c_\eta}\right)
 \frac{w(\sigma)^\top C}{\norm{w(\sigma)^\top C}}.
\end{equation}
\begin{lemma}[Finite-interval row expansion]\label{main:rows}
Under Assumption~\ref{ass:task}, fix $K,b,c_\eta$, a seed with
$u^\top v_1\ne0$, and a finite $S>0$.
The row equations are defined on $[0,\infty)$ and
$w(\sigma)^\top v_1\ne0$ for every $\sigma\ge0$.
Uniformly for $0\le\sigma\le S$,
\begin{equation}\label{eq:row-expansion}
 \begin{aligned}
 y_1^\eps(\sigma)&=w(\sigma)+O_{U,S}(\eps^{2(b-1)}+\eps^2),\\
 y_a^\eps(\sigma)&=\eps^{b-1}
 \bigl[h_a(\sigma)+O_{U,S}(\eps^{2(b-1)}+\eps^2)\bigr],
 \qquad 2\le a\le m.
 \end{aligned}
\end{equation}
\end{lemma}
\begin{proof}(of Lemma~\ref{main:rows}).
\label{proof:main:rows}
First consider the first equation in \eqref{eq:row-equations}. The
matrix polynomial identity in Lemma~\ref{lem:update-map}, applied to a
one-row matrix, gives \eqref{eq:dominant-row}. Introduce a scalar time
$\vartheta$ by
\[
 \frac{\dd\vartheta}{\dd\sigma}
 =\frac{p_K(2\norm{w(\sigma)^\top C}/(2\norm{w(\sigma)^\top C}+c_\eta))}{\norm{w(\sigma)^\top C}},
 \qquad\vartheta(0)=0.
\]
As long as the solution is defined, this number is positive and
\[
 w(\sigma)^\top=u^\top e^{\vartheta(\sigma)C},\qquad
 w(\sigma)^\top v_1=(u^\top v_1)e^{\lambda_1\vartheta(\sigma)}.
\]
In particular the last quantity never vanishes, and $\norm{w(\sigma)^\top C}>0$.
The bound $p_K(s)\le1$ on $[0,1]$ gives
$\norm{\dd w/\dd\sigma}\le1$ and hence
$\norm{w(\sigma)}\le\norm u+\sigma$. A finite endpoint of the solution
would therefore have a bounded state and
$|w(\sigma)^\top v_1|\ge|u^\top v_1|>0$; the locally smooth vector field
extends through that endpoint by \citet[Theorems~2.2, 2.9, 2.11, 2.13 and Corollary~2.15]{teschl2012ordinary}.
Thus $w$ exists on the whole positive half-line. On each compact interval
$A(\sigma)$ is continuous and bounded. The integral equation for each
$h_a$ and Gronwall's inequality give
\[
 \norm{h_a(\sigma)}\le\norm{u_a}
 \exp\!\left(\int_0^\sigma\norm{A(s)}_\op\dd s\right).
\]
This proves the asserted existence of all row equations.

We next quantify their relation to the actual matrix. Use two independent
real parameters $a,z$ and let $Y^{a,z}$ solve
\[
 \frac{\dd Y^{a,z}}{\dd\sigma}
 =P_K\!\left(
 \frac{2Y^{a,z}[C-z\Lop((Y^{a,z})^\top Y^{a,z})]}
 {2\norm{Y^{a,z}[C-z\Lop((Y^{a,z})^\top Y^{a,z})]}_F+c_\eta}
 \right),\qquad
 Y^{a,z}(0)=\diag(1,a,\ldots,a)U.
\]
At $(a,z)=(0,0)$ this solution is $e_1w(\sigma)^\top$.
For fixed $S$, the matrices $e_1w(\sigma)^\top C$, $0\le\sigma\le S$,
form a compact set avoiding zero. There is consequently a tube about
this solution, and a neighborhood of $(a,z)=(0,0)$, on which the matrix
inside the norm also avoids zero. The vector field is $C^\infty$ there:
it is a matrix polynomial composed with linear operations and the
smooth norm away from zero, and its denominator is at least $c_\eta$.
Finite-time dependence in \citet[Theorems~2.2, 2.9, 2.11, 2.13 and Corollary~2.15]{teschl2012ordinary}, on a
finite cover of $[0,S]$, supplies one parameter neighborhood on which
$Y^{a,z}$ and its parameter derivatives through order three are uniformly
bounded. The parameter derivatives are taken on this nonzero-gradient tube.

The transformation $Y\mapsto\diag(1,-1,\ldots,-1)Y$ preserves $Y^\top Y$ and commutes with $P_K$ on the left. Uniqueness and the initial values give
$Y^{-a,z}=\diag(1,-1,\ldots,-1)Y^{a,z}$. The first
row is therefore even in $a$ and every other row is odd in $a$.
At $a=0$ all weak rows vanish for every small $z$, because a zero row
of $Y$ stays zero in its right-multiplier equation. Taylor's formula
on the common compact parameter neighborhood yields, uniformly on
$[0,S]$,
\begin{align*}
 (Y^{a,z})_{1,:}&=w(\sigma)^\top+O_{U,S}(a^2+|z|),\\
 (Y^{a,z})_{a_0,:}&=
 a\left[\partial_a(Y^{a,0})_{a_0,:}\big|_{a=0}
       +O_{U,S}(a^2+|z|)\right],\qquad 2\le a_0\le m.
\end{align*}
For each weak-row entry, oddness eliminates its constant and quadratic
Taylor terms in $a$, leaving its derivative at $a=0$ and an $O(a^3)$
remainder. The mixed parameter derivative bounds the change of that
first derivative between $z$ and $0$ by $O(|z|)$.
The bounds are uniform over the finitely many row entries.

It remains to identify the derivative in this formula. At
$Y=e_1w^\top$ a perturbation supported on the weak rows has
\[
 \delta(Y^\top Y)=(\delta Y)^\top e_1w^\top+w e_1^\top\delta Y=0,
 \qquad \delta\norm{YC}_F^2=\tr(C\delta(Y^\top Y)C)=0.
\]
At $z=0$ the vector field is
\[
 Y\frac{2C}{2\norm{YC}_F+c_\eta}
 q_K\!\left(\frac{4CY^\top YC}{(2\norm{YC}_F+c_\eta)^2}\right).
\]
The displayed first variations make the derivative of this entire right
coefficient zero in a weak-row perturbation. Its coefficient at the
base solution is exactly $A(\sigma)$. Thus the $a$ derivative of weak
row $a_0$ solves $\dd h_{a_0}^\top/\dd\sigma=h_{a_0}^\top A$,
with $h_{a_0}(0)=u_{a_0}$. Uniqueness identifies it with
\eqref{eq:row-equations}. Substituting $a=\eps^{b-1}$ and $z=\eps^2$
gives both bounds in \eqref{eq:row-expansion}. All parameter neighborhoods
and all constants were chosen after fixing $S$.
\end{proof}

Let $V_\perp=(v_2,\ldots,v_r)$ and define
\begin{equation}\label{eq:zeta-definition}
 \zeta_a(\sigma)=h_a(\sigma)^\top V_\perp
 -\frac{h_a(\sigma)^\top v_1}{w(\sigma)^\top v_1}
 w(\sigma)^\top V_\perp,
\end{equation}
\begin{equation}\label{eq:B-initial}
 B(\sigma)=V_\perp^\top A(\sigma)V_\perp
 -\frac{V_\perp^\top A(\sigma)v_1\,w(\sigma)^\top V_\perp}
 {w(\sigma)^\top v_1}.
\end{equation}
\begin{lemma}[Projected weak-row transport]\label{main:projected}
For the data of Lemma~\ref{main:rows},
\begin{equation}\label{eq:initial-normal-transport}
 \frac{\dd\zeta_a}{\dd\sigma}=\zeta_a B(\sigma).
\end{equation}
The matrix solution
\begin{equation}\label{eq:Psi-initial}
 \Psi'(\sigma)=\Psi(\sigma)B(\sigma),\qquad\Psi(0)=I_{r-1}
\end{equation}
is invertible at every finite $\sigma$, and
\begin{equation}\label{eq:cutoff-transport}
 \zeta_a(\sigma)=\zeta_a(0)\Psi(\sigma),\qquad
 \zeta_a(0)=u_a^\top V_\perp-
 \frac{u_a^\top v_1}{u^\top v_1}u^\top V_\perp.
\end{equation}
Conditional on $u$, at every deterministic finite $\sigma$ these rows
are independent centered Gaussian rows with covariance
\begin{equation}\label{eq:normal-covariance}
 \operatorname{Cov}(\zeta_a(\sigma)^\top\mid u)
 =\Psi(\sigma)^\top\left[I_{r-1}+
 \frac{(V_\perp^\top u)(V_\perp^\top u)^\top}{(u^\top v_1)^2}\right]
 \Psi(\sigma)\succ0.
\end{equation}
For every fixed finite $S$, the following expansion is uniform on $[0,S]$:
\begin{equation}\label{eq:actual-row-elimination}
 \begin{aligned}
 &y_a^\eps(\sigma)^\top V_\perp
 -\frac{y_a^\eps(\sigma)^\top v_1}{y_1^\eps(\sigma)^\top v_1}
 y_1^\eps(\sigma)^\top V_\perp\\
 &\hspace{15mm}=\eps^{b-1}
 \bigl[\zeta_a(\sigma)+O_{U,S}(\eps^{2(b-1)}+\eps^2)\bigr].
 \end{aligned}
\end{equation}
The denominator on the left is nonzero throughout $[0,S]$ for every
sufficiently small $\eps>0$.
\end{lemma}
\begin{proof}(of Lemma~\ref{main:projected}).
\label{proof:main:projected}
The denominator $w(\sigma)^\top v_1$ is nonzero by
Lemma~\ref{main:rows}. The row
\[
 h_a(\sigma)^\top-
 \frac{h_a(\sigma)^\top v_1}{w(\sigma)^\top v_1}w(\sigma)^\top
\]
annihilates $v_1$, so it equals $\zeta_a(\sigma)V_\perp^\top$.
The two row equations consequently give the exact quotient derivative
\begin{align*}
 \frac{\dd}{\dd\sigma}
 \frac{h_a(\sigma)^\top v_1}{w(\sigma)^\top v_1}
 &=\frac{
 [h_a(\sigma)^\top A(\sigma)v_1][w(\sigma)^\top v_1]
 -[h_a(\sigma)^\top v_1][w(\sigma)^\top A(\sigma)v_1]}
 {[w(\sigma)^\top v_1]^2}\\
 &=\frac{\zeta_a(\sigma)V_\perp^\top A(\sigma)v_1}
 {w(\sigma)^\top v_1}.
\end{align*}
Differentiating \eqref{eq:zeta-definition}, grouping the two terms
that contain a differentiated row, and then using this quotient
identity gives
\begin{align*}
 \frac{\dd\zeta_a}{\dd\sigma}
 &=\left[h_a(\sigma)^\top-
 \frac{h_a(\sigma)^\top v_1}{w(\sigma)^\top v_1}w(\sigma)^\top\right]
 A(\sigma)V_\perp\\
 &\quad-
 \frac{\zeta_a(\sigma)V_\perp^\top A(\sigma)v_1}
 {w(\sigma)^\top v_1}\,w(\sigma)^\top V_\perp\\
 &=\zeta_a(\sigma)B(\sigma).
\end{align*}
The quotient derivative accounts for the motion of the occupied row.

The coefficient $B$ is continuous on finite intervals. Let $\Psi$ solve \eqref{eq:Psi-initial}. A row solution that vanishes at a finite time vanishes initially by backward uniqueness. Since $\Psi(0)=I$, its rows remain independent, and $\Psi$ is invertible. The row $\zeta_a(0)\Psi$ solves the same equation and has the same initial value as $\zeta_a$, proving \eqref{eq:cutoff-transport}.

Conditional on $u$, the Gaussian coordinates $u_a^\top v_1$ and
$u_a^\top V_\perp$ are independent, with variances $1$ and $I_{r-1}$.
They have zero means, so the covariance of the initial projected row is
\[
 I_{r-1}+\frac{(V_\perp^\top u)(V_\perp^\top u)^\top}{(u^\top v_1)^2}.
\]
Its quadratic form at nonzero $x$ is $\norm x^2+(u^\top V_\perp x)^2/(u^\top v_1)^2>0$. The coefficients $A,B,\Psi$ depend only on $u$ and the fixed model. Conditional covariance transport at deterministic $\sigma$ gives
\[
 \Psi(\sigma)^\top\left[I_{r-1}+\frac{(V_\perp^\top u)(V_\perp^\top u)^\top}{(u^\top v_1)^2}\right]\Psi(\sigma)\succ0.
\]
Conditional independence of the weak rows is preserved by these separate deterministic linear transformations.

Finally fix $S$. Continuity and nonvanishing give
$\inf_{0\le\sigma\le S}|w(\sigma)^\top v_1|>0$. Lemma~\ref{main:rows} gives,
uniformly there,
\begin{align*}
 \frac1{y_1^\eps(\sigma)^\top v_1}
 &=\frac1{w(\sigma)^\top v_1}+O_{U,S}(\eps^{2(b-1)}+\eps^2),\\
 \frac{y_a^\eps(\sigma)^\top v_1}{y_1^\eps(\sigma)^\top v_1}
 &=\eps^{b-1}\left[
 \frac{h_a(\sigma)^\top v_1}{w(\sigma)^\top v_1}
 +O_{U,S}(\eps^{2(b-1)}+\eps^2)\right].
\end{align*}
The first equality follows by subtracting the reciprocals and bounding
both denominators below; all row factors in the second are bounded on
$[0,S]$. Multiplication by
$y_1^\eps(\sigma)^\top V_\perp=w(\sigma)^\top V_\perp+O_{U,S}(\eps^{2(b-1)}+\eps^2)$
and subtraction from the weak-row expansion prove the stated actual
projection estimate.
\end{proof}

Put $\kappa=(\lambda_1-\lambda_2)/\lambda_1\in(0,1)$.
\begin{lemma}[Leading-row asymptotic scale]\label{lem:main-initial-approach}
Under Assumption~\ref{ass:task}, for every fixed $u$ with
$u^\top v_1\ne0$,
\begin{equation}\label{eq:initial-direction}
 \frac{w(\sigma)^\top}{\sigma}
 =\sign(u^\top v_1)v_1^\top+O_u(\sigma^{-\kappa})
 \qquad(\sigma\to\infty).
\end{equation}
Consequently, for every prescribed neighborhood of
$\{Oe_1v_1^\top:O\in\Ogroup(m)\}$ and every prescribed positive
upper bound on $1/\sigma_0$, there is a sufficiently large finite
$\sigma_0(U)$ such that $Q^\eps(\eps\sigma_0)$ belongs to this
neighborhood for all sufficiently small $\eps$. This choice is fixed
when $\eps\downarrow0$.
\end{lemma}
\begin{proof}(of Lemma~\ref{lem:main-initial-approach}).
\label{proof:lem:main-initial-approach}
Use the scalar time $\vartheta$ from the proof of Lemma~\ref{main:rows}. Spectral decomposition gives
\begin{align*}
 \norm{u^\top e^{\vartheta C}C}^2
 &=\lambda_1^2(u^\top v_1)^2e^{2\lambda_1\vartheta}
 +\sum_{l=2}^r\lambda_l^2(u^\top v_l)^2e^{2\lambda_l\vartheta},\\
 \norm{u^\top e^{\vartheta C}C}
 &=\lambda_1|u^\top v_1|e^{\lambda_1\vartheta}
 \bigl[1+O_u(e^{-2(\lambda_1-\lambda_2)\vartheta})\bigr].
\end{align*}
The second equality follows by dividing by the first summand and using $|\sqrt{1+x}-1|\le x$ for $x\ge0$. Define
\[
 s=\frac{2\norm{u^\top e^{\vartheta C}C}}
 {2\norm{u^\top e^{\vartheta C}C}+c_\eta}.
\]
As $\vartheta\to\infty$, $1-s=O_u(e^{-\lambda_1\vartheta})$. Since $p_K(1)=1$, polynomial smoothness gives $p_K(s)=1+O_{K,u}(e^{-\lambda_1\vartheta})$ and the same estimate for its reciprocal. Thus
\[
 \frac{\dd\sigma}{\dd\vartheta}
 =\frac{\norm{u^\top e^{\vartheta C}C}}{p_K(s)}
 =\lambda_1|u^\top v_1|e^{\lambda_1\vartheta}
 +O_u(e^{(2\lambda_2-\lambda_1)\vartheta})+O_u(1).
\]
This derivative is positive and continuous at finite $\vartheta$, and its integral diverges. Because $0<\lambda_2<\lambda_1$, both error terms are $O_u(e^{\lambda_2\vartheta})$. Integration gives
\[
 \sigma=|u^\top v_1|e^{\lambda_1\vartheta}+O_u(e^{\lambda_2\vartheta}).
\]
In particular $\sigma$ is comparable to $e^{\lambda_1\vartheta}$. Spectral decomposition also gives
\[
 u^\top e^{\vartheta C}
 =(u^\top v_1)e^{\lambda_1\vartheta}v_1^\top
 +O_u(e^{\lambda_2\vartheta}).
\]
Subtracting $\sign(u^\top v_1)\sigma v_1^\top$ and dividing by $\sigma$ gives error
$O_u(e^{-(\lambda_1-\lambda_2)\vartheta})=O_u(\sigma^{-\kappa})$.
For a prescribed neighborhood choose a sufficiently large finite $\sigma_0$. At this fixed time, Lemma~\ref{main:rows} gives
$Q^\eps(\eps\sigma_0)\to e_1w(\sigma_0)^\top/\sigma_0$. The same choice makes $1/\sigma_0$ arbitrarily small, proving the last assertion.
\end{proof}

\begin{lemma}[Asymptotic independent growth coefficients]\label{main:coefficients}
Under Assumption~\ref{ass:task}, for every fixed $u$ with
$u^\top v_1\ne0$,
\begin{equation}\label{eq:sigma-B}
 \sigma B(\sigma)=\frac{a_K}{\lambda_1}
 \diag(\lambda_2,\ldots,\lambda_r)+O_u(\sigma^{-\kappa}).
\end{equation}
In the logarithmic initial-layer time $\mu$ of \eqref{eq:mu-time},
Then
\begin{equation}\label{eq:linear-normal}
 \begin{aligned}
 \frac{\dd}{\dd\mu}[e^{-\mu}\zeta_a(e^\mu)]
 &=e^{-\mu}\zeta_a(e^\mu)
 [\diag_{2\le l\le r}(g_l^{(1)})+E_u(\mu)],\\
 g_l^{(1)}&=\frac{a_K\lambda_l}{\lambda_1}-1.
 \end{aligned}
\end{equation}
where $E_u$ depends only on $u$ and the fixed model parameters, and
$\norm{E_u(\mu)}_\op=O_u(e^{-\kappa\mu})$ for all sufficiently
large $\mu$.
\end{lemma}
\begin{proof}(of Lemma~\ref{main:coefficients}).
\label{proof:main:coefficients}
Divide the numerator and denominator in \eqref{eq:A-initial} by the
appropriate powers of $\sigma$ to obtain the exact expression
\[
 \sigma A(\sigma)=
 \frac{2C}{2\norm{w(\sigma)^\top C/\sigma}+c_\eta/\sigma}
 q_K\!\left(
 \frac{4C[w(\sigma)/\sigma][w(\sigma)/\sigma]^\top C}
 {(2\norm{w(\sigma)^\top C/\sigma}+c_\eta/\sigma)^2}\right).
\]
By Lemma~\ref{lem:main-initial-approach}, its row argument differs from
$\sign(u^\top v_1)v_1$ by $O_u(\sigma^{-\kappa})$ and its scalar
denominator differs from $2\lambda_1$ by the same order. Here
$\sigma^{-1}=O(\sigma^{-\kappa})$ because $0<\kappa<1$.
All denominators stay bounded below. Finite matrix polynomial
multiplication and the identity
$X^j-Y^j=\sum_{i=0}^{j-1}X^i(X-Y)Y^{j-1-i}$ give a uniform Lipschitz
bound for $q_K$ on the resulting bounded set. Hence
\[
 \sigma A(\sigma)=\frac C{\lambda_1}
 q_K(v_1v_1^\top)+O_u(\sigma^{-\kappa})
 =\frac C{\lambda_1}
 [a_KI+(1-a_K)v_1v_1^\top]+O_u(\sigma^{-\kappa}).
\]
In the last equality a projection has eigenvalues zero and one,
$q_K(0)=a_K$, and $q_K(1)=1$.
Projection onto $V_\perp$ gives the diagonal term in the assertion.
Projection on the left onto $V_\perp$ and on the right onto $v_1$
gives $\sigma V_\perp^\top A(\sigma)v_1=O_u(\sigma^{-\kappa})$.
Moreover the leading-row estimate and its nonzero leading coordinate give
\[
 \frac{w(\sigma)^\top V_\perp}{w(\sigma)^\top v_1}
   =O_u(\sigma^{-\kappa}).
\]
Thus the subtracted term in $\sigma B$ has size
$O_u(\sigma^{-2\kappa})$, and the claimed bound on $\sigma B$ follows.
For $\mu=\log\sigma$, the chain rule gives exactly
\[
 \frac{\dd}{\dd\mu}\left(e^{-\mu}\zeta_a(e^\mu)\right)
 =e^{-\mu}\zeta_a(e^\mu)[e^\mu B(e^\mu)-I].
\]
Substitution of the coefficient estimate proves the normalized equation
with the diagonal entries $a_K\lambda_l/\lambda_1-1$ and the error
$O_u(e^{-\kappa\mu})$.
\end{proof}

The coefficient error is integrable on every sufficiently late
half-line in $\mu$. Section~\ref{sec:gaussian-data} uses this fact to
construct the Gaussian coefficients that control independent matrix
volumes. The fixed-interval expansion and the large-$\sigma$ estimates
are used in that order.

\subsection{Asymptotic coefficients, minors, and first entry}\label{sec:gaussian-data}
The integrable error in the normalized independent-row equation permits
a measurable asymptotic basis. Its Gaussian coefficients are fixed
before choosing the seed-dependent matching time. Exterior products then
identify all retained independent scales at that time.
\begin{lemma}[Asymptotic basis of the independent-row equation]\label{lem:asymptotic-basis}
Under Assumption~\ref{ass:task}, fix $u^\top v_1\ne0$.
There is an invertible matrix function
$F_u(\mu)\in\R^{(r-1)\times(r-1)}$, whose rows solve
\eqref{eq:linear-normal}, such that
\begin{equation}\label{eq:asymptotic-basis}
 F_u(\mu)=\diag_{2\le l\le r}(e^{g_l^{(1)}\mu})[I_{r-1}+o_u(1)]
 \qquad(\mu\to\infty).
\end{equation}
It can be chosen measurably as a function of $u$ and continued invertibly
 to every finite real $\mu$. Define $\Xi\in\R^{(m-1)\times(r-1)}$ by
\begin{equation}\label{eq:gaussian-coefficients}
 \Xi_{a-1,:}=\zeta_a(1)F_u(0)^{-1},\qquad
 e^{-\mu}\zeta_a(e^\mu)=\Xi_{a-1,:}F_u(\mu),\quad 2\le a\le m.
\end{equation}
Conditional on $u$, the rows of $\Xi$ are independent centered Gaussians
with common positive definite covariance
\[
 F_u(0)^{-\top}\Psi(1)^\top
 \left[I_{r-1}+\frac{(V_\perp^\top u)(V_\perp^\top u)^\top}{(u^\top v_1)^2}\right]
 \Psi(1)F_u(0)^{-1}\succ0.
\]
\end{lemma}
\begin{proof}(of Lemma~\ref{lem:asymptotic-basis}).
\label{proof:lem:asymptotic-basis}
Choose $\mu_b$ as the first positive integer such that
\[
 \sqrt{r-1}\int_{\mu_b}^\infty\norm{E_u(s)}_\op\dd s<\tfrac12.
\]
For each task label $l$, map a bounded continuous row function $x$ on $[\mu_b,\infty)$ to the row function whose $k$th coordinate at $\mu$ is
\[
 \begin{cases}
 \displaystyle\int_{\mu_b}^\mu
 e^{(g_k^{(1)}-g_l^{(1)})(\mu-s)}
 [(e_l^\top+x(s))E_u(s)]_k\dd s,
 &g_k^{(1)}<g_l^{(1)},\\[2mm]
 \displaystyle-\int_\mu^\infty
 e^{(g_k^{(1)}-g_l^{(1)})(\mu-s)}
 [(e_l^\top+x(s))E_u(s)]_k\dd s,
 &g_k^{(1)}\ge g_l^{(1)}.
 \end{cases}
\]
Here $x$ is the generic input of the map. Every kernel is bounded by one on its integration range. In the supremum of the Euclidean row norm, combining the $r-1$ coordinate bounds gives Lipschitz constant at most the displayed tail integral multiplied by $\sqrt{r-1}$. The inhomogeneous term is bounded as well. The contraction principle \citep[Theorem~2.1]{teschl2012ordinary} gives a unique fixed point. Define row $l$ of $F_u(\mu)$ by adding $e_l^\top$ to this fixed point and multiplying by $e^{g_l^{(1)}\mu}$. Differentiating the convergent integrals verifies the row equation, including equal-rate coordinates.

The normalized correction
$e^{-g_l^{(1)}\mu}(F_u(\mu))_{l,:}-e_l^\top$ is bounded. For $g_k^{(1)}\ge g_l^{(1)}$, its $k$th coordinate is bounded in absolute value by
\[
 \left(1+\sup_{\nu\ge\mu_b}
 \norm{e^{-g_l^{(1)}\nu}(F_u(\nu))_{l,:}-e_l^\top}\right)
 \int_\mu^\infty\norm{E_u(s)}_\op\dd s,
\]
which tends to zero. For $g_k^{(1)}<g_l^{(1)}$, split the integral at $(\mu+\mu_b)/2$. The same prefactor multiplies
\[
 e^{(g_k^{(1)}-g_l^{(1)})(\mu-\mu_b)/2}
 \int_{\mu_b}^\infty\norm{E_u(s)}_\op\dd s
 +\int_{(\mu+\mu_b)/2}^\infty\norm{E_u(s)}_\op\dd s.
\]
Both terms tend to zero. The normalized rows of $F_u$ therefore tend to the standard basis, proving invertibility at late times. Continue by
$F_u'=F_u[\diag_{2\le l\le r}(g_l^{(1)})+E_u]$.
Finite-interval row uniqueness as in Lemma~\ref{main:projected} preserves invertibility at every finite real time.

The tail integral is a measurable limit of finite integrals in $u$, so the first-integer choice of $\mu_b$ is measurable. Contraction iteration from zero, its uniform limit, and finite-time linear continuation give a measurable choice of $F_u$.

The time $\mu=0$ corresponds to $\sigma=1$. The rows
$e^{-\mu}\zeta_a(e^\mu)$ and $\zeta_a(1)F_u(0)^{-1}F_u(\mu)$ solve the same linear equation and agree there. Uniqueness gives
\[
 e^{-\mu}\zeta_a(e^\mu)=\Xi_{a-1,:}F_u(\mu),
 \qquad \Xi_{a-1,:}=\zeta_a(1)F_u(0)^{-1}.
\]
By Lemma~\ref{main:projected}, $\zeta_a(1)=\zeta_a(0)\Psi(1)$. Conditional on $u$, these rows are independent centered Gaussians, and their common covariance after multiplication is
\[
 F_u(0)^{-\top}\Psi(1)^\top
 \left[I_{r-1}+\frac{(V_\perp^\top u)(V_\perp^\top u)^\top}{(u^\top v_1)^2}\right]
 \Psi(1)F_u(0)^{-1}\succ0.
\]
The bracket has strictly positive quadratic form on nonzero vectors, and both multiplying matrices are invertible and measurable in $u$. This proves all coefficient assertions.
\end{proof}

For a matrix $Z=(z_2,\ldots,z_r)\in\R^{(m-1)\times(r-1)}$, set
\[
 V_h(Z)=\norm{z_2\wedge\cdots\wedge z_{h+1}},\qquad V_0(Z)=1.
\]
The induced matrix $\wedge^h Z$ has as columns all $h$-fold exterior
products of distinct columns of $Z$; its entries are the corresponding
minors in orthonormal bases. Its Frobenius norm is independent of those
bases. The matrix with rows $\zeta_a(\sigma)/\sigma$, $2\le a\le m$,
is $\Xi F_u(\log\sigma)$.
For an $h$-element column set $J$, write $(\wedge^h\Xi)_J$ for that
column of the exterior matrix.

\begin{lemma}[Nonzero Gaussian exterior coefficients]\label{lem:gaussian-minors}
Under Assumption~\ref{ass:task}, there is a measurable Gaussian-full
event on which $u^\top v_1\ne0$ and every column $(\wedge^h\Xi)_J$ is
nonzero, for $1\le h\le\min\{m-1,r-1\}$ and all $h$-element sets
$J\subset\{2,\ldots,r\}$.
Under the additional Assumption~\ref{ass:prefix}, for each seed in this
event and every $1\le h\le p-1$, put $J_h=\{2,\ldots,h+1\}$. Then
\begin{equation}\label{eq:exterior-asymptotic}
 e^{-\mu\sum_{l=2}^{h+1}g_l^{(1)}}\wedge^h [\Xi F_u(\mu)]
 \longrightarrow (\wedge^h\Xi)_{J_h}\,e_{J_h}^\top.
\end{equation}
Here $e_{J_h}$ is the corresponding standard exterior basis vector.
Consequently $V_h(\Xi F_u(\log\sigma))>0$ for all sufficiently large $\sigma$,
and
\[
 \frac{\norm{\wedge^h \Xi F_u(\log\sigma)}_F^2}{V_h(\Xi F_u(\log\sigma))^2}
 \longrightarrow1.
\]
For $h=r-1$ there is only one column, and the ratio is identically one
where it is defined.
\end{lemma}
\begin{proof}(of Lemma~\ref{lem:gaussian-minors}).
\label{proof:lem:gaussian-minors}
Conditional on $u$, the rows of $\Xi$ have a nonsingular joint Gaussian density. For each $h$-element column set $J$, vanishing of $(\wedge^h\Xi)_J$ implies that the minor in the first $h$ rows and columns $J$ vanishes. This is a nonzero polynomial: it equals one when that selected submatrix is the identity. The zero-set theorem of \citet{mityagin2020zero} makes its zero set Lebesgue null. There are finitely many relevant pairs $(h,J)$. Intersecting their full-probability events and integrating conditional probabilities, also excluding $u^\top v_1=0$, gives a measurable Gaussian-full event.

Assume the separated-prefix hypothesis and fix $1\le h\le p-1$. The asymptotic basis formula and the exterior identity for a matrix product give
\[
 \wedge^h[\Xi F_u(\mu)]
 =(\wedge^h\Xi)\diag_{|J|=h}
 \left(e^{\mu\sum_{l\in J}g_l^{(1)}}\right)[I+o_u(1)].
\]
The last factor tends to the identity because fixed exterior powers are polynomial in the entries. The set $J_h=\{2,\ldots,h+1\}$ has the unique largest summed exponent. When $h<r-1$, every competitor removes an exponent at least $g_{h+1}^{(1)}$ and adds one at most $g_{h+2}^{(1)}$, giving positive gap at least $g_{h+1}^{(1)}-g_{h+2}^{(1)}$. Consequently
\[
 e^{-\mu\sum_{l=2}^{h+1}g_l^{(1)}}\wedge^h[\Xi F_u(\mu)]
 \longrightarrow (\wedge^h\Xi)_{J_h}e_{J_h}^\top.
\]
This is \eqref{eq:exterior-asymptotic}. Its leading column is nonzero on the established event, while every competing column is negligible relative to it. Thus the leading volume is eventually positive and the squared full-exterior to leading-volume ratio tends to one. For $h=r-1$ there is a single exterior column; its normalized limit is nonzero and the ratio is identically one whenever defined. Repeated exponents outside the tested prefix preserve these conclusions.
\end{proof}

For a matrix $Q$ whose first task column $Qv_1$ is nonzero, project the
remaining task columns onto $\Ran(Qv_1)^\perp$ and express the projections
in orthonormal hidden coordinates. Denote the resulting matrix by
$Z(Q)$. The norms and volumes above are independent of this coordinate
choice. On the actual path write $Z^\eps(t)=Z(Q^\eps(t))$.

\begin{lemma}[Gaussian independent scales at first entry]\label{lem:gaussian-entry}
Under Assumption~\ref{ass:task}, for every fixed seed with
$u^\top v_1\ne0$ and every fixed finite
$\sigma>0$,
\begin{equation}\label{eq:gaussian-normal-entry}
 \norm{Z^\eps(\eps\sigma)}_F=O_U(\eps^{b-1}).
\end{equation}
Under Assumption~\ref{ass:prefix}, on the Gaussian-full event of
Lemma~\ref{lem:gaussian-minors}, for every
$\zeta>0$ there is a sufficiently large finite $\sigma_0(U)$ such that,
for all sufficiently small $\eps>0$ and every $1\le h\le p-1$,
\begin{align}
 \log V_h(Z^\eps(\eps\sigma_0))
 &=h(b-1)\log\eps+O_U(1),\label{eq:gaussian-volume-entry}\\
 \norm{\wedge^h Z^\eps(\eps\sigma_0)}_F^2
 &\le(1+\zeta^2)V_h(Z^\eps(\eps\sigma_0))^2.
 \label{eq:gaussian-cone-entry}
\end{align}
The same $\sigma_0(U)$ places $Q^\eps(\eps\sigma_0)$ in any prescribed
sufficiently small neighborhood of
$\{Oe_1v_1^\top:O\in\Ogroup(m)\}$ and makes $1/\sigma_0$ as small as
prescribed. It is chosen after the seed and held fixed as $\eps\downarrow0$.
\end{lemma}
\begin{proof}(of Lemma~\ref{lem:gaussian-entry}).
\label{proof:lem:gaussian-entry}
Use ordinary entries and submatrices in task coordinates. When $Q_{11}\ne0$, the columns of
\[
 \begin{pmatrix}-Q_{2:m,1}^\top/Q_{11}\\ I_{m-1}\end{pmatrix}
 \left(I_{m-1}+\frac{Q_{2:m,1}Q_{2:m,1}^\top}{Q_{11}^2}\right)^{-1/2}
\]
form an orthonormal frame perpendicular to the first column of $Q$. Multiplication by that column gives zero; the Gram matrix before the normalization is the matrix inside the inverse square root. In this frame the projected matrix is
\[
 Z(Q)=\left(I_{m-1}+\frac{Q_{2:m,1}Q_{2:m,1}^\top}{Q_{11}^2}\right)^{-1/2}
 \left(Q_{2:m,2:r}-\frac{Q_{2:m,1}Q_{1,2:r}}{Q_{11}}\right).
\]
For $Q=Y^\eps(\sigma)/\sigma$ at fixed $\sigma>0$, the denominator has a nonzero limit and
$Q_{2:m,1}/Q_{11}=O_U(\eps^{b-1})$ by Lemma~\ref{main:rows}. Spectral calculus bounds the normalization's difference from the identity by $O_U(\eps^{2(b-1)})$. Lemma~\ref{main:projected} applied to the second factor, together with the factor $1/\sigma$, therefore gives
\[
 Z^\eps(\eps\sigma)=\eps^{b-1}
 \left[\Xi F_u(\log\sigma)+O_{U,\sigma}(\eps^{2(b-1)}+\eps^2)\right].
\]
This also proves the fixed-time norm estimate.

Fix a seed in the Gaussian-full event of Lemma~\ref{lem:gaussian-minors}. There are finitely many retained orders. Choose a sufficiently large finite $\sigma_0$ so that their leading volumes of $\Xi F_u(\log\sigma_0)$ are nonzero and their squared exterior ratios lie strictly below $1+\zeta^2$. Lemma~\ref{lem:main-initial-approach} allows the same choice to meet the required rank-one neighborhood and upper bound on $1/\sigma_0$. Keep this time fixed while $\eps\downarrow0$. Multilinearity yields
\[
 \eps^{-h(b-1)}\wedge^h Z^\eps(\eps\sigma_0)
 \longrightarrow\wedge^h[\Xi F_u(\log\sigma_0)].
\]
The leading-column norm on the right is positive and finite. Its normalized counterpart is bounded above and below by positive seed-dependent constants, giving
\[
 \log V_h(Z^\eps(\eps\sigma_0))=h(b-1)\log\eps+O_U(1).
\]
The strict ratio margins pass to the actual matrix for sufficiently small $\eps$, and the strong-row expansion gives the required shape neighborhood. All time choices are made after fixing a seed in the previously established coefficient event.
\end{proof}

The conclusions concern finitely many nonzero volumes at a fixed
initial-layer time. The next sections prove their relative propagation
over the much longer actual time intervals on which independent
components reach a fixed size.

\subsection{Shapes, full linearization, and connections}\label{sec:local-structure}
The local calculations below apply to a symmetric nonsingular task
matrix in arbitrary positive factor dimensions. They establish the
shape equations used in the main recursion, the contracting tangent
blocks, and a connecting solution between successive leading ranks.
\begin{lemma}[Gram matrices at zeros of $F_K$]
\label{lem:gram-zeros}
Every nonzero $Q$ satisfying $F_K(Q)=0$ obeys
\[
 C\Ran(Q^\top)=\Ran(Q^\top),\qquad
 z^\top Cz>0\quad\text{for }z\in\Ran(Q^\top)\setminus\{0\}.
\]
For $K\ge1$, the subspace $\Ran(Q^\top)$ determines $Q^\top Q$
uniquely. In particular, for $Q=Q_{I,K}$, there are positive
coefficients $q_l=\norm{Q_{I,K}v_l}$, $l\in I$, satisfying the equations \eqref{eq:zero-system}.
Conversely, for $K\ge1$, $|I|\le m$, and $\lambda_l>0$ on $I$, every positive
solution of these equations constructs such a matrix. The coefficients
$q_l$ and $s_{I,K}$ are unique when they exist.

At rank one, $Q=uv^\top$ for unit vectors $u,v$ with $Cv=\lambda v$
and $\lambda>0$, and $\norm{QC}_F=\lambda$.
At $K=0$, the zeros are exactly the matrices of Frobenius norm one
whose row space lies in a single positive eigenspace of $C$.
In particular $Q_{1,K}=e_1v_1^\top$ may be chosen for every $K$, and
$s_{1,K}=\lambda_1$.
\end{lemma}
\begin{proof}(of Lemma~\ref{lem:gram-zeros}).
\label{proof:lem:gram-zeros}
Since $Q\ne0$ and $C$ is nonsingular, $\norm{QC}_F>0$. Take a compact singular value decomposition
\[
 \frac{QC}{\norm{QC}_F}=\sum_{a=1}^d\sigma_a u_a z_a^\top,
 \qquad \sigma_a>0,\qquad\sum_{a=1}^d\sigma_a^2=1.
\]
The equation $F_K(Q)=0$ and polynomial action on singular values give
$Q=\sum_a p_K(\sigma_a)u_az_a^\top$.
Multiplying by $C$ and comparing the coefficients of $u_a$ yields
\[
 p_K(\sigma_a)z_a^\top C=\norm{QC}_F\sigma_a z_a^\top,
 \qquad Cz_a=\frac{\norm{QC}_F\sigma_a}{p_K(\sigma_a)}z_a.
\]
The displayed eigenvalue is positive. These vectors span
$\Ran(Q^\top)$, proving both the equality of subspaces and the positive
quadratic form in the statement. This conclusion is derived from the
full matrix equation.

For $K\ge1$, Lemma~\ref{lem:update-map} says that $q_K$ is strictly
decreasing from $a_K$ to $1$ on $[0,1]$. In any fixed orthonormal
eigenbasis of the indicated row subspace, an eigenvalue $\lambda$
therefore has its normalized squared singular value determined by
\[
 \sigma^2=q_K^{-1}(\norm{QC}_F/\lambda).
\]
Each such value strictly decreases as $\norm{QC}_F$ increases on its admissible
domain. Their sum must be one, so at most one value of $\norm{QC}_F$ is possible.
The squared singular values, and then the Gram matrix, are unique.
Equal active eigenvalues have equal normalized singular values.
Within a repeated eigenspace the resulting Gram matrix is a scalar
multiple of the orthogonal projector onto the prescribed active
subspace, so changing its orthonormal basis leaves that Gram matrix
unchanged.

For the prescribed row space indexed by $I$, the right singular vectors
can consequently be taken to be the fixed $v_l$. Set
$q_l=p_K(\sigma_l)=\norm{Q_{I,K}v_l}$. Comparing $Q_{I,K}C$ with its
normalized singular value decomposition gives
\[
 \lambda_lq_l=s_{I,K}\sigma_l,\qquad
 \sum_{l\in I}\sigma_l^2=1.
\]
Substitution proves all three equations in \eqref{eq:zero-system}.
For later scalar calculations, they imply
\[
 \sum_{l\in I}q_K^{-1}(s_{I,K}/\lambda_l)=1,
 \qquad
 \max_{l\in I}\lambda_l\le s_{I,K}<a_K\min_{l\in I}\lambda_l
\]
when all normalized components are positive. At rank one $s_{I,K}$
is the corresponding positive eigenvalue.

Conversely, given a positive solution of \eqref{eq:zero-system},
choose orthonormal $u_l\in\R^m$ and put
$Q=\sum_{l\in I}q_lu_lv_l^\top$.
Its normalization has singular values
$\lambda_lq_l/s_{I,K}>0$, whose squares sum to one. Applying $P_K$
returns the coefficients $q_l$ by the second equation, so $F_K(Q)=0$
and its row space is the prescribed subspace.

At rank one the normalized singular value is one. Since $p_K(1)=1$,
the nonzero singular value of $Q$ is one and its scale is $\lambda>0$.
At $K=0$, $QC/\norm{QC}_F=Q$ means $QC=sQ$; its rows lie in the eigenspace for
$\norm{QC}_F>0$. The identity $\norm{QC}_F=\norm{QC}_F=\norm{QC}_F\norm Q_F$ gives $\norm Q_F=1$.
Conversely every such matrix satisfies $F_0(Q)=0$. The simple largest
eigenvalue in Assumption~\ref{ass:task} then gives the asserted leading
rank-one choice.
\end{proof}

\begin{lemma}[Existence after adding an input direction]
\label{lem:add-zero}
Suppose $K\ge1$, $Q_{I,K}$ exists, $m>|I|$, and
$0<\lambda_l<\min_{a\in I}\lambda_a$ for $l\notin I$.
A zero of $F_K$ with row space
$\operatorname{span}\{v_a:a\in I\cup\{l\}\}$ exists if and only if
\[
 a_K\lambda_l>s_{I,K}.
\]
For $K=0$, the same equivalence holds under the strict inequality
$\lambda_l<\min_{a\in I}\lambda_a$.
Thus, when $\lambda_d>\lambda_{d+1}>0$ and $d<m$,
existence of $Q_{d+1,K}$ is equivalent to
$a_K\lambda_{d+1}/s_{d,K}-1>0$.
\end{lemma}
\begin{proof}(of Lemma~\ref{lem:add-zero}).
\label{proof:lem:add-zero}
First suppose $K\ge1$. Equations~\eqref{eq:zero-system} and the
inverse in Lemma~\ref{lem:update-map} give
$\sum_{a\in I}q_K^{-1}(s_{I,K}/\lambda_a)=1$.
If $a_K\lambda_l>s_{I,K}$, then at this old scale, the old components sum to one and the new component is strictly positive. At the upper endpoint $a_K\lambda_l$, the new component is zero and the old sum is strictly smaller than one. Both values lie in the domain of the old inverse components because $\lambda_l$ is at most every old eigenvalue. The intermediate value theorem gives a root between them with all components positive. Conversely, at any new root the old components sum to less than one. Strict decrease and positivity of the new component place that root strictly between $s_{I,K}$ and $a_K\lambda_l$, proving necessity.

For $K=0$, Lemma~\ref{lem:gram-zeros} says that all eigenvalues
used by a zero of $F_K$ coincide. Since $\lambda_l$ is strictly
below every active eigenvalue, the enlarged matrix does
not exist. Also $a_0\lambda_l=\lambda_l<s_{I,0}$, so both sides
of the asserted equivalence are false. Taking
$I=\{1,\ldots,d\}$ gives the leading-prefix version.
\end{proof}

\begin{lemma}[Full derivative at a zero of $F_K$]
\label{lem:spectrum}
Let $Q_{I,K}$ have rank $d$. For any unused task eigenvector
$v_l$, and any unit $u_\perp$ orthogonal to the hidden range of
$Q_{I,K}$,
\[
 DF_K(Q_{I,K})[u_\perp v_l^\top]
 =\left(\frac{a_K\lambda_l}{s_{I,K}}-1\right)u_\perp v_l^\top.
\]
For $K\ge1$, the derivative on
$\operatorname{span}\{Q_{I,K}v_av_a^\top:a\in I\}$ has
strictly negative eigenvalues.
If $K\ge1$ and every active eigenvalue is strictly larger than
every unused eigenvalue, the complete $mr$-dimensional spectrum
is as follows; $a\in I$ and $l\notin I$.
\begin{center}
\begin{tabular}{@{}lll@{}}\toprule
Block & Dimension & Eigenvalues\\\midrule
Amplitudes & $d$ & Strictly negative\\
Active--active pairs & $d(d-1)$ & One negative and one zero per pair\\
Unused hidden, active input & $(m-d)d$ & Zero\\
Active hidden, unused input & $d(r-d)$ & $\lambda_l/\lambda_a-1$\\
Unused hidden and input & $(m-d)(r-d)$ & $a_K\lambda_l/s_{I,K}-1$\\\bottomrule
\end{tabular}
\end{center}
The zeros in the active--active and unused-hidden/active-input
blocks are tangent to $\{OQ_{I,K}:O\in\Ogroup(m)\}$.
Additional zeros in the last block are transverse to the
rank-$d$ manifold. After factors with the same Gram matrix are
identified, the derivative along that manifold is Hurwitz.
At $K=0$ the same conclusions hold at the leading rank-one
zero, with both unused-input rates equal to
$\lambda_l/\lambda_1-1$.
\end{lemma}
\begin{proof}(of Lemma~\ref{lem:spectrum}).
\label{proof:lem:spectrum}
Use left and right bases in which $Q_{I,K}$ has active diagonal entries $q_a=\norm{Q_{I,K}v_a}$ and its normalized
product $Q_{I,K}C/s_{I,K}$ has entries $s_a=\lambda_aq_a/s_{I,K}$. Equation
\eqref{eq:zero-system} gives $q_a=p_K(s_a)$. The normalization derivative is
\[
 D\!\left(\frac{QC}{\norm{QC}_F}\right)_{Q_{I,K}}[E]
 =\frac{EC}{s_{I,K}}-\frac{Q_{I,K}C}{s_{I,K}^3}\langle Q_{I,K}C,EC\rangle.
\]
Differentiating the first cubic iterate gives
\begin{equation}\label{eq:cubic-derivative}
 DP_1(X)[E]=\tfrac32E-\tfrac12(EX^\top X+XE^\top X+XX^\top E).
\end{equation}
These two formulas account for all blocks.

Unused hidden and input directions.

For $E=u_\perp v_l^\top$ with $u_\perp$ orthogonal to the active
hidden span and $l\notin I$,
\[
 EC=\lambda_lE,\qquad \langle Q_{I,K}C,EC\rangle=0,
 \qquad
 D(QC/\norm{QC}_F)_{Q_{I,K}}[E]=\frac{\lambda_l}{s_{I,K}}E.
\]
Orthogonality on both sides gives
\[
 E(Q_{I,K}C)^\top(Q_{I,K}C)=0,\qquad
 (Q_{I,K}C)E^\top(Q_{I,K}C)=0,\qquad
 (Q_{I,K}C)(Q_{I,K}C)^\top E=0.
\]
Positive scalar normalization preserves these equalities. Each successive polynomial iterate preserves the active left and right singular subspaces, so the cubic correction vanishes at every differentiation step. The increment is multiplied by $3/2$ at each step and hence by $a_K$ in total. Multiplication by $\lambda_l/s_{I,K}$ and subtraction of $E$ give the rate $a_K\lambda_l/s_{I,K}-1$.

One-sided changes of orientation.

On an active hidden direction and unused input direction, the derivative of $P_K$ multiplies by $p_K(s_a)/s_a$. One way to see this is to differentiate a rotation of the right singular vector, for which singular values do not change. The resulting rate is
\[
 \frac{\lambda_l}{s_{I,K}}\frac{p_K(s_a)}{s_a}-1
 =\frac{\lambda_l}{\lambda_a}-1,
\]
using the identity in \eqref{eq:zero-system}. A rotation of the left singular vector gives the same multiplier with $\lambda_a$ in place of $\lambda_l$, hence rate zero. These are exactly the $(m-d)d$ left-rotation directions.

Amplitude changes.

For a diagonal perturbation, $\dd s_{I,K}=\sum_b\lambda_b s_b\dd q_b$ and
\[
 \dd s_a=\frac{\lambda_a}{s_{I,K}}\dd q_a-\frac{s_a}{s_{I,K}}\dd s_{I,K}.
\]
Consequently the amplitude derivative is
\[
 \diag(s_ap_K'(s_a)/q_a-1)-\alpha\beta^\top,
 \quad
 \alpha_a=\frac{p_K'(s_a)s_a}{s_{I,K}},\quad\beta_a=\lambda_as_a.
\]
For $d\ge2$ each $s_a\in(0,1)$, so $0<s_ap_K'(s_a)/q_a<1$: the upper inequality is equivalent to $q_K'(s_a^2)<0$, supplied by Lemma~\ref{lem:update-map}. Conjugating by $\diag(\sqrt{\beta_a/\alpha_a})$ makes the rank-one term symmetric, with entries $\sqrt{\alpha_a\beta_a}\sqrt{\alpha_b\beta_b}$. Thus the conjugated matrix is a negative definite diagonal matrix minus a positive semidefinite matrix. All amplitude eigenvalues are strictly negative. At rank one, positive radial scaling leaves the normalized input unchanged, and the radial eigenvalue is directly $-1$.

An active off-diagonal pair.

For distinct active singular values $s_a,s_b$, differentiation of $Xq_K(X^\top X)$ on the pair $(E_{ab},E_{ba})$ gives the symmetric multiplier matrix with diagonal and off-diagonal coefficients
\[
 A_{ab}=\frac{s_aq_a-s_bq_b}{s_a^2-s_b^2},\qquad
 B_{ab}=\frac{s_bq_a-s_aq_b}{s_a^2-s_b^2}.
\]
Indeed the divided difference of $q_K$ is
$\{q_a/s_a-q_b/s_b\}/(s_a^2-s_b^2)$; multiplying out the derivative of $Xq_K(X^\top X)$ gives these two coefficients. Since the normalization derivative is zero on this pair, the block of $DF_K$ is
\[
 \begin{pmatrix}
 A_{ab}\lambda_b/s_{I,K}-1&B_{ab}\lambda_a/s_{I,K}\\
 B_{ab}\lambda_b/s_{I,K}&A_{ab}\lambda_a/s_{I,K}-1
 \end{pmatrix}.
\]
The vector $(q_b,-q_a)$ is a left-rotation direction and has eigenvalue zero. The other eigenvalue is the trace. Substituting $\lambda_a/s_{I,K}=s_a/q_a$ and $\lambda_b/s_{I,K}=s_b/q_b$ reduces that trace to
\begin{equation}\label{eq:pair-spectrum}
 -1+\frac{s_as_b(q_a^2-q_b^2)}{q_aq_b(s_a^2-s_b^2)}.
\end{equation}
If $s_a>s_b$, monotonicity gives $1<q_a/q_b<s_a/s_b$. Since $z\mapsto z-z^{-1}$ is strictly increasing on $(1,\infty)$, the fraction in \eqref{eq:pair-spectrum} lies in $(0,1)$, proving negativity. When $s_a=s_b$, also $q_a=q_b$. Taking the divided-difference
limits gives
\[
 A_{ab}=\frac12\left(p_K'(s_a)+\frac{q_a}{s_a}\right),\qquad
 B_{ab}=\frac12\left(p_K'(s_a)-\frac{q_a}{s_a}\right).
\]
Indeed, differentiation of $s\mapsto sp_K(s)$ gives the first limit;
differentiation of $s\mapsto p_K(s)/s$ in the off-diagonal expression
gives the second. Since $\lambda_a/s_{I,K} =\lambda_b/s_{I,K}=s_a/q_a$,
the trace is $-1+s_ap_K'(s_a)/q_a<0$. Here $0<s_a<1$ because the
active rank is at least two, and the strict inequality follows from
$q_K'(s_a^2)<0$. Matrix polynomial differentiation is continuous, so
these limiting coefficients are the actual derivative at the repeated
singular value.

The block dimensions sum to
\[
 d+d(d-1)+(m-d)d+d(r-d)+(m-d)(r-d)=mr.
\]
To identify the Gram derivative, the differential of $Q\mapsto Q^\top Q$
is $E\mapsto Q_{I,K}^\top E+E^\top Q_{I,K}$. In the active diagonal coordinates,
a tangent increment to the rank-$d$ factor manifold has zero
unused-hidden/unused-input block. Its Gram differential has arbitrary
symmetric active block and arbitrary active--unused block; its unused
principal block is zero. This is exactly the tangent space to the
rank-$d$ PSD manifold. If that differential is zero, the active increment
has the form $E_{11}=\Omega_{11}\diag(q_a)$ with
$\Omega_{11}^\top=-\Omega_{11}$, the active--unused increment is zero,
and the unused-hidden/active-input increment is an arbitrary
$\Omega_{21}\diag(q_a)$. These are precisely $E=\Omega Q_{I,K}$ for a
skew-symmetric $\Omega$. Thus the differential is onto with exactly
the left-rotation kernel. Differentiating the equivariance of the Gram
map under the factor flow intertwines the two linearizations.
Its induced isomorphism on the quotient identifies the remaining
factor eigenvalues with the fixed-rank Gram eigenvalues. The preceding
block calculation proves that all of these eigenvalues are negative. For $K=0$, differentiate $QC/\norm{QC}_F-Q$ directly: the radial rate is $-1$, hidden rotations have rate zero, and all unused-input directions have rate $\lambda_l/\lambda_1-1$.
\end{proof}

\label{def:rank-connection}
Fix $K\ge1$ and $1\le d<\min\{m,r\}$. Assume $Q_{d,K}$
exists and
\[
 \lambda_d>\lambda_{d+1}>0,\qquad g_{d+1}^{(d)}>0.
\]
For $1\le i\le d$, set
$u_i=Q_{d,K}v_i/\norm{Q_{d,K}v_i}$, and choose a unit $u_{d+1}$
orthogonal to these vectors.
A connection with positive initial growth in the
$u_{d+1}v_{d+1}^\top$ direction means a solution tending to
$Q_{d,K}$ as $v\to-\infty$ whose positive new amplitude is
asymptotic to a positive multiple of $e^{g_{d+1}^{(d)}v}$.
Fix its time translation at a sufficiently small prescribed
positive value of that amplitude.

\begin{lemma}[Connection to the next rank]\label{lem:connection}
Under these conditions, the connection exists in
$\{\sum_{i=1}^{d+1}q_i u_iv_i^\top:q_i>0\}$ and converges to
$Q_{d+1,K}$ up to a left orthogonal transformation. On every
fixed compact segment, its linear transport of the remaining
unused-hidden directions is positive diagonal in the original
task labels. In particular, for $l>d+1$ its multiplier on
$[v_a,v_b]$ is
\[
 \exp\!\left\{\int_{v_a}^{v_b}
 \left(\frac{a_K\lambda_l}{\norm{Q(v)C}_F}-1\right)\dd v\right\}>0.
\]
\end{lemma}
\begin{proof}(of Lemma~\ref{lem:connection}).
\label{proof:lem:connection}
In the diagonal factor subspace supported on $\{1,\ldots,d+1\}$, Lemma~\ref{lem:spectrum} makes the old amplitudes stable and the new amplitude the unique positive direction, of rate $g_{d+1}^{(d)}$. The field is analytic near $Q_{d,K}$. A smooth compactly supported cutoff equal to one there gives a complete $C^2$ field with the same linearization. Lemma~\ref{lem:local-exit-background} constructs its departure with positive new amplitude. The following bounds continue this local departure under the original field.

All diagonal amplitudes are positive. Define their normalized squared coordinates
\[
 x_i=\frac{\lambda_i^2\norm{Q(v)v_i}^2}{\norm{Q(v)C}_F^2},
 \qquad 1\le i\le d+1.
\]
The amplitude equation is
\[
 \frac{\dd}{\dd v}\norm{Q(v)v_i}
 =\norm{Q(v)v_i}\left\{\frac{\lambda_i q_K(x_i)}{\norm{Q(v)C}_F}-1\right\}.
\]
Its multiplicative form preserves positivity at finite times. Differentiating the norm and the normalized coordinates gives
\begin{equation}\label{eq:simplex}
\begin{aligned}
 x_i'&=\frac{2x_i}{\norm{Q(v)C}_F}
 \left\{\lambda_iq_K(x_i)-\sum_{a=1}^{d+1}\lambda_a x_aq_K(x_a)\right\},\\
 \frac{\dd}{\dd v}\norm{Q(v)C}_F
 &=\sum_{i=1}^{d+1}\lambda_i x_iq_K(x_i)-\norm{Q(v)C}_F.
\end{aligned}
\end{equation}
Thus $x_i>0$ and $\sum_i x_i=1$.

The function $x\mapsto\sum_i\lambda_i\int_0^{x_i}q_K(u)\dd u$ is strictly concave on the simplex; its Hessian has entries $\lambda_iq_K'(x_i)<0$ in the interior. By Lemma~\ref{lem:add-zero}, the vector with coordinates $q_K^{-1}(s_{d+1,K}/\lambda_i)$ is interior, sums to one, and has all gradient entries equal to $s_{d+1,K}$. Strict concavity makes it the unique maximizer. The relative-entropy derivative is
\begin{equation}\label{eq:entropy}
\begin{aligned}
 &\frac{\dd}{\dd v}\sum_iq_K^{-1}(s_{d+1,K}/\lambda_i)
 \log\frac{q_K^{-1}(s_{d+1,K}/\lambda_i)}{x_i}\\
 &\quad=-\frac2{\norm{Q(v)C}_F}
 \sum_i\bigl[q_K^{-1}(s_{d+1,K}/\lambda_i)-x_i\bigr]\lambda_iq_K(x_i)\\
 &\quad\le-\frac2{\norm{Q(v)C}_F}
 \sum_i\lambda_i\int_{x_i}^{q_K^{-1}(s_{d+1,K}/\lambda_i)}q_K(u)\dd u.
\end{aligned}
\end{equation}
The last line uses the concave tangent inequality. Positive weights make an entropy sublevel bounded away from every simplex boundary.

The scalar polynomial bounds also give
\[
 \frac{\lambda_{d+1}}{\lambda_1}-\norm{Q(v)}_F
 \le\frac{\dd}{\dd v}\norm{Q(v)}_F
 \le\sqrt{d+1}-\norm{Q(v)}_F.
\]
For the lower bound use $p_K(\sqrt{x_i})\ge\sqrt{x_i}$ and
$\norm{Q(v)C}_F\le\lambda_1\norm{Q(v)}_F$; the upper bound follows from $p_K\le1$ and Cauchy--Schwarz. Scalar integration keeps the scale bounded above and eventually bounded below by a positive constant, giving global continuation. Integrating \eqref{eq:entropy} makes its nonnegative final sum integrable over every forward tail. Its derivative is bounded on the resulting compact set. If this sum had infinitely many uniformly positive values, the derivative bound would give disjoint intervals of fixed positive length with uniformly positive mass, contradicting integrability. Thus the sum tends to zero. Compactness and the unique maximizer imply
\[
 x_i(v)\longrightarrow q_K^{-1}(s_{d+1,K}/\lambda_i).
\]
The sum $\sum_i\lambda_i x_i(v)q_K(x_i(v))$ consequently tends to $s_{d+1,K}$. Variation of constants in the scale equation gives
\begin{align*}
 \norm{Q(v)C}_F-s_{d+1,K}
 ={}&e^{-v}\bigl(\norm{Q(0)C}_F-s_{d+1,K}\bigr)\\
 &+\int_0^v e^{-(v-u)}
 \left[\sum_i\lambda_i x_i(u)q_K(x_i(u))-s_{d+1,K}\right]\dd u.
\end{align*}
Splitting at a sufficiently late fixed time proves convergence of the scale. Since $\norm{Q(v)v_i}=\norm{Q(v)C}_F\sqrt{x_i(v)}/\lambda_i$, the diagonal solution converges to $Q_{d+1,K}$.

For each remaining unused input label $l$, the normal linearization is multiplication by $a_K\lambda_l/\norm{Q(v)C}_F-1$. Its multiplier across a fixed segment is
\begin{equation}\label{eq:connection-transfer}
 \exp\left\{\int_{v_a}^{v_b}
 \left(\frac{a_K\lambda_l}{\norm{Q(v)C}_F}-1\right)\dd v\right\}>0.
\end{equation}
Labels evolve independently. The product for $h$ labels is ordered by their summed signed eigenvalues, since the common coefficient $a_K/\norm{Q(v)C}_F$ is positive.
\end{proof}

The solution connects the two local neighborhoods. The Gaussian data and the subsequent passage results
identify the independent coefficients and show how the actual
trajectory follows the prescribed additions.

The algebraic construction in Section~\ref{sec:entrance} uses only
existing supports. The following proposition verifies the finite
procedure and identifies its exceptional exponent set.
\begin{proposition}[Clocks and exceptional exponents]
\label{prop:clock-domain}
Under Assumptions~\ref{ass:task} and~\ref{ass:prefix}, the two
algebraic procedures in Section~\ref{sec:entrance} are well defined. Every candidate support used by either procedure exists;
all its normalizations are uniquely determined. The finite coefficients
$c_{d,K}$ are independent of $b$, positive for $d\ge2$, and strictly
increasing along the algebraic sequence. Every finite candidate time satisfies
\[
 x_d=(b-1)c_{d,K},\qquad |\mathcal B_{C,K}|\le p-1.
\]
The set $\mathcal B_C$ is at most countable.
For $b\notin\mathcal B_{C,K}$ every evaluated finite time differs from
one. For $b\notin\mathcal B_C$ this holds simultaneously at every
finite depth. All spectral boundaries used in these procedures are
strict by Assumption~\ref{ass:prefix}.
\end{proposition}
\begin{proof}(of Proposition~\ref{prop:clock-domain}).
\label{proof:prop:clock-domain}
The initial zero $Q_{1,K}=e_1v_1^\top$ exists for every finite $K$,
with $s_{1,K}=\lambda_1$. Suppose a leading support of size $d<p$ has
been constructed. All of its eigenvalues and $\lambda_{d+1}$ are
positive and $\lambda_d>\lambda_{d+1}$. If its next rate is positive,
Lemma~\ref{lem:add-zero} constructs the next support and
Lemma~\ref{lem:gram-zeros} gives its unique Gram matrix and scale.
The hidden dimension permits the addition since $d<p\le m$.
If the rate is nonpositive, the algebraic procedure stops and assigns
infinite coefficients thereafter, without evaluating any nonexistent
normalization. This proves existence and uniqueness at every evaluated
step by induction. At $K=0$ the first rate is
$\lambda_2/\lambda_1-1<0$, so the initial step already stops.
There are at most $p-1$ additions.

The coefficient $c_{2,K}$ is positive when finite. At each subsequent
finite step, $1-\lambda_{d+1}/\lambda_d>0$ and the rate denominator is
positive. The added term in \eqref{eq:tau-clock} is positive, proving
strict increase and positivity of every finite $c_{d,K}$ for $d\ge2$.
All inputs to this equation depend only on $C,K$, so the coefficients
are independent of $b$.

For the actual candidate-time procedure, the first finite quotient
is $(b-1)c_{2,K}$. If $x_d=(b-1)c_{d,K}$ has been established, direct
substitution gives
\[
 x_{d+1}=(b-1)c_{d,K}
 +\frac{(b-1)(1+c_{d,K})(1-\lambda_{d+1}/\lambda_d)}
 {g_{d+1}^{(d)}}=(b-1)c_{d+1,K}.
\]
Its used supports form an initial segment of the previously constructed
algebraic sequence; stopping at a clock at least one creates no new
normalization. At a nonpositive rate both constructions stop. This
proves the identity at every evaluated finite candidate.

A finite candidate can equal one only if
$b=1+1/c_{d,K}$. There are at most $p-1$ such values at fixed $K$.
The union over integer $K\ge0$ is countable and hence has Lebesgue
measure zero: enumerate its points and cover them by intervals whose
lengths form a geometric series with arbitrarily small total sum.
The definitions now give both noncriticality assertions. Finally,
every tested boundary has index at most $\min\{p,r-1\}$.
Assumption~\ref{ass:prefix} therefore makes its eigenvalue difference
strictly positive, independently of the outcome of either procedure.
\end{proof}

\subsection{Exact normal equation and integrated relative estimates}\label{sec:relative-statements}
Local Gram coordinates separate stable tangent motion from the independent
unused columns. The exact projected equation is homogeneous in the full
normal matrix. This permits relative volume estimates even when those
columns are arbitrarily small.
For $I\subset\{1,\ldots,r\}$ put $Q_I=(Qv_l)_{l\in I}$.
Whenever the first $d$ task columns are independent, project the remaining
columns onto $\Ran(Q_{\{1,\ldots,d\}})^\perp$ and express them in
orthonormal hidden coordinates:
\[
 Z(Q)=(z_{d+1},\ldots,z_r)\in\R^{(m-d)\times(r-d)}.
\]
This extends the rank-one definition in Section~\ref{sec:gaussian-data}.
For $0\le h\le\min\{m-d,r-d\}$ set
\begin{equation}\label{eq:volumes}
 V_h(Z)=\norm{z_{d+1}\wedge\cdots\wedge z_{d+h}},\qquad V_0(Z)=1.
\end{equation}
At rank $d$, the induction retains the orders
\label{def:prefixes}\begin{equation}\label{eq:prefix-family}
 1\le h\le p-d,\qquad J_{d,h}=\{d+1,\ldots,d+h\},\qquad p=\min\{m,n_+(C)\}.
\end{equation}
At $d=p$, the retained family is empty.
For any order admitting a competing column set, define
\begin{equation}\label{eq:explicit-exterior-gap}
 \sigma_{d,h}=\sum_{l=d+1}^{d+h}g_l^{(d)}-
 \max_{\substack{J\subset\{d+1,\ldots,r\},\ |J|=h\\J\ne J_{d,h}}}
 \sum_{l\in J}g_l^{(d)}.
\end{equation}
Set $\sigma_{d,h}=+\infty$ if the leading column set is the sole set.
The incoming powers of the consecutive ratios $V_h/V_{h-1}$ are specified
in Lemma~\ref{lem:main-relative-growth}.
The quotient $V_h/V_{h-1}$, when its denominator is positive, is the
length of the last column after orthogonal projection away from its
predecessors.

\begin{lemma}[Leading exterior gaps]\label{lem:gap-equivalence}
For an existing $Q_{d,K}$ and any order
$1\le h\le\min\{m-d,r-d\}$ with $d+h<r$,
\[
 \sigma_{d,h}=\frac{a_K}{s_{d,K}}
 (\lambda_{d+h}-\lambda_{d+h+1})\ge0.
\]
Under Assumption~\ref{ass:prefix}, this gap is strictly positive for
every retained order $1\le h\le p-d$. Each active boundary
$\lambda_d>\lambda_{d+1}$ with $d\le p$ and $d<r$ is also strict.
For $d+h=r$ the leading column set is the only set and the gap is
$+\infty$ by convention.
\end{lemma}
\begin{proof}(of Lemma~\ref{lem:gap-equivalence}).
\label{proof:lem:gap-equivalence}
The rates have the same signed order as the task eigenvalues because
$a_K/s_{d,K}>0$. The largest sum over $h$ distinct unused indices is their
first $h$-term sum. Among different sets a largest sum replaces its last
member by the next unused index: replacing an earlier term removes
an equal or larger rate, and adding a later index adds an equal or
smaller rate. The difference is therefore
$g_{d+h}^{(d)}-g_{d+h+1}^{(d)}
=(a_K/s_{d,K})(\lambda_{d+h}-\lambda_{d+h+1})\ge0$.
For a retained order $d+h\le p$, Assumption~\ref{ass:prefix}
makes this expression strictly positive whenever a competing set exists.
When every unused index is chosen there is exactly one set. The
active-boundary assertion is a direct part of Assumption~\ref{ass:prefix}.
\end{proof}

The coordinate identities and norm estimates in this subsection apply
in any positive matrix dimensions under their stated active-column and
spectral-boundary conditions. A neighborhood of $Q_{d,K}$ means the
inverse image under $Q\mapsto Q^\top Q$ of a small neighborhood of
$Q_{d,K}^\top Q_{d,K}$; its definition is independent of the hidden
orthogonal coordinates. For the local estimates, write $S=Q^\top Q$ in the task basis, with block
indices $\{1,\ldots,d\}$ and $\{d+1,\ldots,r\}$, and define
\[
 y=(S_{11}-(Q_{d,K}^\top Q_{d,K})_{11},S_{12}),\qquad
 H=S_{22}-S_{21}S_{11}^{-1}S_{12}=Z^\top Z.
\]
Choose an orthonormal complementary frame $U_\perp$ at the initial point
and extend it by the linear equation
\[
 U_\perp'=-Q_{\{1,\ldots,d\}}
 (Q_{\{1,\ldots,d\}}^\top Q_{\{1,\ldots,d\}})^{-1}
 (Q_{\{1,\ldots,d\}}')^\top U_\perp.
\]
Here and below the prime denotes differentiation in $v=\log t$.
Differentiation gives $Q_{\{1,\ldots,d\}}^\top U_\perp=0$ and
$U_\perp^\top U_\perp=I$ throughout the interval; also
$U_\perp^\top U_\perp'=0$. The projected matrix is
$Z=U_\perp^\top Q_{\{d+1,\ldots,r\}}$.
Retain the actual coefficients
\[
 R_t=C-t^2\Lop(S),\qquad
 s_t=\norm{QR_t}_F+\frac{c_\eta\eps}{2t},\qquad
 \TF=\frac{R_t}{s_t}q_K\!\left(\frac{R_tSR_t}{s_t^2}\right).
\]
With $\Gamma_d=\diag_{d<l\le r}(g_l^{(d)})$, define
\[
 E=\TF_{22}-I-\TF_{21}S_{11}^{-1}S_{12}-\Gamma_d.
\]
The identity $\norm{QR_t}_F^2=\tr(R_tSR_t)$ expresses this coefficient
entirely in Gram coordinates, $t^2$, and $\eps/t$.

The limiting Gram equation used below is
\begin{equation}\label{eq:gram-field}
 T(S)=\frac C{\sqrt{\tr(CSC)}}
 q_K\!\left(\frac{CSC}{\tr(CSC)}\right)-I,
 \qquad S'=T(S)^\top S+ST(S).
\end{equation}
The prime denotes differentiation in $v=\log t$.
Let $J_d$ be the derivative of its $y$ equation with respect to $y$ at
$(y,H)=(0,0)$. By Lemma~\ref{lem:spectrum}, its linear equation is stable under the
strict active-boundary condition. We measure $y$ in a fixed norm equivalent to the product Frobenius
norm and chosen so that this linear equation contracts. For a nonempty normal block, put
$g=g_{d+1}^{(d)}$, the largest unused rate.
For each retained admissible order $h$, the leading-volume cone
of width $\zeta$ means
\[
 \norm{\wedge^h Z}_F^2\le(1+\zeta^2)V_h(Z)^2.
\]
Only orders required in a given result are retained. The local equations
and tangent contraction use $\lambda_d>\lambda_{d+1}$ when $d<r$.
The estimates for individual volumes additionally use
$\sigma_{d,h}>0$ for their retained orders. The estimates for
$\norm Z_F$ alone use the largest rate $g$, also when it is zero. We write
\[
 D^+ a(v)=\limsup_{h\downarrow0}\frac{a(v+h)-a(v)}h
\]
for the upper right Dini derivative of a scalar function.

\begin{lemma}[Integrated relative volume estimates]\label{lem:integral}
In a sufficiently small coordinate neighborhood of $Q_{d,K}$, the
actual equations satisfy
\begin{equation}\label{eq:chart-estimate}
 \begin{aligned}
 y'&=J_dy+O(\norm y^2+\norm Z_F^2+t^2+\eps/t),\\
 Z'&=Z(\Gamma_d+E),\\
 \norm E_\op&\le C(\norm y+\norm Z_F^2+t^2+\eps/t),\\
 D^+\norm y&\le-\omega\norm y+C(\norm Z_F^2+t^2+\eps/t)
 \end{aligned}
\end{equation}
The identity for $Z$ in this display holds throughout every interval
where the active task columns are independent and the normalized update is defined, in the
prescribed frame. The other relations are local estimates near $Q_{d,K}$.
Here $\omega>0$ is fixed and the last inequality uses the contracting
equivalent norm described above. Consider any interval $[t_a,t_b]$ in
this neighborhood with
$t_a\ge c_U\eps$ for fixed $c_U>0$ and $t_b\le t_*$ for a fixed small
$t_*>0$. Suppose the incoming tangent coordinate is in a fixed small
ball and all required leading-volume cones hold at $t_a$. The neighborhood,
$t_*$ and an upper bound on $\eps/t_a$ are chosen so that the coefficient
error is uniformly small enough for cone invariance. The cones then
persist on this interval. For each retained order with $V_h(Z(t_a))>0$,
the principal volume remains positive and
\begin{equation}\label{eq:local-relative-volume}
 \left|\log\frac{V_h(Z(t))}{V_h(Z(t_a))}
 -\Bigl(\sum_{l\in J_{d,h}}g_l^{(d)}\Bigr)\log(t/t_a)\right|
 \le C_h\int_{\log t_a}^{\log t}\norm E_\op\,\dd v,
 \quad\text{for each retained order}.
\end{equation}
Here the constants are independent of the size of the leading volumes.
Suppose $g>0$ and
the first volume has logarithmic growth at least $g/2$. Then, for
$t_a\le s\le t_b$,
\[
 \norm{Z(s)}_F\le\sqrt{1+\zeta^2}\,\norm{Z(t_b)}_F(s/t_b)^{g/2},
\]
and
\begin{equation}\label{eq:integral-error}
 \begin{aligned}
 \int_{\log t_a}^{\log t_b}\norm Z_F^2\,\dd v
 &\le\frac{C}{g}\norm{Z(t_b)}_F^2,\\
 \int_{\log t_a}^{\log t_b}(\norm y+\norm E_\op)\,\dd v&\le C_U.
 \end{aligned}
\end{equation}
The endpoint may be any stopping time for which the stated local
conditions hold. These constants are independent of the interval length.
For a sufficiently narrow first-column cone and sufficiently small
coefficient error, there is a fixed $C\ge1$ such that
\[
 C^{-1}\tr H\le (\tr H)'\le C\,\tr H
\]
on the same positive-growth interval.

For the following norm estimate, assume the local equations and tangent
contraction above, with any multiplicities among the unused rates;
no volume-cone hypothesis is imposed. Suppose
\[
 \norm{Z(t_a)}_F\le C_U\eps^e,\qquad e>0,\qquad
 \log(t_*/t_a)=\theta\log(1/\eps)+O_U(1),\qquad \theta\ge0.
\]
If $e-g\theta>0$, choose a fixed $\nu>0$ such that
$e-(g+\nu)\theta>0$ and a neighborhood with $\norm E_\op\le\nu$.
Then the same integrated bound holds up to $t_*$,
\[
 \norm{Z(t)}_F\le C_U\eps^e(t/t_a)^{g+\nu},
 \qquad t_a\le t\le t_*.
\]
The supremum of $\norm Z_F$ on this interval tends to zero as a positive
power of $\eps$, and its squared logarithmic-time integral tends to zero.
These conclusions include $g\le0$.
If all unused task eigenvalues are negative, a sufficiently small
coefficient error instead gives
\[
 \norm{Z(t)}_F\le\norm{Z(t_a)}_F(t/t_a)^{-1/2};
\]
the squared-normal and coefficient integrals are bounded without a cone
condition or an incoming power-law assumption. If the normal block is
empty, the conclusions hold with $Z=0$.
All interval assertions are local: they hold while the trajectory is in
the stated coordinate neighborhood.
\end{lemma}
\begin{proof}(of Lemma~\ref{lem:integral}).
\label{proof:lem:integral}
Fix $Q_{d,K}$ and order the first $d$ task coordinates first. Throughout this proof primes denote derivatives in $v=\log t$.

The horizontal-frame equation prescribed before the lemma has continuous coefficients while the active columns are independent. Differentiation gives
\begin{align*}
 (Q_{\{1,\ldots,d\}}^\top U_\perp)'
 &=(Q_{\{1,\ldots,d\}}')^\top U_\perp
 -(Q_{\{1,\ldots,d\}}')^\top U_\perp=0,\\
 U_\perp^\top U_\perp'&=0,\qquad
 (U_\perp^\top U_\perp)'=0.
\end{align*}
Thus $U_\perp$ remains an orthonormal complementary frame. The polynomial update identity gives the complete actual equation $Q'=Q(\TF-I)$, with the empirical $R_t$ and regularized $s_t$ defined above. Direct differentiation of $Z=U_\perp^\top Q_{\{d+1,\ldots,r\}}$ gives
\begin{align*}
 Z'&=U_\perp^\top Q_{\{d+1,\ldots,r\}}'
      -U_\perp^\top Q_{\{1,\ldots,d\}}'S_{11}^{-1}S_{12},\\
 U_\perp^\top Q_{\{1,\ldots,d\}}'&=Z\TF_{21},\qquad
 U_\perp^\top Q_{\{d+1,\ldots,r\}}'=Z(\TF_{22}-I).
\end{align*}
Consequently,
\begin{equation}\label{eq:schur-normal}
 Z'=Z(\TF_{22}-I-\TF_{21}S_{11}^{-1}S_{12})=Z(\Gamma_d+E).
\end{equation}
This identity preserves every vanishing independent component under the complete empirical equation.

The limiting Gram equation follows by differentiating $S=Q^\top Q$. At its base point,
$\tr(CQ_{d,K}^\top Q_{d,K}C)=s_{d,K}^2>0$.
Thus its polynomial expression extends analytically to an open symmetric-matrix neighborhood with positive denominator and invertible $S_{11}$. For every differentiable symmetric $S$ there, differentiation of the Schur complement gives
\begin{align*}
 (S_{22}-S_{21}S_{11}^{-1}S_{12})'
 ={}&S_{22}'-S_{21}'S_{11}^{-1}S_{12}
 -S_{21}S_{11}^{-1}S_{12}'\\
 &+S_{21}S_{11}^{-1}S_{11}'S_{11}^{-1}S_{12}.
\end{align*}
Substitute $S'=(\TF-I)^\top S+S(\TF-I)$ in these four terms. The terms containing $\TF_{12}$ cancel between the first three terms; the terms containing $\TF_{11}-I$ cancel with the last. The remaining terms factor as
\begin{equation}\label{eq:H-schur}
 H'=(\Gamma_d+E)^\top H+H(\Gamma_d+E),
 \qquad H=S_{22}-S_{21}S_{11}^{-1}S_{12}.
\end{equation}
This algebraic identity holds throughout that symmetric neighborhood. For actual factors, orthogonal projection gives $H=Z^\top Z\succeq0$.

In the coordinates $y$ defined above, $S_{22}=H+S_{21}S_{11}^{-1}S_{12}$. Invertible block elimination gives
$\rank S=d+\rank H$ while $S_{11}\succ0$. Thus $H=0$ describes the fixed-rank Gram manifold. The derivative of its tangent equation is $J_d$. Lemma~\ref{lem:spectrum}, Taylor expansion, and Lemma~\ref{lem:rescaling} prove the first three relations in \eqref{eq:chart-estimate}; the squared-normal term comes from $H=Z^\top Z$.

To obtain a contracting tangent norm for the Hurwitz matrix $J_d$, retain
\[
 P_d=\int_0^\infty e^{J_d^\top u}e^{J_du}\dd u,
 \qquad \norm y=(y^\top P_dy)^{1/2}.
\]
The integral converges and is positive definite; integration of its derivative gives
$J_d^\top P_d+P_dJ_d=-I$.
For $y'=J_dy$, the squared norm derivative is $-\norm y_2^2$. Norm equivalence gives contraction. Choose the common $\omega>0$ with enough margin to absorb the quadratic remainder in a smaller fixed neighborhood. The forcing terms are bounded there, giving the last relation in \eqref{eq:chart-estimate}. These constants are uniform on the fixed chart.

The determinant identity
\[
 V_h(Z)^2=\det\bigl((z_{d+a}^\top z_{d+b})_{a,b=1}^h\bigr)
\]
characterizes column independence. Orthogonal elimination gives $V_h/V_{h-1}$ as the last column's distance from the preceding span. For example, right multiplication of $\diag(\eps,\eps^3,\eps^6)$ by $I+\eps e_2e_3^\top$ changes the third column to $(0,\eps^4,\eps^6)^\top$; its third independent scale stays $\eps^6$. The incoming power assumptions in Lemma~\ref{lem:main-relative-growth} provide two-sided estimates for these independent scales.

The matrix $\wedge^h Z$ has the $h$-fold column wedges as its columns. The product rule applied to \eqref{eq:schur-normal} gives
\begin{equation}\label{eq:compound}
 (\wedge^h Z)'=(\wedge^h Z)(\Gamma_d^{[h]}+E^{[h]}),
 \qquad\norm{E^{[h]}}_\op\le h\norm E_\op.
\end{equation}
For any square matrix, the superscript $[h]$ denotes the operator obtained by inserting it into one exterior slot at a time and summing. On the full tensor product the summands each have its operator norm; restriction to the alternating subspace preserves the resulting upper bound. The diagonal operator $\Gamma_d^{[h]}$ has rate $\sum_{l\in J}g_l^{(d)}$ on group $J$.

For use throughout this subsection, let $B_h$ collect the columns of $\wedge^hZ$ other than $J_{d,h}$. The distinguished column has norm $V_h(Z)$. When a competing group exists, norm differentiation and the one-sided Dini bounds at zeros give
\begin{equation}\label{eq:dini}
\begin{aligned}
 D^+\norm{B_h}_F
 &\le\left(\sum_{l\in J_{d,h}}g_l^{(d)}-\sigma_{d,h}\right)\norm{B_h}_F
 +h\norm E_\op\bigl(V_h(Z)+\norm{B_h}_F\bigr),\\
 D^-V_h(Z)
 &\ge\left(\sum_{l\in J_{d,h}}g_l^{(d)}\right)V_h(Z)
 -h\norm E_\op\bigl(V_h(Z)+\norm{B_h}_F\bigr).
\end{aligned}
\end{equation}
Here $D^-$ is the lower right Dini derivative. At a positive principal volume, the quotient rule yields
\[
 D^+\frac{\norm{B_h}_F}{V_h(Z)}
 \le-\sigma_{d,h}\frac{\norm{B_h}_F}{V_h(Z)}
 +h\norm E_\op\left(1+\frac{\norm{B_h}_F}{V_h(Z)}\right)^2.
\]
Choose the coefficient error small enough that the right side is negative at ratio $\zeta$. This proves cone invariance. Inside the cone, the principal logarithmic derivative is bounded, so its initially positive volume cannot vanish at a finite time. Its deviation from the principal rate is at most $h\sqrt{1+\zeta^2}\norm E_\op$. Hence
\begin{equation}\label{eq:logvolume}
 \left|\log\frac{V_h(Z(t))}{V_h(Z(t_a))}
 -\left(\sum_{l\in J_{d,h}}g_l^{(d)}\right)\log(t/t_a)\right|
 \le h\sqrt{1+\zeta^2}\int_{t_a}^t\norm{E(s)}_\op\frac{\dd s}s.
\end{equation}
This proves \eqref{eq:local-relative-volume}. If the distinguished group is the sole group, $B_h$ is empty and the logarithmic estimate follows directly; its cone requires no gap estimate.

A residence interval can have length proportional to $\log(1/\eps)$. Integration of the logarithmic-volume equation accumulates the coefficient error across that interval, so we now bound its integral uniformly in the interval length.

For positive growth, the first cone gives
$V_1(Z)\le\norm Z_F\le\sqrt{1+\zeta^2}V_1(Z)$.
When $(\log V_1)'\ge g/2$, backward integration from any endpoint $t_b$ gives
\[
 \norm{Z(s)}_F\le\sqrt{1+\zeta^2}\norm{Z(t_b)}_F(s/t_b)^{g/2},
 \quad t_a\le s\le t_b.
\]
Thus
\[
 \int_{t_a}^{t_b}\norm{Z(s)}_F^2\frac{\dd s}s
 \le\frac{1+\zeta^2}{g}\norm{Z(t_b)}_F^2.
\]
The tangent inequality integrates to
\[
 \norm{y(t)}\le(t_a/t)^\omega\norm{y(t_a)}
 +C\int_{t_a}^t(s/t)^\omega
 \bigl(\norm{Z(s)}_F^2+s^2+\eps/s\bigr)\frac{\dd s}s.
\]
A second integration and interchange of nonnegative integrals give
\[
 \int_{t_a}^{t_b}\norm{y(t)}\frac{\dd t}t
 \le\frac{\norm{y(t_a)}}\omega
 +\frac C\omega\int_{t_a}^{t_b}
 \bigl(\norm{Z(t)}_F^2+t^2+\eps/t\bigr)\frac{\dd t}t.
\]
The remaining integral is explicit:
\begin{equation}\label{eq:forcing-integral}
 \int_{t_a}^{t_b}(t^2+\eps/t)\frac{\dd t}t
 =\frac{t_b^2-t_a^2}{2}+\eps\left(\frac1{t_a}-\frac1{t_b}\right)
 \le\frac{t_*^2}{2}+\frac1{c_U}.
\end{equation}
The bound for $E$ in \eqref{eq:chart-estimate} now proves \eqref{eq:integral-error}. The inequality $t_a\ge c_U\eps$ bounds this integral, while a sufficiently small $\eps/t_a$ supplies the pointwise error bound needed for cone invariance.

For the trace estimate, positivity and the first cone imply
\[
 \tr(\Gamma_dH)\ge
 \frac{g-\zeta^2\norm{\Gamma_d}_\op}{1+\zeta^2}\tr H.
\]
Choose $\zeta$ so that this coefficient is positive. Equation~\eqref{eq:H-schur} and
$|\tr(HE)|\le\norm E_\op\tr H$ give the positive lower trace rate when $E$ is small. Boundedness of $\Gamma_d+E$ gives the upper rate, also for every PSD $H$ in the Gram chart.

For the final norm estimate choose $\nu>0$ with $e-(g+\nu)\theta>0$. As long as $\norm E_\op\le\nu$,
\[
 D^+\norm Z_F\le(g+\nu)\norm Z_F,
 \qquad \norm{Z(t)}_F\le C_U\eps^e(t/t_a)^{g+\nu}.
\]
If $g+\nu\ge0$, the supremum is $O_U(\eps^{e-(g+\nu)\theta})$; if $g+\nu<0$, it is bounded by $C_U\eps^e$. In both cases its squared integral is bounded by a positive power of $\eps$ times $O_U(\log(1/\eps))$ and tends to zero. The tangent convolution then closes the neighborhood bootstrap after the initial tangent radius, $t_*$, and $\eps/t_a$ are chosen sufficiently small. It also gives the coefficient integral bound. If all unused $\lambda_l$ are negative, all $g_l^{(d)}<-1$; choosing $\norm E_\op\le1/2$ gives $D^+\norm Z_F\le-\norm Z_F/2$, whose squared integral is bounded directly. For an empty normal block, set $Z=0$ in the same tangent argument.
\end{proof}

\subsection{First sections and propagation through one passage}
A single outgoing section must work for the actual trajectories and for
the limiting exit. Transversality fixes its first crossing, after which
projection and a positive relative transfer preserve the remaining
independent exponents.
\begin{lemma}[Uniform first-section limit]\label{lem:uniform-first-section}
Suppose $g_{d+1}^{(d)}>0$ is the unique largest unused rate.
Choose a sufficiently small cone width $\zeta>0$ and fix any $C_0>0$.
For a sufficiently small outgoing level $h_0>0$ and $0<h<h_0$, define
in the symmetric Gram chart
\begin{equation}\label{eq:first-section-data}
 \mathcal A_h=\left\{(y,H):H\succeq0,\ \tr H=h^2,\
 H_{d+1,d+1}\ge\frac{h^2}{1+\zeta^2},\ \norm y\le C_0h^2\right\}.
\end{equation}
There is no rank restriction on $H$ in this set.
For every initial value in $\mathcal A_h$, the first exit of the
limiting Gram equation \eqref{eq:gram-field} through $\tr H=h_0^2$
exists and is transverse. As $h\downarrow0$, these exit states converge
uniformly over $\mathcal A_h$ to the corresponding first crossing in
Gram coordinates of the connecting solution of
Lemma~\ref{lem:connection} through the same section.
The outgoing level and convergence rate may depend on $C_0$, the
matrix $Q_{d,K}$, and $K$. Only $h$ tends to zero.
\end{lemma}
\begin{proof}(of Lemma~\ref{lem:uniform-first-section}).
\label{proof:lem:uniform-first-section}

Identification of the connecting Gram trajectory.

Let $e_l$ denote the standard coordinate in $\R^{r-d}$ with task label $l$.
In the coordinates $(y,H)$, the full Gram Jacobian is
\begin{equation}\label{eq:Gram-Jacobian}
 \begin{pmatrix}J_d&\left[D_Hy\prime\right]_{(0,0)}\\0&[H\mapsto\Gamma_dH+H\Gamma_d]\end{pmatrix}.
\end{equation}
The lower eigenvalues are $g_l^{(d)}+g_k^{(d)}$. For a unique largest positive $g_{d+1}^{(d)}$, the simple largest eigenvalue is $2g_{d+1}^{(d)}$ and its principal \emph{left} functional is
\begin{equation}\label{eq:left-functional}
 (y,H)\longmapsto H_{d+1,d+1}.
\end{equation}
The right eigenvector normalized by $H_{d+1,d+1}=1$ is
\[
 \left(
 (2g_{d+1}^{(d)}I-J_d)^{-1}
       \left[D_Hy\prime\right]_{(0,0)}(e_{d+1}e_{d+1}^\top),
       e_{d+1}e_{d+1}^\top
 \right).
\] The connecting trajectory of Lemma~\ref{lem:connection} gives this Gram solution, up to time translation. Indeed its new factor amplitude is proportional to $e^{g_{d+1}^{(d)}v}$ as $v\to-\infty$, so $H_{d+1,d+1}$ is proportional to $e^{2g_{d+1}^{(d)}v}$; its tangent component solves the forced stable equation and has the right eigenvector just displayed. Uniqueness of the solution with the prescribed positive leading growth from the local-exit theorem identifies the two descriptions. 

The simplicity of $2g_{d+1}^{(d)}$ is a statement in the full symmetric Gram space. The corresponding factor direction has multiplicity $m-d$; the Gram left functional is simple and satisfies the relative projection bound used below.

For $H\succeq0$ choose any square-root factor $H=Z^\top Z$ to check the norm inequalities; this is only a device for the Gram calculation and imposes no new condition on the actual hidden width. The cone in \eqref{eq:first-section-data} is the first column cone in Gram form, since $\tr H=\norm Z_F^2$ and $H_{d+1,d+1}=\norm{z_{d+1}}^2$. Choose $\zeta$ with $\norm{\Gamma_d}_\op\zeta^2<g_{d+1}^{(d)}/4$, and then shrink the chart so that the coefficient error has norm less than a fixed sufficiently small multiple of $g_{d+1}^{(d)}$. Inside the cone,
\[
 \tr(\Gamma_dH)\ge g_{d+1}^{(d)}H_{d+1,d+1}-\norm{\Gamma_d}_\op\sum_{l\ne d+1}H_{ll}
 \ge(g_{d+1}^{(d)}-\norm{\Gamma_d}_\op\zeta^2)H_{d+1,d+1}.
\]
Differentiating $H=Z^\top Z$ using the normal equation of
Lemma~\ref{lem:integral} gives
$H'=(\Gamma_d+E)^\top H+H(\Gamma_d+E)$. The error contributes $2\tr(HE)$, whose absolute value is at most $2\norm E_\op\tr H$ for PSD $H$. After choosing the error small, the previous lower bound therefore gives $(\tr H)'\ge(g_{d+1}^{(d)}/2)\tr H$. The upper trace-growth estimate follows from Lemma~\ref{lem:integral}. Cone invariance follows from Lemma~\ref{lem:integral}. The stable tangent estimate keeps $y$ in a smaller chart while $\tr H\le h_0^2$: its initial size is at most $C_0h^2$ and its forcing is bounded by a constant times $\tr H$. Thus the first exit exists, is transverse, and is unique before leaving this chart.

For Lemma~\ref{lem:local-exit-background}, the field must be $C^2$, its largest eigenvalue must be a simple positive real number, and the incoming displacement must have uniformly positive relative projection on the associated left functional. Equation~\eqref{eq:gram-field} gives an analytic extension to an open symmetric-matrix neighborhood. A smooth compactly supported cutoff supplies a complete auxiliary field agreeing with it on the chart. Equations~\eqref{eq:Gram-Jacobian}--\eqref{eq:left-functional} give the largest eigenvalue $2g_{d+1}^{(d)}$ and left functional $H_{d+1,d+1}$. Finally, on $\mathcal A_h$,
\[
 \norm{(y,H)}\le(C_0+1)h^2,\qquad
 H_{d+1,d+1}\ge h^2/(1+\zeta^2).
\]
In particular
\[
 \frac{H_{d+1,d+1}}{\norm{(y,H)}}\ge\frac1{(1+\zeta^2)(C_0+1)}>0,
\]
which is the required relative projection bound independent of $h$. The hypotheses of that lemma allow arbitrary Jordan blocks in the remaining spectrum.

Apply Lemma~\ref{lem:local-exit-background} with phase shift $(2g_{d+1}^{(d)})^{-1}\log(1/H_{d+1,d+1})$. The shifted Gram flow converges to the same Gram trajectory on each compact phase interval. The convergence is uniform over $\mathcal A_h$: otherwise, a sequence of incoming states at levels tending to zero would satisfy the same uniform positive relative-alignment bound and violate the theorem. Local continuous dependence upgrades fixed-phase convergence to compact-phase convergence. Since the connecting solution crosses $\tr H=h_0^2$ transversely and the normal norm increases until that crossing, compact-phase convergence also identifies the first crossing and its state. This proves the asserted uniform convergence of the first exits.
\end{proof}

\begin{lemma}[Actual first-section limit]\label{lem:exit}
Fix $Q_{d,K}$ with a unique largest unused rate
$g_{d+1}^{(d)}>0$, a small coordinate neighborhood, and a sufficiently
narrow first-column cone. Choose the constant $C_0$ in
Lemma~\ref{lem:uniform-first-section} at least as large as the uniform
coefficient of $h^2$ obtained by combining the backward normal bound
and tangent contraction in Lemma~\ref{lem:integral}. Then choose
$h_0$ sufficiently small for that $C_0$ as in
Lemma~\ref{lem:uniform-first-section}.
Consider any actual trajectory family on $[t_a^\eps,t_b^\eps]$
satisfying \eqref{eq:chart-estimate} and that cone throughout.
Require the coefficient error to be uniformly small enough for the
positive lower and upper trace-growth bounds. Suppose
$y(t_a^\eps)$ lies in a fixed small ball,
\[
 0<\norm{Z(t_a^\eps)}_F\longrightarrow0,\qquad
 t_a^\eps\ge c_U\eps
\]
for fixed $c_U>0$, and $t_b^\eps$ is the first crossing of
$\norm Z_F=h_0$. Suppose also
\[
 t_b^\eps\longrightarrow0,\qquad \eps/t_b^\eps\longrightarrow0.
\]
The outgoing Gram converges to the Gram matrix at the first crossing
of the connecting solution of Lemma~\ref{lem:connection}. With that crossing assigned
phase zero in logarithmic time, convergence holds on each fixed compact
phase interval within the coordinate neighborhood or its fixed
continuation segment.
For every fixed $0<h<h_0$, let $t_h^\eps$ be the preceding first crossing
of $\norm Z_F=h$. There is a constant $C\ge1$, determined by the fixed
local data and independent of the seed and of $h$, such that
\begin{equation}\label{eq:relative-predecessor}
 \limsup_{\eps\downarrow0}\norm{y(t_h^\eps)}\le Ch^2,
 \qquad H_{d+1,d+1}(t_h^\eps)\ge\frac{h^2}{1+\zeta^2},
\end{equation}
and throughout the residence
\begin{equation}\label{eq:relative-trace-growth}
 C^{-1}\tr H\le\frac{\dd}{\dd\log t}\tr H\le C\tr H.
\end{equation}
\end{lemma}
\begin{proof}(of Lemma~\ref{lem:exit}).
\label{proof:lem:exit}
Fix a coordinate neighborhood and the contraction rate $\omega>0$
from Lemma~\ref{lem:integral}. Use a fixed $C\ge1$ for its trace-growth lower bound. Before a crossing $\norm Z_F=h$, it yields
$\norm{Z(v)}_F^2\le h^2e^{(v-\log t_h^\eps)/C}$.
The normal term in the tangent convolution is therefore at most
$Ch^2/(\omega+C^{-1})$. Choose $C_0$ strictly larger than this fixed
coefficient and then choose $h_0$ as in Lemma~\ref{lem:uniform-first-section}.
Shrinking this outgoing level preserves the constants fixed on the
larger neighborhood. The remaining incoming and nonautonomous terms
are shown below to vanish at every fixed $h$.
The first cone, bounded coefficients, and the positive trace-growth estimate in Lemma~\ref{lem:integral} compare $\norm Z_F$ to an exponentially increasing quantity, with uniform upper and lower rates. Because its incoming norm tends to zero, the logarithmic residence length tends to infinity. For any fixed $0<h<h_0$, let $t_h^\eps$ be the first crossing of $\norm Z_F=h$. The time between levels $h$ and $h_0$ is bounded above by a constant depending only on $h,h_0$ and the local rates. Hence $t_h^\eps/t_b^\eps$ is bounded below by a positive constant for this fixed $h$. It follows that
\[
 t_h^\eps\to0,\qquad\eps/t_h^\eps\to0,
 \qquad t_a^\eps/t_h^\eps\to0.
\]
The first cone at the auxiliary crossing gives
\begin{equation}\label{eq:coarse-projection}
 H_{d+1,d+1}=\norm{z_{d+1}}^2\ge\frac{h^2}{1+\zeta^2}.
\end{equation}

We next estimate the tangent coordinates at this crossing. The backward normal estimate from Lemma~\ref{lem:integral}, now ending at level $h$, inserted in the tangent convolution gives a contribution at most $Ch^2$, where $C$ is independent of the seed and of $h$. The initial tangent term is bounded by $C(t_a^\eps/t_h^\eps)^\omega$ and tends to zero. The $t^2$ forcing contributes at most $C(t_h^\eps)^2$. The regularization contribution is
\[
 C\eps(t_h^\eps)^{-\omega}\int_{t_a^\eps}^{t_h^\eps}s^{\omega-2}\dd s.
\]
For $0<\omega<1$ this is bounded by a constant times
$(\eps/t_a^\eps)(t_a^\eps/t_h^\eps)^\omega$; for $\omega>1$ it is $O(\eps/t_h^\eps)$; for $\omega=1$ it is
$O((\eps/t_h^\eps)\log(t_h^\eps/t_a^\eps))=o(1)$ since $t_a^\eps\ge c_U\eps$. Thus
$\norm{y(t_h^\eps)}\le Ch^2+o_{U,\eps}(1)$ for each fixed $h$. Every predecessor accumulation state consequently satisfies
\begin{equation}\label{eq:positive-relative}
 \norm y\le C_0h^2,\qquad H_{d+1,d+1}\ge h^2/(1+\zeta^2)>0,
\end{equation}
and lies in \eqref{eq:first-section-data} with a deterministic constant $C_0$.

Take any sequence $\eps_n\downarrow0$ and any accumulation point of the outgoing states. For the fixed $h$, compactness also supplies a predecessor limit and a limit of the bounded inter-section durations. On this phase interval, $t$ is within fixed positive factors of $t_b^{\eps_n}$; therefore $t^2+\eps_n/t\to0$ uniformly. The $C^1$ estimate in Lemma~\ref{lem:rescaling} and the finite-time comparison from \citet[Lemma~2.7]{teschl2012ordinary} identify the outgoing limit as the first limiting-equation exit of that predecessor. Transversality identifies the limiting exit as the first crossing of the section.

Lemma~\ref{lem:uniform-first-section} bounds its distance to the same crossing of the connecting solution by a quantity tending to zero with $h$. Now let $h\downarrow0$, after taking the fixed-$h$ subsequential limit. All outgoing accumulation points coincide. Compactness therefore yields convergence of the full $\eps$ family. Once the outgoing state is identified, finite-time continuous dependence gives convergence on every fixed forward or backward phase window on which the connecting solution is defined. Lemma~\ref{lem:connection} provides the next fixed segment.
\end{proof}

\begin{lemma}[Relative compound transfer]\label{lem:relative-transfer}
Let $V$ be any compound matrix with distinguished input-prefix
column $V_0$ and $\norm V_F^2\le(1+\zeta^2)\norm{V_0}^2$. Let $D$ be any positive
diagonal matrix of compatible size and let $F$ be any perturbation of
the same size. With the distinguished column indexed by $0$,
\begin{equation}\label{eq:relative-lower}
 \norm{(V(D+F))_0}
 \ge\bigl[D_{00}-\sqrt{1+\zeta^2}\norm F_\op\bigr]\norm{V_0}.
\end{equation}
\end{lemma}
\begin{proof}(of Lemma~\ref{lem:relative-transfer}).
\label{proof:lem:relative-transfer}
The output column equals $D_{00}V_0+VFe_0$. The reverse triangle inequality gives
\[
 \norm{D_{00}V_0+VFe_0}
 \ge D_{00}\norm{V_0}-\norm V_\op\norm F_\op.
\]
The hypothesis implies
$\norm V_\op\le\norm V_F\le\sqrt{1+\zeta^2}\norm{V_0}$,
which proves \eqref{eq:relative-lower}. The transfer error is proportional to the incoming distinguished-column norm.
\end{proof}

The next statement applies these local estimates to incoming powers of
$\eps$ and gives the outgoing time and the independent powers available
at the next entry.

\begin{lemma}[Propagation of independent scales through an actual passage]
\label{lem:main-relative-growth}
Fix $Q_{d,K}$ with $F_K(Q_{d,K})=0$.
Retain $h_{\max}$ successive unused directions, where
\[
1\le h_{\max}\le\min\{m-d,r-d\}.
\]
Assume the active spectral boundary is strict and that each
retained leading group is separated:
\[
\lambda_d>\lambda_{d+1},
\qquad
\lambda_{d+h}>\lambda_{d+h+1}
\quad
\text{for }1\le h\le h_{\max}\text{ with }d+h<r.
\]

Suppose an actual trajectory enters a sufficiently small
neighbourhood of $Q_{d,K}$ at $t_{\rm in}^\eps$, where
\[
\log t_{\rm in}^\eps
=
(1-x)\log\eps+O_U(1),
\qquad 0\le x<1,
\]
and $\eps/t_{\rm in}^\eps$ is sufficiently small.
Suppose its incoming independent scales satisfy
\[
\log
\frac{V_h(Z^\eps(t_{\rm in}^\eps))}
     {V_{h-1}(Z^\eps(t_{\rm in}^\eps))}
=
\beta_h\log\eps+O_U(1),
\qquad
0<\beta_1\le\cdots\le\beta_{h_{\max}}.
\]
Assume also that each retained leading volume satisfies
\[
\norm{\bigwedge\nolimits^h Z^\eps(t_{\rm in}^\eps)}_F^2
\le
(1+\zeta^2)
V_h(Z^\eps(t_{\rm in}^\eps))^2,
\qquad 1\le h\le h_{\max},
\]
for a sufficiently small fixed $\zeta>0$.

Put $g=g_{d+1}^{(d)}$.
In either of the two cases below, for all sufficiently small $\eps$ the
trajectory stays in a fixed local neighborhood until the stated endpoint,
the leading-volume inequalities persist, and the retained volumes satisfy
\[
\log V_h(Z^\eps(t))
=
\left(\sum_{a=1}^h\beta_a\right)\log\eps
+
\left(\sum_{a=1}^h g_{d+a}^{(d)}\right)
\log\frac{t}{t_{\rm in}^\eps}
+
O_U(1).
\]
The error is uniform over the passage interval. The integrated bounds of
Lemma~\ref{lem:integral} give
\begin{equation}\label{eq:relative-integrated-statement}
 \int_{\text{passage interval}}
 (\norm y+\norm Z_F^2+\norm E_\op)\,\dd\log t\le C_U,
\end{equation}
The interval ends at the outgoing time in the first case and at a
fixed sufficiently small $t_*>0$ in the second case. Constants in these
bounds are independent of the residence length.

If $g>0$ and $x+\beta_1/g<1$, the actual trajectory reaches a
fixed outgoing level $\norm{Z^\eps}_F=h_d>0$ at a first time
$t_{\rm out}^\eps$ satisfying
\[
\log t_{\rm out}^\eps
=
\left(1-x-\frac{\beta_1}{g}\right)\log\eps+O_U(1).
\]
The crossing is transverse.
After shifting its logarithmic time so that the crossing occurs at zero,
the actual Gram converges on each fixed compact time interval to
$Q(v)^\top Q(v)$, where $Q(v)$ is the connecting solution of
Lemma~\ref{lem:connection} from $Q_{d,K}$ to $Q_{d+1,K}$. Each retained exterior column outside its leading prefix,
divided by the norm of the leading column, tends to zero at the crossing
and after transport along a fixed connection segment.
After projection away from the newly occupied direction and
transport to the next neighbourhood, the remaining independent
exponents are
\[
\beta_{a+1}
-
g_{d+a+1}^{(d)}
\frac{\beta_1}{g},
\qquad 1\le a<h_{\max}.
\]
Each retained volume at the next entry is bounded above and below by
positive constants times its outgoing projected volume, uniformly in
$\eps$. The leading-volume bounds hold at this next entry.

If
\[
\beta_1-g(1-x)>0,
\]
fix $\nu>0$ with $\beta_1-(g+\nu)(1-x)>0$.
The neighborhood, the upper bound on $\eps/t_{\rm in}^\eps$, and $t_*$
can be chosen, depending on this $\nu$, so that
\[
 \norm{Z^\eps(t)}_F\le C_U\eps^{\beta_1}
 (t/t_{\rm in}^\eps)^{g+\nu},
 \qquad t_{\rm in}^\eps\le t\le t_*.
\]
In particular, the supremum on this interval tends to zero as a positive
power of $\eps$, its squared norm has integral tending to zero with
respect to $\dd\log t$, and the trajectory remains in the chosen
neighborhood up to $t_*$.
The local residence and integrated bounds also hold when every unused
task eigenvalue is negative, with the same entry-time assumptions and
without retained-volume or exponent conditions. For any incoming normal block in the fixed small neighborhood,
\[
 \norm{Z(t)}_F\le\norm{Z(t_{\rm in}^\eps)}_F
 (t/t_{\rm in}^\eps)^{-1/2},\qquad t_{\rm in}^\eps\le t\le t_*.
\]
Thus the unused component tends to zero at every fixed $t\in(0,t_*]$.
When the normal block is empty, these residence and integrated conclusions
hold with $Z=0$ and require no retained directions.
\end{lemma}
\begin{proof}(of Lemma~\ref{lem:main-relative-growth}).
\label{proof:lem:main-relative-growth}
Put $g=g_{d+1}^{(d)}$.

Existence and time of the actual crossing.

Suppose first that $g>0$ and $x+\beta_1/g<1$.
Start at the entry in the statement and stop at the first outgoing normal
section, tangent boundary, cone boundary, or fixed small $t_*$.
Lemma~\ref{lem:integral} preserves the incoming cones on this stopped
interval. Apply \eqref{eq:local-relative-volume} with $h=1$ to every
subinterval. Choose the local coefficient-error bound so that the $h=1$ estimate gives $(\log V_1)'\ge g/2$. The positive-growth part of
Lemma~\ref{lem:integral} therefore supplies its integrated estimate on
every stopped prefix, with no prior assumption on which boundary is
reached. The tangent convolution, with
the bound $\norm Z_F\le h_d$, prevents a tangent exit once its incoming
ball, $h_d$, $t_*$, and the incoming regularization ratio are chosen
sufficiently small. The choices depend only on the fixed local data.
Equation~\eqref{eq:local-relative-volume} and Lemma~\ref{lem:integral} therefore give,
on this interval,
\[
 \log V_h(Z^\eps(t))
 =\left(\sum_{a=1}^h\beta_a\right)\log\eps
 +\left(\sum_{a=1}^h g_{d+a}^{(d)}\right)
 \log(t/t_{\rm in}^\eps)+O_U(1).
\]
If $g>0$ and $x+\beta_1/g<1$, failure to cross before $t_*$ would give
\[
 \log V_1(Z^\eps(t_*))
 =\{\beta_1-g(1-x)\}\log\eps+O_U(1)\longrightarrow+\infty,
\]
contradicting the stopped normal bound. Thus a first crossing occurs.
There $h_d/\sqrt{1+\zeta^2}\le V_1\le h_d$, and consequently
\begin{equation}\label{eq:wait}
 \log(t_{\rm out}^\eps/t_{\rm in}^\eps)
 =\frac{\beta_1}{g}\log(1/\eps)+O_U(1).
\end{equation}
This proves the outgoing-time formula and gives
$t_{\rm out}^\eps\to0$ and $\eps/t_{\rm out}^\eps\to0$.
The incoming norm tends to zero by the first cone and $\beta_1>0$.
All hypotheses of Lemma~\ref{lem:exit} have now been verified, so it identifies the outgoing Gram and the limit on each fixed phase interval.
The trace-growth inequality makes this crossing transverse.

Use the complementary exterior columns $B_h$ defined in the proof of Lemma~\ref{lem:integral}. By Lemma~\ref{lem:exit}, the time-shifted actual Gram paths converge to the diagonal connecting Gram path on every fixed backward window $[-A,0]$. The exact normal coefficient
$\TF_{22}-I-\TF_{21}S_{11}^{-1}S_{12}$ is continuous in the Gram coordinates, $t^2$, and $\eps/t$: the active block stays invertible and
$\norm{QR_t}_F^2=\tr(R_tSR_t)$.
Its off-diagonal part $E-\diag(E)$ consequently tends uniformly to zero on that window. At a sufficiently small fixed outgoing section the diagonal exterior gaps stay at least half their values at $Q_{d,K}$.

Within the coarse cone, define the ratio
$r_h=\norm{B_h}_F/V_h(Z)$. For a nonempty complementary group, \eqref{eq:dini} with the diagonal part separated gives
\[
 D^+r_h\le-\sigma_{d,h}r_h/2
       +h(1+\zeta)^2\norm{E-\diag(E)}_\op.
\]
Hence
\begin{equation}\label{eq:prefix-alignment}
 r_h(0)\le\zeta e^{-\sigma_{d,h}A/2}
 +h(1+\zeta)^2\int_{-A}^0e^{\sigma_{d,h}u/2}
       \norm{E(u)-\diag(E(u))}_\op\dd u\longrightarrow0,
\end{equation}
where $\eps\downarrow0$ first at fixed $A$, and then $A\to\infty$.
An empty complement has ratio zero identically.

At the outgoing section adjoin $z_{d+1}\ne0$ to the occupied span. For every remaining ordered group $J$, orthogonal elimination gives
\begin{equation}\label{eq:deflation}
 \left\|z_{d+1}\wedge\bigwedge_{l\in J}z_l\right\|
 =\norm{z_{d+1}}
 \left\|\bigwedge_{l\in J}
 \left(I-\frac{z_{d+1}z_{d+1}^\top}{\norm{z_{d+1}}^2}\right)z_l\right\|.
\end{equation}
Indeed, choose the first basis vector along $z_{d+1}$; parallel components disappear from the wedge. The new principal $h$-volume is $V_{h+1}(Z)/V_1(Z)$. Its competing squared coefficients are bounded by the corresponding old sum divided by $V_1(Z)^2$. Since $V_1(Z)\ge h_d/\sqrt{1+\zeta^2}$, both the relative cones and the independent powers survive this projection.

On a fixed connecting segment $[v_a,v_b]$ use the normal coordinates with the first $d+1$ task columns occupied. They stay independent along this compact segment. Let $\Psi_\eps$ be the right fundamental matrix of the complete homogeneous normal equation, from $v_a$ to $v_b$. Then $Z(v_b)=Z(v_a)\Psi_\eps$ exactly. Uniform coefficient convergence and Gronwall's inequality give convergence of these fundamental matrices. For each retained $h$, let
\[
 D=\diag_{\substack{J\subset\{d+2,\ldots,r\}\\|J|=h}}
 \exp\left\{\int_{v_a}^{v_b}\sum_{l\in J}
 \left(\frac{a_K\lambda_l}{\norm{Q(u)C}_F}-1\right)\dd u\right\},
 \qquad F=\wedge^h\Psi_\eps-D.
\]
Here $Q$ is the connecting solution. Polynomial continuity of exterior powers gives $\norm F_\op\to0$. Lemma~\ref{lem:relative-transfer}, applied directly to $\wedge^h Z(v_a)$, gives
\[
 V_h(Z(v_b))\ge
 \left[D_{00}-\sqrt{1+\zeta^2}\norm F_\op\right]V_h(Z(v_a)).
\]
For sufficiently small $\eps$ the bracket is at least $D_{00}/2>0$, independent of the incoming volume. The upper triangle inequality gives the corresponding upper bound. The diagonal ordering in Lemma~\ref{lem:connection} implies $\max_{j\ne0}D_{jj}\le D_{00}$. The outgoing complementary/principal ratio is therefore at most
\[
 \frac{D_{00}\norm{B_h(v_a)}_F/V_h(Z(v_a))
       +\sqrt{1+\zeta^2}\norm F_\op}
 {D_{00}-\sqrt{1+\zeta^2}\norm F_\op}.
\]
The incoming fine-cone ratio and $\norm F_\op$ both tend to zero. Thus the next inlet cones can be made arbitrarily narrow, and every retained volume transfers by positive bounded factors.

Outgoing independent exponents and the final interval.

For the remaining scales, the new principal $h$-volume equals
$V_{h+1}(Z)/V_1(Z)$ by \eqref{eq:deflation}. Subtracting consecutive logarithmic-volume formulas at the outgoing time
therefore gives the exponent
$\beta_{a+1}-g_{d+a+1}^{(d)}\beta_1/g$ for the $a$th remaining
independent scale. The fixed connection multiplies each retained volume
by factors bounded above and below by positive constants, as proved in
the preceding connection estimate. This completes the transfer assertion.

If $\beta_1-g(1-x)>0$, Lemma~\ref{lem:integral} gives uniform decay of $Z$ on the stopped
interval. The cone and tangent estimates therefore extend that interval
to $t_*$. The integrated coefficient bound in that lemma, inserted
in \eqref{eq:local-relative-volume}, gives the stated logarithmic-volume
formula throughout this final interval, together with
\eqref{eq:relative-integrated-statement}. If all unused eigenvalues
are negative, the direct inequality $D^+\norm Z_F\le-\norm Z_F/2$
gives the displayed contraction without exterior cones. Its squared
integral is bounded by a fixed multiple of the incoming squared norm;
the same tangent convolution proves residence and the integrated bound.
An empty normal block satisfies these estimates identically.
\end{proof}

\subsection{Actual clocks and the deterministic entrance}\label{sec:passage-results}
The preceding estimates are now iterated in the finite order prescribed
by the clocks. Each entry carries the independent suppression left by
all earlier stages. The last residence gives convergence to a
fixed-rank empirical boundary solution.
Fix a task under Assumptions~\ref{ass:task} and~\ref{ass:prefix},
a finite $K$, and $b>1$ with $b\notin\mathcal B_{C,K}$,
and write $D=d_K(b)$. The retained orders
$1\le h\le p-d$, label sets $J_{d,h}$, and volumes $V_h$ are defined
in \eqref{eq:prefix-family} and \eqref{eq:volumes}.
In each rank-$d$ residence, evaluate $y,Z,E$ from
Lemma~\ref{lem:main-relative-growth} on $Q^\eps(t)$ in that chart,
and write $Z_d^\eps(t)$ for its projected matrix.
For each retained label $l$, define
\begin{equation}\label{eq:stage-inlet-exponents}
 e_l^{(1)}=(b-1),\qquad
 e_l^{(d)}=((b-1)+x_d)(1-\lambda_l/\lambda_d)
 \quad (2\le d\le D).
\end{equation}
The lemma below identifies these numbers as the actual incoming
independent exponents.

For an index set $I$, $P_{\Ran Q_I}$ denotes orthogonal projection
onto the hidden span of the columns $Q_I$. Given a fixed outgoing
level $h_d>0$, the section at rank $d<D$ is
\begin{equation}\label{eq:rank-transition-section}
 \left\|\left(I_m-P_{\Ran Q^\eps_{\{1,\ldots,d\}}(t)}\right)
 Q^\eps_{\{d+1,\ldots,r\}}(t)\right\|_F=h_d.
\end{equation}
Its first crossing after entry into the rank-$d$ neighborhood is
$t_{d+1}^\eps$. In the orthonormal coordinates of
Lemma~\ref{lem:main-relative-growth}, the left side is
$\norm{Z_d^\eps}_F$.
A transition records an independent component reaching a fixed
size in the rescaled factor. The positive-initialization factor
has its original algebraic rank at every finite time by
Proposition~\ref{main:global}.

\begin{lemma}[Actual transition times and remaining independent scales]
\label{lem:main-clocks}
For the fixed data above, there
are finitely many coordinate neighborhoods and cone widths,
positive outgoing levels $h_d$ for $d<D$, fixed connecting-solution
segments with logarithmic durations $\Delta v_d>0$,
and $t_*>0$, all chosen from the fixed data, such that the
following holds. For each seed in the Gaussian-full event of
Lemma~\ref{lem:gaussian-entry}, there are
$\sigma_0(U)<\infty$ and $C_U<\infty$ for which all
sufficiently small $\eps>0$ have these properties; the upper
cutoff for $\eps$ may depend on $U$.

The first inlet is $t_{{\rm in},1}^\eps=\eps\sigma_0(U)$.
At later stages,
$t_{{\rm in},d}^\eps=e^{\Delta v_{d-1}}t_d^\eps$, and
\begin{equation}\label{eq:inlet-time-bound}
 \left|\log t_{{\rm in},d}^\eps-(1-x_d)\log\eps\right|
 \le C_U,\qquad 1\le d\le D.
\end{equation}
For $d<D$, the residence ends at the first outgoing section
$t_{d+1}^\eps$; for $d=D$, it ends at $t_*$.
Throughout each residence the actual trajectory stays in its
coordinate neighborhood and all required column cones hold.
Its retained volumes satisfy the two-sided estimates
\begin{equation}\label{eq:stage-volume-bound}
 \left|\log V_h(Z_d^\eps(t))
 -\Bigl(\sum_{l\in J_{d,h}}e_l^{(d)}\Bigr)\log\eps
 -\Bigl(\sum_{l\in J_{d,h}}g_l^{(d)}\Bigr)
       \log\frac{t}{t_{{\rm in},d}^\eps}\right|\le C_U,
 \qquad 1\le h\le p-d,
\end{equation}
and its coefficient error satisfies
\begin{equation}\label{eq:stage-error-budget}
 \int_{\text{residence at }d}
 (\norm{y}+\norm{Z_d^\eps}_F^2+
 \norm{E}_\op)\,\dd\log t\le C_U.
\end{equation}
Each completed crossing is transverse and obeys
\begin{equation}\label{eq:main-clock-law}
 \log t_{d+1}^\eps=(1-x_{d+1})\log\eps+O_U(1),
 \qquad 1\le d<D.
\end{equation}
Its Gram matrix, with logarithmic time shifted to put the
crossing at zero, converges to the Gram matrix of the connecting
trajectory of Lemma~\ref{lem:connection}. Projection and
transport along the fixed segment give the stated inlet
conditions at rank $d+1$.
At the final rank, $Z_D^\eps(t)\to0$ at every fixed
$t\in(0,t_*]$. Assertions about retained volumes are vacuous
when $d=p$.
\end{lemma}
\begin{proof}(of Lemma~\ref{lem:main-clocks}).
\label{proof:lem:main-clocks}
The candidate rank sequence is finite. Use $D=d_K(b)$ from the statement. All geometric neighborhoods and outgoing levels are
chosen from the fixed task, width, finite depth and exponent before
fixing a seed. Seed-dependent constants below are uniform as
$\eps\downarrow0$ for that seed.\par

Geometric choices.

The candidate ranks form a finite sequence. Assumption~\ref{ass:prefix}
and Lemma~\ref{lem:gap-equivalence} give the strict boundaries and the
positive exterior gaps required by Lemma~\ref{lem:main-relative-growth}
at every rank in this sequence. Choose the neighborhoods and outgoing
levels supplied by that lemma. By Lemma~\ref{lem:connection}, each
connecting solution reaches the interior of the next chosen neighborhood
after a fixed finite logarithmic duration $\Delta v_d$. Choose these
segments successively, shrinking the preceding outgoing levels when
needed. The local constants depend only on the task, depth, exponent,
and dimensions. Their finite minimum gives a common small $t_*>0$.

Fix a seed in the measurable Gaussian-full event of
Lemma~\ref{lem:gaussian-entry}. Choose its finite $\sigma_0(U)$ so that the
first entry lies inside the chosen rank-one neighborhood, every required
cone bound holds with room to spare, and $1/\sigma_0$ is below all
applicable regularization-ratio bounds.

First entry.

At $t_{{\rm in},1}^\eps=\eps\sigma_0$, the incoming time law has $x_1=0$.
Equation~\eqref{eq:gaussian-volume-entry} gives
\[
 \log\frac{V_h(Z_1^\eps(\eps\sigma_0))}
 {V_{h-1}(Z_1^\eps(\eps\sigma_0))}
 =(b-1)\log\eps+O_U(1),\qquad 1\le h\le p-1.
\]
These exponents are positive because $b>1$.
Equation~\eqref{eq:gaussian-cone-entry} supplies the required leading
volume bounds. Thus the input conditions of
Lemma~\ref{lem:main-relative-growth} hold before its first application.

A completed stage.

Suppose the entry conditions have been established at rank $d<D$.
Apply Lemma~\ref{lem:main-relative-growth} to the established rank-$d$ entry, retaining its $p-d$ required exterior orders with incoming powers $e_{d+h}^{(d)}$. For $d=1$ all these exponents equal $(b-1)$. For $d\ge2$,
\[
 e_{d+h}^{(d)}=((b-1)+x_d)
 \left(1-\frac{\lambda_{d+h}}{\lambda_d}\right)>0.
\]
Their order is nondecreasing in $h$, since the task eigenvalues are in
decreasing signed order. The entry-time bound gives $0\le x_d<1$ and
a positive seed-dependent lower bound for $t_{{\rm in},d}^\eps/\eps$. At the first stage the small
regularization ratio was chosen above; at later stages $x_d>0$ and
$\eps/t_{{\rm in},d}^\eps\to0$.
The definition of a completed candidate is exactly
\[
 g_{d+1}^{(d)}>0,\qquad
 x_d+\frac{e_{d+1}^{(d)}}{g_{d+1}^{(d)}}=x_{d+1}<1.
\]
All hypotheses have now been checked. The lemma gives the first
transverse crossing, its Gram limit, the estimates
\eqref{eq:stage-volume-bound}--\eqref{eq:stage-error-budget}, and
\[
 \log t_{d+1}^\eps=(1-x_{d+1})\log\eps+O_U(1).
\]
In particular $t_{d+1}^\eps\to0$ and
$\eps/t_{d+1}^\eps\to0$ are outputs of this application.

The exponents at the next entry.

At the first completed stage, $x_2=(b-1)/g_2^{(1)}$.
The transfer conclusion of Lemma~\ref{lem:main-relative-growth} gives,
for every remaining retained label $l$,
\begin{align}
 e_l^{(2)}
 &=(b-1)-g_l^{(1)}x_2
 =(b-1)+x_2-\frac{a_K\lambda_l}{\lambda_1}x_2\notag\\
 &=((b-1)+x_2)\left(1-\frac{\lambda_l}{\lambda_2}\right),
 \label{eq:second-entry-exponents}
\end{align}
because $(b-1)+x_2=(a_K\lambda_2/\lambda_1)x_2$.
At a later completed stage, the defining time identity is
\[
 ((b-1)+x_d)\left(1-\frac{\lambda_{d+1}}{\lambda_d}\right)
 =(x_{d+1}-x_d)\left(\frac{a_K\lambda_{d+1}}{s_{d,K}}-1\right).
\]
Rearranging gives
\[
 \frac{(b-1)+x_d}{\lambda_d}
 +\frac{a_K(x_{d+1}-x_d)}{s_{d,K}}
 =\frac{(b-1)+x_{d+1}}{\lambda_{d+1}}.
\]
For a remaining label, the transfer formula therefore becomes
\begin{align}
 e_l^{(d+1)}
 &=e_l^{(d)}-g_l^{(d)}(x_{d+1}-x_d)\notag\\
 &=(b-1)+x_{d+1}-\lambda_l\left(
 \frac{(b-1)+x_d}{\lambda_d}
 +\frac{a_K(x_{d+1}-x_d)}{s_{d,K}}\right)\notag\\
 &=((b-1)+x_{d+1})
   \left(1-\frac{\lambda_l}{\lambda_{d+1}}\right)>0.
 \label{eq:exponent-update}
\end{align}
The identity $p-(d+1)=(p-d)-1$ ensures that the previous stage
retained all volumes required after the new projection. The lower and
upper transfer bounds and the convergence of the complementary column
ratios, both stated in Lemma~\ref{lem:main-relative-growth}, give the
next entry's volume and cone conditions. The fixed segment adds
$\Delta v_d$ to logarithmic time, so
$t_{{\rm in},d+1}^\eps=e^{\Delta v_d}t_{d+1}^\eps$ satisfies
\eqref{eq:inlet-time-bound}. This completes the induction step.

The final interval.

If $D<p$, apply the final case of Lemma~\ref{lem:main-relative-growth} to the final entry with exponent $e_{D+1}^{(D)}$ and clock $x_D$. The stopping rule and
noncriticality give
\[
 e_{D+1}^{(D)}-g_{D+1}^{(D)}(1-x_D)>0.
\]
For a positive rate this is equivalent to the next candidate being
larger than one. For a nonpositive rate it follows from the positivity
of $e_{D+1}^{(D)}$. The lemma supplies residence up to $t_*$, the
integrated bound, and decay of the normal component.
If $D=p=m$, the unused-hidden block has zero rows and is empty.
If $D=p<m$, the identity $p=\min\{m,n_+(C)\}$ implies
$p=n_+(C)$; nonsingularity of $C$ then makes every unused task
eigenvalue negative. This includes $p=r<m$, where there are no unused
task columns. Apply the respective
last case of the same lemma. Its contraction estimate gives
$Z_D^\eps(t)\to0$ at every fixed $t>0$.

There are finitely many applications. Taking the smallest of their
seed-dependent cutoffs and the largest of their constants gives one
range of $\eps$ and a common $C_U$ for the entire sequence.
\end{proof}

The bound $O_U(1)$ holds with the seed and model parameters fixed.
The clock law states equivalently that
$\log(t_{d+1}^\eps/\eps)/\log(1/\eps)\to x_{d+1}$.
Each completed candidate lies in $(0,1)$, so its transition
time tends to zero while $t_{d+1}^\eps/\eps$ tends to infinity.

\label{def:empirical-entrance}
The next lemma identifies an entrance from the decay of the unused
component. Its conditions concern a fixed rank and apply in any positive
factor dimensions. They are also used for the rank-one conclusion in
Proposition~\ref{prop:rank-one-loss}.

\begin{lemma}[Deterministic entrance at a fixed rank]
\label{lem:deterministic-entrance}
Consider a spanning empirical quadratic task, finite $K$, and a nonzero
matrix $Q_{d,K}$ with $F_K(Q_{d,K})=0$ and
$\Ran(Q_{d,K}^\top)=\operatorname{span}\{v_1,\ldots,v_d\}$.
Assume $C$ is symmetric and nonsingular, its eigenvalues are ordered
by signed value, and $1\le d\le\min\{m,r\}$,
$\lambda_d>\lambda_{d+1}$ when $d<r$, and either $K\ge1$ or $d=1$.
There is a local rank-$d$ solution $\overline W_{K,d}$ of
\[
 \dot W=P_K(G(W)/\norm{G(W)}_F),\qquad W(t)/t\longrightarrow Q_{d,K},
\]
unique in this fixed-rank class up to a constant left orthogonal
transformation. For some $t_*>0$ it satisfies
\[
 \overline W_{K,d}(t)=tQ_{d,K}+O(t^3),\qquad
 \dot{\overline W}_{K,d}(t)=Q_{d,K}+O(t^2).
\]

Let $W^\eps$ be solutions of \eqref{eq:flow} with
$\eta_\eps=c_\eta\eps$ and $W^\eps(0)\to0$.
Suppose there are $t_a^\eps\to0$ with $t_a^\eps\ge\eps$ such that
the rescaled factor $W^\eps(t)/t$ remains in a fixed sufficiently small Gram
neighborhood of $Q_{d,K}$ for $t_a^\eps\le t\le t_*$, and its
projected component satisfies $Z^\eps(t)\to0$ at every fixed
$t\in(0,t_*]$. Then, locally uniformly for $t\in(0,t_*)$,
\[
 \inf_{O\in\Ogroup(m)}
 \norm{OW^\eps(t)-\overline W_{K,d}(t)}_F\longrightarrow0.
\]
At every sufficiently small fixed predictor radius $\rho>0$, the actual
first hits exist, their times converge to the first hit of
$\overline W_{K,d}$, and their predictors converge to its rank-$d$
predictor on that section. The neighborhood and $t_*$ may be chosen
from the fixed task, $K$, and $Q_{d,K}$.
\end{lemma}
\begin{proof}(of Lemma~\ref{lem:deterministic-entrance}).
\label{proof:lem:deterministic-entrance}
Use the Gram coordinates fixed before Lemma~\ref{lem:integral} near $Q_{d,K}$. At $H=0$, the reconstruction
$S_{22}=S_{21}S_{11}^{-1}S_{12}$ gives a PSD Gram matrix of rank $d$ while $S_{11}\succ0$. With zero denominator regularization its exact tangent equation is
\[
 y'=J_dy+R_d(y,t^2),\qquad
 \norm{R_d(y,t^2)}\le C(\norm y^2+t^2),
\]
where $R_d$ includes the full empirical term. On the fixed small chart,
\[
 \norm{R_d(y,t^2)-R_d(z,t^2)}
 \le C(\norm y+\norm z+t^2)\norm{y-z}.
\]
Choose the tangent norm and common contraction rate $\omega$ of Lemma~\ref{lem:integral} with sufficient margin for this derivative bound. The linear kernel then satisfies $\norm{(t/s)^{J_d}}\le(s/t)^\omega$ for $0<s\le t$.

For any bounded solution tending to zero, multiplication by $t^{-J_d}$ gives derivative $t^{-1-J_d}R_d(y(t),t^2)$. Its norm is bounded by $Ct^{\omega-1}(\norm{y(t)}^2+t^2)$, which is integrable at zero, and
$\norm{t^{-J_d}y(t)}\le t^\omega\sup_{0<s\le t_*}\norm{y(s)}\to0$.
Integration directly from zero gives the boundary equation
\begin{equation}\label{eq:volterra}
 y_*(t)=\int_0^t(t/s)^{J_d}R_d(y_*(s),s^2)\frac{\dd s}s.
\end{equation}

On the Banach space of continuous functions on $(0,t_*]$ with finite norm $\sup t^{-2}\norm{y(t)}$, use the radius $R=2C/(\omega+2)$. The linear forcing contributes at most $Ct^2/(\omega+2)$; on this ball the quadratic contribution is at most $CR^2t^4/(\omega+4)$. The difference of two images in the weighted norm is bounded by $C(2R+1)t_*^2$ times their distance. Choose $t_*$ small enough to preserve the ball and make this coefficient less than one. The contraction principle \citep[Theorem~2.1]{teschl2012ordinary} gives a unique $y_*=O(t^2)$.

Every solution tending to zero satisfies the same integral equation. Taking suprema up to $t$ gives
\[
 \sup_{0<s\le t}\norm{y(s)}
 \le C\left(\sup_{0<s\le t}\norm{y(s)}\right)^2+Ct^2.
\]
For small $t$ the quadratic term is absorbed by half the left side. Hence $y=O(t^2)$, and substitution into the integral equation puts it in the same weighted ball after shrinking the interval. This proves uniqueness in the asserted class.

Let $S(t)$ be the resulting rank-$d$ Gram matrix. At the existing endpoint $t_*>0$, choose a physical factor with Gram $t_*^2S(t_*)$, and denote its solution by $\overline W_{K,d}$. Solve backwards under the zero-regularization normalized equation. Its rescaled Gram solves the same equation as $S(t)$ and therefore agrees with it. For small positive $t$, $C-t^2\Lop(S(t))$ is invertible and $\norm{G(\overline W_{K,d}(t))}_F$ is bounded below by a positive constant times $t$. The solution is in the smooth domain; boundedness on every closed interval away from zero gives continuation throughout $(0,t_*]$.

The factor
\[
 \begin{pmatrix}S_{11}^{1/2}&S_{11}^{-1/2}S_{12}\\0&0\end{pmatrix}
\]
has Gram matrix $S$ and depends smoothly on $y$. Thus
$t^{-1}\overline W_{K,d}(t)$ has distance $O(t^2)$ from
$\{OQ_{d,K}:O\in\Ogroup(m)\}$. The rescaled vector field vanishes on this set at $t=0$, is locally Lipschitz, and has empirical perturbation $O(t^2)$. Therefore
$\norm{\partial_v(t^{-1}\overline W_{K,d}(t))}_F=O(t^2)$.
Integration from $v=-\infty$ gives a limit in that orthogonal class and an $O(t^2)$ difference. Choose its constant left orthogonal representative so that
\begin{equation}\label{eq:entrance-derivative}
 \overline W_{K,d}(t)=tQ_{d,K}+O(t^3),\qquad
 \dot{\overline W}_{K,d}(t)=Q_{d,K}+O(t^2).
\end{equation}
Factors with the same Gram at a positive time differ by a left orthogonal transformation. Equivariance and factor uniqueness keep this transformation constant along the solutions, proving the fixed-rank uniqueness.

For the actual family, smooth dependence of the exact Gram field on $H$ and $\eps/t$ gives
\[
 (y^\eps)'=J_dy^\eps+R_d(y^\eps,t^2)
 +O(\norm{Z^\eps}_F^2+\eps/t).
\]
Subtracting the constructed equation and absorbing its small Lipschitz remainder in the common tangent norm gives
\begin{equation}\label{eq:final-comparison}
 D^+\norm{y^\eps-y_*}
 \le-\omega\norm{y^\eps-y_*}+C\norm{Z^\eps}_F^2+C\eps/t.
\end{equation}
Consequently,
\begin{align*}
 \norm{y^\eps(t)-y_*(t)}
 \le{}&C(t_a^\eps/t)^\omega\norm{y^\eps(t_a^\eps)-y_*(t_a^\eps)}\\
 &+Ct^{-\omega}\int_{t_a^\eps}^t s^{\omega-1}\norm{Z^\eps(s)}_F^2\dd s
 +C\eps t^{-\omega}\int_{t_a^\eps}^t s^{\omega-2}\dd s.
\end{align*}
On any fixed compact positive-time interval, the initial term vanishes because its mismatch is bounded and $t_a^\eps\to0$. Extend $Z^\eps$ by zero below $t_a^\eps$. Dominated convergence with majorant $Cs^{\omega-1}$ makes the normal integral tend to zero uniformly there. Since $t_a^\eps\ge\eps$, the regularization term is $O(\eps^\omega)$ for $0<\omega<1$, $O(\eps\log(1/\eps))$ for $\omega=1$, and $O(\eps)$ for $\omega>1$. Thus $y^\eps\to y_*$ locally uniformly. The exact normal equation has bounded coefficients on every compact positive-time interval; equicontinuity and pointwise convergence of $Z^\eps$ give uniform convergence there. Reconstructing $S$ from $y$ and $H=(Z^\eps)^\top Z^\eps$ proves locally uniform Gram convergence.

The factor conclusion follows by compactness: any sequence violating it has convergent subsequences of bounded factors at positive times with equal limiting Grams. Their limits differ by a left orthogonal transformation, contradicting the assumed positive distance from that orthogonal class.

Equation~\eqref{eq:entrance-derivative} yields
\begin{equation}\label{eq:radius-derivative}
 \frac{\dd}{\dd t}\norm{\overline W_{K,d}(t)^\top\overline W_{K,d}(t)}_F
 =2t\norm{Q_{d,K}^\top Q_{d,K}}_F+O(t^3)>0.
\end{equation}
Every sufficiently small fixed $\rho>0$ has a unique local first crossing $t_\rho$. The speed bound is $\norm{\dot W^\eps}_F\le\sqrt{\min\{m,r\}}$. For sufficiently small $\eps$, $\norm{W^\eps(0)}_F\le\sqrt\rho/4$, so the actual predictor norm is at most $\rho/4$ on
\[
 0\le t\le\frac{\sqrt\rho}{4\sqrt{\min\{m,r\}}}.
\]
The deterministic entrance has the same speed bound from zero, giving
$t_\rho\ge\sqrt\rho/\sqrt{\min\{m,r\}}$.
Fix any sufficiently small open interval containing $t_\rho$ with closure in $(0,t_*)$. Local uniform convergence puts the actual norm below $\rho$ at its left endpoint and above $\rho$ at its right endpoint. On the compact interval from the explicit early cutoff to the left endpoint, the limiting norm has a positive margin below $\rho$. Uniform convergence preserves it. Thus the actual first hit eventually belongs to every such interval. Its time tends to $t_\rho$, and uniform Gram convergence gives its predictor limit.
\end{proof}

\begin{lemma}[Isometric embeddings of a fixed-rank entrance]\label{lem:entrance-embedding}
Let $V(t)\in\R^{d\times r}$ be an entrance satisfying
Lemma~\ref{lem:deterministic-entrance}, and let
$E\in\R^{m\times d}$ satisfy $E^\top E=I_d$, with $m\ge d$.
Then $EV(t)$ is the $m$-row entrance with boundary $EQ_{d,K}$.
Its Gram matrix and its entire continuation Gram matrix are the same
as those of $V$. In particular, whenever two widths have Gaussian
entrances of the same rank $d$ and the same boundary Gram matrix,
their selected continuation predictors and terminal losses agree.
\end{lemma}
\begin{proof}(of Lemma~\ref{lem:entrance-embedding}).
\label{proof:lem:entrance-embedding}
Since $E^\top E=I_d$, for every $V\in\R^{d\times r}$ and every
$X\in\R^{d\times r}$,
\[
 \begin{aligned}
 (EV)^\top EV&=V^\top V,\\
 G(EV)&=2EVR(V^\top V)=EG(V),\\
 \norm{EX}_F^2&=\tr(X^\top E^\top EX)=\norm X_F^2.
 \end{aligned}
\]
We prove $P_k(EX)=EP_k(X)$ by induction. It holds for $k=0$.
If it holds for $k$, the cubic term at the next step is
\[
 (EP_k(X))(EP_k(X))^\top EP_k(X)
 =EP_k(X)P_k(X)^\top E^\top EP_k(X)
 =EP_k(X)P_k(X)^\top P_k(X).
\]
Substitution in \eqref{eq:cubic-definition} proves the induction step.
Both the zero-stabilization physical field where $G\ne0$ and the
continuation field $H_{K,\eta}(G)$ therefore commute with multiplication
by $E$. The assertion also holds at $G=0$ because both continuation
fields then vanish. Thus $EV(t)$ satisfies the full $m\times r$
equation, has rank $d$, and has the stated boundary.

Lemma~\ref{lem:deterministic-entrance} identifies it with the local
$m$-row entrance up to a constant left orthogonal transformation.
Positive-time uniqueness extends that identification through the
whole continuation. Left orthogonal transformations preserve the
Gram matrix, and Proposition~\ref{main:global} supplies the terminal
factor limit. Finally, any two factors of the same rank and Gram
matrix can be represented as isometric embeddings of one $d$-row
factor: map the span of its columns isometrically to the span of the
columns of the other factor and extend orthonormal bases. Applying
the identities just proved and uniqueness gives the last assertion.
The comparison identifies selected continuation Grams across widths.
\end{proof}

Lemma~\ref{lem:main-clocks} verifies these conditions for the rank
$d_K(b)$ computed from Gaussian initialization.

\begin{proof}(of Theorem~\ref{thm:passage}).
\label{proof:thm:passage}
Fix the task, $K$, $b$, and $c_\eta$ as in the statement, and set
$d=d_K(b)$. Lemmas~\ref{lem:gram-zeros} and~\ref{lem:add-zero}
provide $Q_{d,K}$. Assumption~\ref{ass:prefix} gives the strict final
spectral boundary when $d<r$; at $K=0$ the procedure stops at $d=1$.
Thus Lemma~\ref{lem:deterministic-entrance} supplies the local rank-$d$
solution and its uniqueness and expansion.

For a seed in the measurable Gaussian-full event of
Lemma~\ref{lem:gaussian-entry}, apply Lemma~\ref{lem:main-clocks}.
Its final incoming time tends to zero. At rank one choose the initial-layer cutoff at least one; at every later rank the time law has $x_d>0$, so
$t_{{\rm in},d}^\eps\ge\eps$ for all sufficiently small $\eps$.
That lemma gives residence in the chosen final neighborhood and
$Z_d^\eps(t)\to0$ at each fixed small positive time. The neighborhoods
may be chosen within those of Lemma~\ref{lem:deterministic-entrance}.
Finally $D_{\eps,b}U\to0$ for every fixed seed.
All conditions of the deterministic lemma are now verified. Its
conclusions are \eqref{eq:full-limit} and \eqref{eq:entrance-main}.

The task and the final zero determine the constructed Gram solution,
so the limit is independent of the seed and has rank $d$.
Proposition~\ref{prop:clock-domain} makes $\mathcal B_C$ at most countable. Thus almost every $b>1$ lies outside it. For each such fixed $b$, the intersection over all finite integer depths is a measurable Gaussian-full set, and all their fixed-depth conclusions hold there. The time intervals and initialization cutoffs retain their stated
dependence on depth and seed.
\end{proof}

\subsection{Dependence on initialization and depth}\label{app:depth-monotonicity}
The algebraic identity \eqref{eq:staircase} separates the initialization
exponent from the spectral coefficients. For every already existing
support of rank at least two and finite $K\ge1$, its scale $s_{I,K}$
increases with $K$, while $s_{I,K}/a_K$ decreases, and the support
continues to exist. Thus positive unused rates increase and finite
$c_{d,K}$ with $d\ge2$ decrease. These statements imply
Proposition~\ref{prop:rank-monotone}.
\begin{proof}(the identity~\eqref{eq:staircase}).
\label{proof:prop:rank-staircase}
Substitute $x_d=(b-1)c_{d,K}$ in
\eqref{eq:first-clock}--\eqref{eq:clock}. The common factor
$b-1$ cancels, leaving exactly \eqref{eq:tau-clock}.
At each candidate with a positive addition rate, induction therefore
identifies the computed time fraction as $x_d=(b-1)c_{d,K}$.
A nonpositive addition rate assigns $+\infty$ to that candidate in
both sets of time formulas. The condition $x_d<1$ is precisely $(b-1)c_{d,K}<1$, proving
\eqref{eq:staircase}. In particular the second direction enters precisely
when $1<b<a_K\lambda_2/\lambda_1$; this interval is empty when the upper
endpoint is at most one.
\end{proof}

\begin{proof}(of Proposition~\ref{prop:rank-monotone}).
\label{proof:prop:rank-monotone}
For $0<s<1$, write the scalar iteration as
\[
 p_{K+1}(s)=\frac32p_K(s)\left(1-\frac{p_K(s)^2}{3}\right).
\]
Since $0<p_K(s)<1$, this implies $p_{K+1}(s)>p_K(s)$ and
\[
 q_{K+1}(s^2)>q_K(s^2),\qquad
 \frac{q_{K+1}(s^2)}{a_{K+1}}
 =\frac{q_K(s^2)}{a_K}\left(1-\frac{p_K(s)^2}{3}\right)
 <\frac{q_K(s^2)}{a_K}.
\]
First fix an existing support $I$ with $|I|\ge2$ and $K\ge1$.
All its normalized squared singular values lie in $(0,1)$.
Dividing the positive equations in \eqref{eq:zero-system} gives
\[
 \sum_{a\in I}q_K^{-1}(s_{I,K}/\lambda_a)=1,
 \qquad
 \max_{a\in I}\lambda_a<s_{I,K}<a_K\min_{a\in I}\lambda_a.
\]
The strict lower inequality follows because if the scale equalled the
largest active eigenvalue, one inverse term would be one and all
other terms would be positive.

At the upper endpoint $(a_{K+1}/a_K)s_{I,K}$ the inequalities for the normalized polynomials imply
\[
 q_{K+1}^{-1}\!\left(\frac{a_{K+1}s_{I,K}}{a_K\lambda_a}\right)
 <q_K^{-1}(s_{I,K}/\lambda_a).
\]
All displayed inverse arguments lie in their domains, by the strict upper bound on $s_{I,K}$. Their new sum is less than one. At the lower endpoint $\max_{a\in I}\lambda_a$, the largest-eigenvalue term of the new inverse sum equals one and all other terms are positive. The same old upper bound ensures that this endpoint is within the domain. Continuity and strict decrease give a unique new root $s_{I,K+1}$ between these endpoints, constructing the same support at depth $K+1$.

At the old unscaled root $s_{I,K}$, the strict inequality $q_{K+1}>q_K$ makes every new inverse term larger than its old counterpart. The new sum there is greater than one. Combining the two comparisons gives
\[
 s_{I,K}<s_{I,K+1}<\frac{a_{K+1}}{a_K}s_{I,K}.
\]
Iterating proves both comparisons for all larger finite depths.
At the leading rank-one support, $s_{1,K}=\lambda_1$ for all $K$,
and $s_{1,K}/a_K$ decreases directly.

The candidate process at $K=0$ stops at rank one, because
$\lambda_2/\lambda_1-1<0$. At every later depth, for each support
that already existed, the positive unused rates
$a_K\lambda_l/s_{I,K}-1$ increase. An existing addition remains
possible, and a previously nonpositive rate may become positive.
The first finite coefficient $c_{2,K}$ decreases with $K$.
In every subsequent update in \eqref{eq:tau-clock}, the map
\[
 c\longmapsto c+
 \frac{(1+c)(1-\lambda_{d+1}/\lambda_d)}{g_{d+1}^{(d)}}
\]
is strictly increasing in $c\ge0$ and decreasing in the positive
rate denominator. Induction over the already existing supports
therefore makes all finite coefficients $c_{d,K}$ decrease with
$K$; newly available supports can replace infinite coefficients by
finite ones. Equation~\eqref{eq:staircase} proves the algebraic
rank monotonicity. At depths satisfying the pointwise conditions of
Theorem~\ref{thm:passage}, the same integers are the actual
first-radius entrance ranks, which proves the last assertion.
\end{proof}

%% file: appendices/comparison.tex
\section{Comparison and All-Time Path Separation}\label{app:comparison}
We first identify the gradient-flow comparison predictor. Entry sets then
allow the selected continuation to be compared at any sufficiently small
section. Strict loss and equal loss have separate tail estimates, which
are combined with the finite-time comparison to give all-time statements.

\subsection{Gradient-flow comparison and its stationary endpoint}\label{app:glrl}
The precise selection law used in
Proposition~\ref{main:gd} is as follows. For any family $A_\eps\succeq0$
tending to zero, consider
\begin{equation}\label{eq:gf-predictor}
 \frac{\dd M_\eps}{\dd s}
 =-2\{M_\eps\nabla\Phi(M_\eps)+\nabla\Phi(M_\eps)M_\eps\},
 \qquad M_\eps(0)=A_\eps.
\end{equation}
If some fixed $c>0$ satisfies
\begin{equation}\label{eq:gf-alignment}
 \langle A_\eps,v_1v_1^\top\rangle\ge c\norm{A_\eps}_F>0
 \quad\text{for every sufficiently small }\eps,
\end{equation}
then for every fixed $s\in\R$,
\begin{equation}\label{eq:gf-stage-limit}
 M_\eps\!\left(\frac1{4\lambda_1}
 \log\frac1{\langle A_\eps,v_1v_1^\top\rangle}+s\right)
 \longrightarrow\mathcal M^{\rm G}(s).
\end{equation}
The time $s$ is ordinary gradient-flow time. The application below
verifies the complete-flow and leading-projection requirements and the
normalization corresponding to \citet[Theorem~5.6]{li2021greedy}.
\begin{proof}(of Proposition~\ref{main:gd}).
\label{proof:main:gd}

Differentiating $M=W^\top W$ under $\dd W/\dd s=-2W\nabla\Phi(W^\top W)$ gives \eqref{eq:gf-predictor}. Its derivative at $M=0$ is $H\mapsto2(CH+HC)$. In the orthonormal symmetric basis formed from eigenvectors of $C$, the eigenvalues are $2(\lambda_a+\lambda_b)$, $a\le b$. The largest is the simple eigenvalue $4\lambda_1$, and its right eigenvector and left projection are $v_1v_1^\top$ and $H\mapsto\langle H,v_1v_1^\top\rangle$. Thus \eqref{eq:gf-alignment} is exactly the positive-projection hypothesis in Lemma~\ref{lem:local-exit-background}, and its time shift is the one in \eqref{eq:gf-stage-limit}.

The predictor field is analytic. For each $A\succeq0$, choose an initial factor with Gram matrix $A$. Along its gradient equation,
\[
 \frac{\dd}{\dd s}\Phi(W^\top W)
 =-4\norm{W\nabla\Phi(W^\top W)}_F^2\le0.
\]
Compactness of the PSD sublevels bounds the predictor and hence the factor, because $\norm W_F^2=\tr(W^\top W)$. \citet[Theorems~2.2, 2.9, 2.11, 2.13 and Corollary~2.15]{teschl2012ordinary} therefore gives a global forward factor solution and a PSD predictor solution. Choose a compact PSD sublevel containing $0$ and all sufficiently small $A_\eps$. Their forward paths stay there by the displayed loss decrease. Multiply the predictor field by a smooth cutoff equal to one on a neighborhood of this set and supported in a compact subset of the full symmetric space. The resulting $C^2$ field is complete, has the same derivative at zero, and agrees with every relevant forward path. Lemma~\ref{lem:local-exit-background} now proves \eqref{eq:gf-stage-limit}. This is the application underlying \citet[Theorem~5.6]{li2021greedy}, with their matrix loss equal to $2\Phi$.

For the initial values $\alpha v_1v_1^\top$, all predictors have rank at most one, so their common limiting trajectory has rank at most one. The nonconstant departure in Lemma~\ref{lem:local-exit-background} selects its positive leading direction. Since zero is a stationary solution, uniqueness prevents this nonconstant trajectory from attaining zero at any finite time; its rank is therefore one throughout. At a positive point of this path, choose a one-row factor. The one-row gradient equation gives the same predictor by uniqueness. Its forward path is bounded because $\Phi$ has compact PSD sublevels. The general analytic-curve assertion of
Proposition~\ref{main:global} applies to this bounded one-row gradient
flow, which satisfies the three inequalities \eqref{eq:descent} with
$a_0=1$. Consequently the factor and its predictor converge to a stationary limit. On a nonconstant segment the loss is strictly below $\Phi(0)$ and decreases thereafter, so the limiting predictor is nonzero. Closedness of the rank-at-most-one set then gives rank one. Stationarity of the one-row factor gives $M_1R(M_1)=0$ in the empirical setting.

The empirical loss in Lemma~\ref{lem:empirical-identities} and compactness in Proposition~\ref{main:global} supply the hypotheses used in the preceding argument. We now apply that argument to the prescribed Gaussian initialization.

For the actual initialization, write $u_a^\top$ for row $a$ of $U$. Then
\[
 A_\eps=U^\top D_{\eps,b}^2U=\eps^2u_1u_1^\top+\eps^{2b}\sum_{a=2}^m u_au_a^\top,
 \qquad \eps^{-2}A_\eps\longrightarrow u_1u_1^\top.
\]
Since $b>1$, $\norm{u_1u_1^\top}_F=\norm{u_1}^2$, and a Gaussian $u_1$ has $u_1^\top v_1\ne0$ almost surely,
\begin{equation}\label{eq:gf-seed}
 \frac{\langle A_\eps,v_1v_1^\top\rangle}{\norm{A_\eps}_F}
 \longrightarrow\frac{(u_1^\top v_1)^2}{\norm{u_1}^2}>0.
\end{equation}
For each such fixed seed, half the positive limit gives a constant $c$ in \eqref{eq:gf-alignment}. The time-shifted convergence also implies the actual $K=0$ path approaches $M_1$ arbitrarily closely as initialization vanishes: first choose a finite $s$ with $\mathcal M^{\rm G}(s)$ close to $M_1$, then take $\eps$ small in \eqref{eq:gf-stage-limit}. Lemma~\ref{lem:time-change} preserves the infimum of distance over the path.
\end{proof}

\begin{corollary}[Negative curvature at the comparison factor]\label{cor:comparison-saddle}
Under Assumption~\ref{ass:comparison}, every factor
$W_1\in\R^{m\times r}$ with $W_1^\top W_1=M_1$ is stationary
for $f$ and has a unit Frobenius direction $X$ satisfying
\[
 \langle X,\nabla^2f(W_1)[X]\rangle
 =-2\lambda_{\max}(R_1)<0.
\]
Thus the comparison factor is a strict saddle in every hidden
width $m\ge2$ considered here.
\end{corollary}
\begin{proof}(of Corollary~\ref{cor:comparison-saddle}).
\label{proof:cor:comparison-saddle}
Proposition~\ref{main:gd} gives $M_1R_1=0$. Every factor $W_1$ of $M_1$ satisfies
$\norm{W_1R_1}_F^2=\tr(R_1M_1R_1)=0$, so $G(W_1)=0$.
Its rank is one. Since $m\ge2$, choose the already quantified $X$ as the outer product of any unit vector in $\ker W_1^\top$ and any unit eigenvector of $R_1$ for $\lambda_{\max}(R_1)$. Then
\[
 \norm X_F=1,\qquad W_1^\top X=0,\qquad
 \tr(XR_1X^\top)=\lambda_{\max}(R_1).
\]
For every real $s$, the perturbed predictor is $M_1+s^2X^\top X\succeq0$, and the quadratic loss identity gives
\[
 f(W_1+sX)-f(W_1)
 =-\lambda_{\max}(R_1)s^2
 +\frac{s^4}{2}\langle X^\top X,\Lop(X^\top X)\rangle.
\]
The second derivative at zero is the claimed negative value. The variational characterization of the least Hessian eigenvalue proves the strict-saddle assertion. The same display supplies a PSD predictor of loss strictly below $\phi_1$ for sufficiently small nonzero $s$.
\end{proof}

\subsection{Entry sets and section-independent terminal loss}\label{sec:entry-loss}
For the continuation \eqref{eq:continuation-flow}, put
$\ell_\infty(Z)=\lim_{\tau\to\infty}f(\varphi_K(\tau,Z))$ at the fixed
finite depth $K$. Compactness and monotonicity suffice for the maximum
terminal loss on a compact set to be attained. Finite-time comparison
then connects entry sets at different small sections.
\begin{lemma}[Maximum terminal loss on a compact set]\label{lem:terminal-maximum}
For any nonempty compact set $E\subset\R^{m\times r}$, the maximum
$\max_{Z\in E}f(\varphi_K(T,Z))$ exists at every $T\ge0$, is nonincreasing
in $T$, is bounded below by zero, and satisfies
\begin{equation}\label{eq:terminal-maximum}
 \lim_{T\to\infty}\max_{Z\in E}f(\varphi_K(T,Z))
 =\max_{Z\in E}\ell_\infty(Z).
\end{equation}
\end{lemma}
\begin{proof}(of Lemma~\ref{lem:terminal-maximum}).
\label{proof:lem:terminal-maximum}
Proposition~\ref{main:global} gives global existence and decreasing loss. At each finite time, continuous dependence and compactness of $E$ give attainment of the maximum in the statement. These maxima are nonincreasing and nonnegative, so their limit exists.

For each $T\ge0$, the closed subset of $E$ displayed below is nonempty: its threshold is no larger than the maximum at time $T$.
\[
 \left\{Z\in E:
 f(\varphi_K(T,Z))\ge
 \lim_{s\to\infty}\max_{Y\in E}f(\varphi_K(s,Y))\right\}.
\]
These closed subsets are nested as $T$ increases. Every finite subcollection has nonempty intersection, because its intersection is the member corresponding to the largest time in that subcollection. Compactness of $E$ therefore gives
\[
 \bigcap_{T\ge0}
 \left\{Z\in E:
 f(\varphi_K(T,Z))\ge
 \lim_{s\to\infty}\max_{Y\in E}f(\varphi_K(s,Y))\right\}\ne\varnothing.
\]
For every element of this intersection, passing to the limit along its own trajectory shows that its terminal loss is at least the limit on the right. Conversely, for every $Z\in E$ and every $T\ge0$,
\[
 \ell_\infty(Z)\le f(\varphi_K(T,Z))
 \le\max_{Y\in E}f(\varphi_K(T,Y)).
\]
Taking the infimum over $T$ gives the reverse inequality for every terminal loss. Hence the limit of the finite-time maxima equals the maximum terminal loss, and this maximum is attained.
\end{proof}

\begin{definition}[Entry set and terminal-loss test]\label{def:entry-loss}
Fix a finite depth $K$, an exponent $b$, and a seed $U$ for which the first
hits exist at every sufficiently small fixed radius for all sufficiently
small $\eps$. At any such radius define
\begin{equation}\label{eq:entryset}
 E_\rho(K,b,U):=\{Z:\ W_K^{\eps_n,b}(T_\rho^{\eps_n};U)\to Z
                   \text{ for some }\eps_n\downarrow0\},
\end{equation}
and its terminal-loss test by
\begin{equation}\label{eq:entry-loss}
 \mathcal J_\rho(K,b,U):=\lim_{T\to\infty}
   \max_{Z\in E_\rho(K,b,U)}f(\varphi_K(T,Z)).
\end{equation}
Choose a radius bound satisfying
\begin{equation}\label{eq:small-ball}
 0<\bar\rho<\min\left\{
 \frac{\sigma_{\min}(C)}{2n^{-1}\sum_l\norm{x_l}^4},
 \frac{\norm{M_1}_F}{2}\right\}.
\end{equation}
Here $\sigma_{\min}(C)>0$ is the smallest singular value of $C$.
The minimum depth passing this test is
\begin{equation}\label{eq:Kb}
 \kloss(b):=\min\{K\in\N_0:\mathcal J_\rho(K,b,U)<\phi_1
                 \text{ for almost every }U\}.
\end{equation}
Eligible depths have almost-sure first hits at every sufficiently small
fixed section. The test uses any such radius below $\bar\rho$;
Lemma~\ref{lem:section-independence} states its independence of this choice.
Almost-sure statements refer to the Gaussian seed law, and
$\min\varnothing=+\infty$.
\end{definition}
\begin{lemma}[Well-defined terminal-loss test]\label{lem:section-independence}
Under Definition~\ref{def:entry-loss}, each entry set $E_\rho(K,b,U)$ is
nonempty and compact. If $0<\rho_1<\rho_2<\bar\rho$ and the inner first
hits exist for all sufficiently small $\eps$, then the outer first hits
also exist, and their elapsed continuation times are bounded independently
of $\eps$. Moreover,
\[
 \{\ell_\infty(Z):Z\in E_{\rho_1}(K,b,U)\}
 =\{\ell_\infty(Z):Z\in E_{\rho_2}(K,b,U)\}.
\]
Consequently $\mathcal J_{\rho_1}(K,b,U)=\mathcal J_{\rho_2}(K,b,U)$.
\end{lemma}
\begin{proof}(of Lemma~\ref{lem:section-independence}).
\label{proof:lem:section-independence}
$E_\rho$ is a compact, nonempty set. At its defining first hits, $M\succeq0$, $\rank M\le q$, and $\norm M_F=\rho$, so
\[
 \norm W_F^2=\tr M\le\sqrt q\,\rho.
\]
Thus every sequence of hit factors has a convergent subsequence. The set of their subsequential limits as $\eps\downarrow0$ is closed: equivalently, it is the intersection, over positive integers $j$, of the closures of the hit factors with $0<\eps<1/j$. 
Use $\sigma_{\min}(C)>0$ and $\bar\rho$ from \eqref{eq:small-ball}. The measurement formula gives
\[
 \norm{\Lop(M)}_{\op}\le \frac1n\sum_l
  |x_l^\top Mx_l|\,\norm{x_l}^2
 \le \left(n^{-1}\sum_l\norm{x_l}^4\right)\norm M_F.
\]
For $\norm M_F\le\bar\rho$, singular-value perturbation therefore yields
\begin{align*}
 \sigma_{\min}(R(M))
 &\ge\sigma_{\min}(C)-\norm{\Lop(M)}_{\op}\ge\sigma_{\min}(C)/2,\\
 \norm{G(W)}_F&=2\norm{WR(M)}_F\ge\sigma_{\min}(C)\norm W_F.
\end{align*}

All derivatives in the next calculation use continuation time $\tau$. Before a first hit inside this ball, the angle inequality in Proposition~\ref{main:global} gives
\[
 -f'\ge a_K^{-1}\norm G_F\norm{W'}_F
 \ge\frac{\sigma_{\min}(C)}{a_K}\norm W_F\norm{W'}_F
 \ge\frac{\sigma_{\min}(C)}{2a_K}\left|\frac{d}{d\tau}\tr(W^\top W)\right|.
\]
Integrating the last inequality and then using the signed change in the trace, which is at most its total variation, proves
\[
 f(W(T_\rho^\eps))\le f(W(0))-
 \frac{\sigma_{\min}(C)}{2a_K}\{\tr M(T_\rho^\eps)-\tr M(0)\}.
\]
The point denoted by $T_\rho^\eps$ is the same hit point in either clock.

Take $0<\rho_1<\rho_2<\bar\rho$ and suppose the inner first hits exist. Since $\tr M(T_{\rho_1}^\eps)\ge\rho_1$, $\tr M(0)\to0$, and $f(W(0))\to\Phi(0)$, the preceding estimate gives, for all sufficiently small $\eps$,
\[
 f(W(T_{\rho_1}^\eps))\le\Phi(0)-\frac{\sigma_{\min}(C)\rho_1}{4a_K}.
\]
After this inner hit, and for as long as the outer hit has not occurred, loss monotonicity and \eqref{eq:loss-expand} imply
\[
 \frac{\sigma_{\min}(C)\rho_1}{4a_K}\le\Phi(0)-\Phi(M)\le\ip C M
       \le\norm C_F\norm M_F.
\]
Consequently the trajectory cannot return arbitrarily close to zero on that interval. Its loss decreases at a rate bounded away from zero:
\[
 -f'\ge\norm G_F^2\ge\sigma_{\min}(C)^2\tr M
       \ge\frac{\sigma_{\min}(C)^3\rho_1}{4a_K\norm C_F}.
\]
Because the loss is nonnegative, the elapsed continuation time between the two sections is at most $4a_K\Phi(0)\norm C_F/[\sigma_{\min}(C)^3\rho_1]$. If the outer hit never occurred, integration of this strict decrease would give negative loss. This proves existence of the outer hit as well as a bound independent of $\eps$ on the intervening continuation time.

Extracting convergent hit factors and elapsed times from a sequence at either radius gives the two image-set relations
\[
 E_{\rho_2}\subseteq\bigcup_{Z\in E_{\rho_1}}
 \left\{\varphi_K(s,Z):0\le s\le
 \frac{4a_K\Phi(0)\norm C_F}{\sigma_{\min}(C)^3\rho_1}\right\},
\]
and, for every $Z\in E_{\rho_1}$,
\[
 \left\{\varphi_K(s,Z):0\le s\le
 \frac{4a_K\Phi(0)\norm C_F}{\sigma_{\min}(C)^3\rho_1}\right\}
 \cap E_{\rho_2}\ne\varnothing.
\]
To verify these relations, apply \eqref{eq:gronwall} with $\eta=c_\eta\eps$ on the common bounded continuation interval. Continuity in time of the limiting solution then gives convergence at the converging inter-section times. The semigroup property preserves terminal loss at every point of each image set. The two relations imply equality of the sets of attainable terminal losses. Equation~\eqref{eq:terminal-maximum} yields $\mathcal J_{\rho_1}=\mathcal J_{\rho_2}$. The same argument extends the conclusion to every outer radius below $\bar\rho$.
\end{proof}

\subsection{Strict loss transfer and all-time separation}
Under the pointwise hypotheses of Theorem~\ref{thm:passage}, the selected
continuation has the same terminal predictor and loss for every positive
entrance point and hidden frame. On its Gaussian-full event, at every
sufficiently small fixed section,
\begin{equation}\label{eq:J-deterministic}
 \mathcal J_\rho(K,b,U)=\ell_{K,d_K(b)}.
\end{equation}
Every entry factor has this terminal loss. If $\kloss(b)<\infty$ and the
hypotheses of Theorem~\ref{thm:separation} apply at $K=\kloss(b)$, then
$\kpath(b)\le\kloss(b)$. We prove these characterizations together with
the strict-loss transfer.
\begin{proof}(of Theorem~\ref{thm:separation} and the entry-set characterization).
\label{proof:thm:separation}
For two positive points of the same local selected entrance, the change of time $\dd t/\dd\tau=\norm{G(W)}_F$ has a finite positive duration between them, since the gradient is continuous and nonzero on that compact segment. The semigroup identity
\[
 \varphi_K(\tau,\varphi_K(s,Z))=\varphi_K(\tau+s,Z)
\]
therefore identifies their continuations after a time shift. For $O\in\Ogroup(m)$,
$G(OW)=OG(W)$ and $H_{K,0}(OG)=OH_{K,0}(G)$, so uniqueness gives
$\varphi_K(\tau,OZ)=O\varphi_K(\tau,Z)$.
Proposition~\ref{main:global} gives the terminal factor. Its Gram and loss are unchanged by the time shift or constant orthogonal transformation, proving that $M_{s,K}$ and $\ell_{K,d_K(b)}$ are well defined and $\Phi(M_{s,K})=\ell_{K,d_K(b)}$.

At a sufficiently small first-radius section, Theorem~\ref{thm:passage} gives the same limiting Gram for every entry factor. Two factors with the same Gram differ by a left orthogonal transformation: their corresponding columns define an isometry between column spans, which extends to $\R^m$. All entry factors consequently have the selected terminal loss, and Lemma~\ref{lem:terminal-maximum} gives \eqref{eq:J-deterministic}.

Assume now \eqref{eq:bottleneck} and $\ell_{K,d_K(b)}<\phi_1$. Choose $\Delta_*>0$ with $\ell_{K,d_K(b)}<\phi_1-4\Delta_*$. Choose a fixed small section $\rho<\norm{M_1}_F/2$ whose selected loss exceeds $\phi_1-4\Delta_*$. This is possible because its loss tends to $\Phi(0)>\phi_1$ at zero. Stop the selected continuation at its first crossing of $\phi_1-4\Delta_*$, which occurs at a finite deterministic elapsed time.

The predictor image of this compact segment is disjoint from $M_1$. Every factor $W$ of $M_1$ satisfies
\[
 \norm{WR_1}_F^2=\tr(R_1M_1R_1)=0
\]
and is stationary. Local uniqueness prevents a nonconstant solution from meeting such a factor at finite time. The segment is nonconstant because its endpoint has loss below $\phi_1$. Its normalized predictor distance from $M_1$ therefore has a positive minimum.

Before the first radius hit the normalized distance is at least
$1-\rho/\norm{M_1}_F$. At any point below loss $\phi_1-\Delta_*$, the exact expansion
\begin{equation}\label{eq:tail}
 \Phi(M)-\phi_1=-\langle R_1,M-M_1\rangle
 +\tfrac12\langle M-M_1,\Lop(M-M_1)\rangle
\end{equation}
gives
\[
 \Delta_*\le\langle R_1,M-M_1\rangle
 \le\norm{R_1}_F\norm{M-M_1}_F.
\]
Choose the theorem constant $d_*>0$ smaller than half the normalized minimum on the compact selected segment, smaller than $1-\rho/\norm{M_1}_F$, and smaller than $\Delta_*/(\norm{R_1}_F\norm{M_1}_F)$. These choices depend only on the selected continuation.

For each fixed seed in the entrance event, apply one matching left orthogonal transformation to its actual path at the section and keep that transformation fixed during comparison. Theorem~\ref{thm:passage} and \eqref{eq:gronwall}, with $\eta=c_\eta\eps$, give uniform factor convergence on the entire chosen compact segment. Predictor convergence follows from
\[
 \norm{A^\top A-B^\top B}_F
 \le(\norm A_F+\norm B_F)\norm{A-B}_F.
\]
For sufficiently small $\eps$, the actual segment stays at normalized distance at least $d_*$ and ends below loss $\phi_1-\Delta_*$. Loss monotonicity preserves this gap at all later times, where \eqref{eq:tail} gives the required distance bound. The pre-section estimate covers all earlier times. This proves both assertions, with seed-dependent initialization cutoff. The positive time change covers the forward half-line, so the loss crossing is finite in physical time as well. Applying the conclusion at $K=\kloss(b)$ under its stated entrance hypotheses gives $\kpath(b)\le\kloss(b)$.
\end{proof}

\subsection{Selected capture, equal loss, and the combined corollary}
The full-width loss statement uses only the full-rank seed event.
Capture follows from a finite selected segment. For equal loss at a
different endpoint, a local analytic length estimate controls the
remaining tail. We prove these parts in an order that makes the
full-width loss improvement available when needed.

\begin{proof}(of Corollary~\ref{thm:conditional-classification}).

(iii) \label{proof:cor:actual-optimal-loss}
The PSD rank-one perturbation in the proof of Corollary~\ref{cor:comparison-saddle} has loss below $\phi_1$. Compact PSD sublevels give a minimizer, so $\Phi_*<\phi_1$.
The determinant of the first $r$ rows of $U$ is a nonzero polynomial. Its zero set has Lebesgue measure zero \citep{mityagin2020zero}. On this one Gaussian-full event, $U$ has column rank $r$, and multiplication by every invertible $D_{\eps,b}$ preserves it, simultaneously for all $\eps>0$ and $b>1$. Corollary~\ref{cor:full-rank-optimum} applies to each positive initialization and each finite $K$, giving terminal loss $\Phi_*$. Convergence forces crossing of $(\phi_1+\Phi_*)/2$ at a finite time. Every such trajectory has infimal loss $\Phi_*$, proving the small-initialization strict loss inequality.

(i) \label{proof:lem:selected-capture}
Fix a seed in the entrance event and restart at a small fixed section, using constant left transformations as above. Fix any open predictor ball about $M_1$ of positive radius. Convergence of the selected continuation gives a finite point inside the concentric ball of half that radius. At this fixed time, Corollary~\ref{cor:flow-comparison} makes the actual predictor error less than the remaining half radius for sufficiently small $\eps$. Thus every such ball contains an actual path point for all sufficiently small initializations. Nonnegativity of $d_1$ proves $d_1\to0$. The comparisons use the same entrance event, with a finite comparison time chosen for each ball.

(ii) \label{proof:lem:equal-loss-separation}
Fix a small first-radius section and its selected factor $Z$. Proposition~\ref{main:global} gives the limit $W_s$ of $\varphi_K(\cdot,Z)$, with $W_s^\top W_s=M_{s,K}$ and $f(W_s)=\phi_1$. Since $M_{s,K}\ne M_1$, choose $R>0$ so that every factor in $\overline B_R(W_s)$ has predictor distance at least $\tfrac12\norm{M_{s,K}-M_1}_F$. Decrease $R$ so that \citet[Lemma~2.1]{absil2005convergence} gives $c>0$ and $0\le\theta<1$ with
\[
 \norm{\nabla f(W)}_F\ge c|f(W)-\phi_1|^\theta
\]
throughout this ball. Choose a loss width
\[
 0<a\le\left(\frac{c(1-\theta)R}{8a_K}\right)^{1/(1-\theta)}.
\]
These constants concern only the selected continuation.

On a segment in this ball with $|f-\phi_1|\le a$, the uniform angle bound
$-f'\ge a_K^{-1}\norm{\nabla f}_F\norm{W'}_F$
controls its length. Above $\phi_1$, differentiation of $(f-\phi_1)^{1-\theta}$ and integration give length at most
$a_Ka^{1-\theta}/[c(1-\theta)]\le R/8$.
Below $\phi_1$, differentiation of $(\phi_1-f)^{1-\theta}$ gives the same bound. The integrals converge at the crossing because $\theta<1$. Constant-loss intervals have zero gradient and velocity. The total length in the ball and loss interval is therefore at most $R/4$.

Choose a finite point of $\varphi_K(\cdot,Z)$ in $B_{R/8}(W_s)$ with loss gap less than $a/4$. Fix its elapsed time before taking $\eps\downarrow0$. At the actual first section, apply one constant hidden orthogonal transformation to each path so that its factor converges to $Z$. Indeed, a contrary sequence has a convergent subsequence of bounded factors; its limiting Gram is $Z^\top Z$, and equality of Grams gives a left orthogonal equivalence, contradicting positive distance from that class. Equivariance and Corollary~\ref{cor:flow-comparison} then give convergence on the compact segment ending at the chosen point.

For each seed in the entrance event and sufficiently small $\eps$, the transformed actual endpoint lies in $B_{R/4}(W_s)$ and has loss gap less than $a/2$. Until it reaches loss $\phi_1-a$, monotonicity keeps it in the loss interval. Exiting $B_R(W_s)$ would require length at least $3R/4$, whereas the proved bound is $R/4$. Thus either the path remains in the ball forever, maintaining predictor distance at least $\tfrac12\norm{M_{s,K}-M_1}_F$, or it reaches loss $\phi_1-a$ while still in the ball. In the latter case \eqref{eq:tail} and monotonicity give, at every later time,
\[
 a\le\langle R_1,M-M_1\rangle
 -\tfrac12\langle M-M_1,\Lop(M-M_1)\rangle
 \le\norm{R_1}_F\norm{M-M_1}_F.
\]

Before the first section, take its radius below $\norm{M_1}_F/2$, so the predictor distance is at least $\norm{M_1}_F/2$. The selected compact segment has positive minimum distance from $M_1$: every factor of $M_1$ is stationary, and uniqueness would make a solution meeting it constant, contradicting its different terminal predictor. Finite-time comparison transfers half this minimum to the actual segment. The minimum of the early bound, the transferred segment bound, the terminal-ball bound, and $a/\norm{R_1}_F$, divided by $\norm{M_1}_F$, gives deterministic $d_*>0$. Only the initialization cutoff may depend on the seed. The physical-time change covers the full forward half-line. At $m\ge r$, the full-width part of the corollary supplies the strict actual loss crossing, completing bypass.

(iv) \label{proof:thm:conditional-classification}
If $\ell_{K,d_K(b)}<\phi_1$, Theorem~\ref{thm:separation} gives bypass,
and $\Phi(M_{s,K})=\ell_{K,d_K(b)}$ implies $M_{s,K}\ne M_1$.
If $\ell_{K,d_K(b)}=\phi_1$ and $M_{s,K}\ne M_1$, Corollary~\ref{thm:conditional-classification}(ii) gives bypass. These cover the entire
premise $\ell_{K,d_K(b)}\le\phi_1$ when the terminal predictor differs.
If it equals $M_1$, Corollary~\ref{thm:conditional-classification}(i) gives visitation,
which contradicts the positive all-time distance in the definition of
bypass. This proves the reverse implication and the equivalence.
\end{proof}

\subsection{Rank-one and constant-rank consequences}
A rank-one continuation follows the ordinary gradient-flow path after a
positive change of time. This gives a lower bound on the loss-test depth,
including repeated unused eigenvalues. Fixed-rank entrance uniqueness
also makes the selected loss constant on an admissible exponent interval
where the entrance rank is unchanged.
\begin{proposition}[The rank-one lower bound on loss-test depth]\label{prop:rank-one-loss}
Assume the empirical task satisfies Assumption~\ref{ass:task} and
\eqref{eq:bottleneck}, with $b>1$ and $c_\eta>0$. A rank-one entrance
satisfying $W(t)/t\to Q_{1,K}$ has terminal continuation loss $\phi_1$.
For every fixed finite integer $K$ with $a_K\lambda_2/\lambda_1<b$,
the Gaussian first-radius limit exists almost surely at all sufficiently
small radii and has rank one. Its terminal-loss test equals $\phi_1$. Repetitions among
$\lambda_2,\ldots,\lambda_r$ are allowed in this assertion.
Consequently
\begin{equation}\label{eq:lowerK}
 \kloss(b)\ge\left\lceil\log_{3/2}(b\lambda_1/\lambda_2)\right\rceil.
\end{equation}
\end{proposition}
\begin{proof}(of Proposition~\ref{prop:rank-one-loss}).
\label{proof:prop:rank-one-loss}

Rank-one continuation.

For a rank-one factor, $G(W)$ has rank at most one. At a nonzero such
gradient, $p_K(1)=1$ gives
\[
 P_K(G(W)/\norm{G(W)}_F)=G(W)/\norm{G(W)}_F,
 \qquad H_{K,0}(G(W))=G(W).
\]
The second identity also holds at $G(W)=0$. Thus the normalized
rank-one equation, and its boundary matrix $Q_{1,K}=e_1v_1^\top$,
are independent of $K$. Lemma~\ref{lem:deterministic-entrance}
identifies a single such rank-one trajectory.

To match its gradient-flow parametrization to Proposition~\ref{main:gd},
use \eqref{eq:entrance-expansion} to obtain
\[
 \norm{G(W(t))}_F=2\lambda_1t+O(t^3),\qquad
 \frac{\dd\tau}{\dd t}=\frac1{2\lambda_1t}+O(t).
\]
Choose the additive origin of continuation time so that
$\tau=(2\lambda_1)^{-1}\log t+O(t^2)$.
Consequently its predictor in this time variable has the asymptotic form
\[
 M(\tau)=e^{4\lambda_1\tau}v_1v_1^\top
          +O(e^{8\lambda_1\tau})\qquad(\tau\to-\infty).
\]
Starting this solution at successively earlier times gives initial
predictors whose normalized projection on $v_1v_1^\top$ tends to one.
Proposition~\ref{main:gd} therefore identifies it, up to a constant
time shift, with $\mathcal M^{\rm G}$. For example, the shift
$(4\lambda_1)^{-1}\log(1/\langle M(\tau),v_1v_1^\top\rangle)$ equals
$-\tau+o(1)$, so the two limiting parametrizations agree at every fixed
time. Its terminal predictor is therefore $M_1$ and its terminal loss
is $\phi_1$.

Actual rank-one selection below the threshold.

Fix $K$ with $a_K\lambda_2/\lambda_1<b$ and a seed satisfying
$u^\top v_1\ne0$. The latter event has Gaussian probability one.
The largest unused rate is
\[
 g_2^{(1)}=a_K\lambda_2/\lambda_1-1,
 \qquad (b-1)-g_2^{(1)}=b-a_K\lambda_2/\lambda_1>0.
\]
By
Lemmas~\ref{lem:main-initial-approach} and~\ref{lem:gaussian-entry}, a
sufficiently large fixed $\sigma_0\ge1$ places the actual rescaled matrix in a
small rank-one neighborhood at $\eps\sigma_0$ and gives
\[
 \norm{Z^\eps(\eps\sigma_0)}_F=O_U(\eps^{b-1}).
\]
The rank-one tangent equation is stable by Lemma~\ref{lem:spectrum},
using $\lambda_1>\lambda_2$. Apply the norm part of
Lemma~\ref{lem:integral}, which uses only the largest unused rate, on the interval stopped at the first chart boundary or at a
fixed small $t_*$. Choosing the chart, $1/\sigma_0$, and $t_*$
sufficiently small makes $\norm E_\op\le(b-a_K\lambda_2/\lambda_1)/2$ there, and yields
\[
 \norm{Z^\eps(t)}_F
 =O_U\!\left(\eps^{b-1}(t/(\eps\sigma_0))^{g_2^{(1)}+(b-a_K\lambda_2/\lambda_1)/2}\right).
\]
After the task, $K,b$, seed, chart, $\sigma_0$, and $t_*$ are fixed, both estimates are uniform over the stopped interval and all sufficiently small $\eps$. Its supremum is at most a constant times
\[
 \eps^{\min\{b-1,\,(b-a_K\lambda_2/\lambda_1)/2\}},
\]
which tends to zero. Its squared logarithmic-time integral also tends
to zero, since the interval length is $O_U(\log(1/\eps))$.

The tangent inequality in Lemma~\ref{lem:integral}
excludes a first chart exit: after multiplication by $e^{\omega v}$
and integration, it bounds $\norm y$ by its contracted incoming size
plus $C/\omega$ times the supremum of
$\norm Z_F^2+t_*^2+1/\sigma_0$. Choose the incoming tangent size,
$t_*^2$, and $1/\sigma_0$ to leave a strict margin inside the chart;
then choose $\eps$ small so that the normal term fits the same margin.
The normal estimate excludes exit through its own boundary. The stopped
interval therefore extends to $t_*$.

We have established $(\eps\sigma_0)\to0$, $(\eps\sigma_0)\ge\eps$, local
residence, and $Z^\eps(t)\to0$ for every fixed $t>0$. Also
$W_K^{\eps,b}(0;U)\to0$. Lemma~\ref{lem:deterministic-entrance}
now gives the rank-one first-radius limit and existence of all its
sufficiently small actual sections. Every limiting factor on such a
section has the same Gram matrix as the rank-one solution already
identified. Left orthogonal equivariance and
Lemma~\ref{lem:terminal-maximum} give
$\mathcal J_\rho(K,b,U)=\phi_1$.

Every nonnegative integer strictly below the ceiling in
\eqref{eq:lowerK} satisfies the strict inequality just proved and hence
fails the terminal-loss test. This proves the bound, while retaining
the original strict inequality at the boundary.
\end{proof}

\begin{proposition}[Terminal loss on a constant-rank exponent interval]\label{prop:constant-rank-loss}
Fix a finite depth $K$. On any interval of exponents $b$ for which the
hypotheses of Theorem~\ref{thm:passage} hold and $d_K(b)$ is constant,
the selected entrance continuation and its terminal loss are independent
of $b$, up to the positive starting point and hidden orthogonal frame.
\end{proposition}
\begin{proof}(of Proposition~\ref{prop:constant-rank-loss}).
\label{proof:prop:constant-rank-loss}
The boundary matrix $Q_{d_K(b),K}$ depends on $K$ and the constant rank
$d_K(b)$. Theorem~\ref{thm:passage} identifies the entrance with the unique
fixed-rank solution having this boundary condition. Hence the selected
continuation is the same throughout the exponent interval, up to its
positive starting point and hidden orthogonal frame.
The continuation identity proved with \eqref{eq:J-deterministic}
gives independence of terminal loss from these choices.
\end{proof}

%% file: appendices/depth.tex
\section{Residual-Inertia Certificate and the Depth Tail}\label{app:depth}
A spectral estimate first gives an explicit cutoff for entrance capacity.
The critical loss band then converts preserved finite-time rank into a
strict selected loss decrease. The final proof combines these two inputs
with the all-time transfer theorem.

\subsection{Capacity saturation and an explicit sufficient cutoff}
For $2\le d\le p$, the following sufficient estimate holds under
Assumptions~\ref{ass:task} and~\ref{ass:prefix}:
\begin{equation}\label{eq:highK-bound}
 a_K\lambda_d>
 \bigl(1+2(d-1)\max\{1,b-1\}\bigr)
 \left(\sum_{h=1}^d\lambda_h^2\right)^{1/2}
 \quad\Longrightarrow\quad d_K(b)\ge d.
\end{equation}
The estimate proves that every intermediate clock completes; the strict
integer convention in \eqref{eq:capacity-depth} then applies it with
$d=p$.
\begin{proof}(the sufficient bound~\eqref{eq:highK-bound}).
\label{proof:cor:finite-depth}
At any existing support $\{1,\ldots,h\}$,
\[
 s_{h,K}^2=\sum_{a\le h}\lambda_a^2p_K(s_a)^2
 \le\sum_{a=1}^h\lambda_a^2\le\sum_{a=1}^d\lambda_a^2.
\]
Condition \eqref{eq:highK-bound} and Lemma~\ref{lem:add-zero} therefore give successive existence of all supports up to $d$, and every required rate exceeds $2(d-1)\max\{1,b-1\}$. Since $1-\lambda_{h+1}/\lambda_h<1$, the time formulas give
\[
 x_{h+1}\le x_h+\frac{b-1+x_h}{2(d-1)\max\{1,b-1\}}.
\]
Starting at $x_1=0$, induction gives
\[
 x_h\le(b-1)\left\{\left(1+\frac{1}{2(d-1)\max\{1,b-1\}}\right)^{h-1}-1\right\}.
\]
For $2\le h\le d$,
\begin{align*}
 \left(1+\frac1{2(d-1)\max\{1,b-1\}}\right)^{h-1}
 &\le\exp\!\left(\frac{h-1}{2(d-1)\max\{1,b-1\}}\right)\\
 &<1+\frac{h-1}{(d-1)\max\{1,b-1\}}.
\end{align*}
The last exponent is at most $1/2$, and $e^z-1<2z$ on $0<z\le1/2$ follows by integrating $e^s\le e^{1/2}<2$.
Consequently the bound for $x_h$ is strictly below
$(b-1)/\max\{1,b-1\}\le1$. Every candidate is completed.
\end{proof}

\begin{proof}(of Proposition~\ref{prop:capacity-tail}).
\label{proof:prop:capacity-tail}
The strict floor-plus-one convention in \eqref{eq:capacity-depth} implies, for $K\ge K_{\rm cap}(b)$,
\[
 a_K\lambda_p>
 [1+2(p-1)\max\{1,b-1\}]
 \left(\sum_{i=1}^p\lambda_i^2\right)^{1/2}.
\]
Apply the sufficient bound~\eqref{eq:highK-bound} with $d=p$. Every required support exists and every candidate is strictly below one. Capacity stops the process after rank $p$. Assumption~\ref{ass:prefix} supplies the required gaps, so Theorem~\ref{thm:passage} gives the actual rank-$p$ entrance. Intersecting its full-measure events over the countable finite depths $K\ge K_{\rm cap}(b)$ proves the simultaneous seed assertion.

The function $K_{\rm cap}(b)$ is nondecreasing in $b$. Thus for
$1<b\le B$ and $K\ge K_{\rm cap}(B)$ every candidate is strictly
below one. No member of $\mathcal B_{C,K}$ lies in $(1,B]$ at
these depths. There remain $K_{\rm cap}(B)$ nonnegative integer
depths below the cutoff, and Proposition~\ref{prop:clock-domain}
gives at most $p-1$ critical values at each. Their finite union has
the stated cardinality bound.
\end{proof}

\subsection{The critical loss band and sufficient rank condition}
The definitions in \eqref{eq:criticalset} concern stationary PSD
predictors with loss between $\phi_1$ and $\Phi(0)$. Their residual
inertia is bounded below uniformly by an integer. The dynamic bound
excludes this entire set as a selected endpoint when the entrance rank
is large enough.
\begin{lemma}[Positive residual directions above the comparison loss]\label{lem:positive-residual}
Under \eqref{eq:bottleneck}, $\CE_{\ge\phi_1}^{(q)}$ is nonempty and compact,
and $p_*^{(q)}$ is a well-defined integer satisfying $1\le p_*^{(q)}\le n_+(R_1)\le n_+(C)-1$.
\end{lemma}
\begin{proof}(of Lemma~\ref{lem:positive-residual}).
\label{proof:lem:positive-residual}
The set is closed and lies in a compact PSD sublevel, so it is compact; it is nonempty because it contains $M_1$. The value $p_*^{(q)}$ is a minimum of a nonempty set of integers between $0$ and $r$, so it is well defined.

The positive residual eigenvalue at $M_1$ gives a PSD predictor with
loss strictly below $\phi_1$, by the rank-one perturbation identity in
the proof of Corollary~\ref{cor:comparison-saddle}.
Thus the PSD-optimal loss is strictly below $\phi_1$. Every member of \eqref{eq:criticalset} is consequently nonglobal. If one had $R(M)\preceq0$, its stationarity would imply PSD optimality by Lemma~\ref{lem:psd-optimality}, a contradiction. It must have at least one positive residual eigenvalue.

Finally $M_1\in\CE_{\ge\phi_1}^{(q)}$ gives
$p_*^{(q)}\le n_+(R_1)$. Lemma~\ref{lem:static-inertia} at the
rank-one stationary predictor $M_1$ yields
$n_+(R_1)\le n_+(C)-1$, proving the stated upper bound.
\end{proof}

\begin{corollary}[A sufficient rank condition for loss crossing]\label{cor:rank-loss}
Under Theorem~\ref{thm:passage} and \eqref{eq:bottleneck}, if $d_K(b)+p_*^{(q)}>r$, then $\ell_{K,d_K(b)}<\phi_1$ and $K$ is successful in \eqref{eq:Kb}.
\end{corollary}
\begin{proof}(of Corollary~\ref{cor:rank-loss}).
\label{proof:cor:rank-loss}
Choose a small positive time on the selected
entrance. By \eqref{eq:entrance-expansion}, its predictor has the form
\[
 M(t)=t^2Q_{d_K(b),K}^\top Q_{d_K(b),K}+O(t^4).
\]
The row space of $Q_{d_K(b),K}$ lies in positive task eigendirections,
so
$\langle C,Q_{d_K(b),K}^\top Q_{d_K(b),K}\rangle
=\sum_{i=1}^{d_K(b)}\lambda_iq_i^2>0$.
The empirical expansion \eqref{eq:loss-expand} now gives
\[
 \Phi(M(t))=\Phi(0)
 -t^2\langle C,Q_{d_K(b),K}^\top Q_{d_K(b),K}\rangle+O(t^4)<\Phi(0)
\]
for a sufficiently small fixed positive time. Start the continuation
there. Proposition~\ref{main:global} gives convergence of the selected
factor and preserves its initial rank $d_K(b)$ at finite times. Its
already defined terminal predictor $M_{s,K}$ is PSD, has rank at most
$d_K(b)\le q$, satisfies $M_{s,K}R(M_{s,K})=0$, and has loss at most
$\Phi(0)$ by descent.

If $\Phi(M_{s,K})\ge\phi_1$, these properties place $M_{s,K}$ in
$\CE_{\ge\phi_1}^{(q)}$. Its residual then has at least
$p_*^{(q)}$ positive eigenvalues. Lemma~\ref{lem:inertia}, applied
with the initial rank $d_K(b)$, instead gives
\[
 d_K(b)+n_+(R(M_{s,K}))\le r.
\]
This contradicts $d_K(b)+p_*^{(q)}>r$. Hence
$\ell_{K,d_K(b)}=\Phi(M_{s,K})<\phi_1$.
The identity \eqref{eq:J-deterministic} makes the same strict
inequality hold for the terminal-loss test at every sufficiently
small section, proving success in \eqref{eq:Kb}.
\end{proof}

\subsection{The eventual finite-depth theorem}
The proof is pointwise in each finite depth. Its final countable
intersection combines the probability-one events while retaining the
individual constants and initialization cutoffs.
\begin{proof}(of Theorem~\ref{thm:eventual-depth}).
\label{proof:thm:eventual-depth}
Fix $b>1$ and $c_\eta>0$, and first fix one finite integer
$K\ge K_{\rm cap}(b)$. Proposition~\ref{prop:capacity-tail}
proves existence of all candidate supports, strict completion of
every clock, and the pointwise applicability of
Theorem~\ref{thm:passage}. In particular the actual selected
entrance has rank $d_K(b)=p$.

The inequality \eqref{eq:task-inertia-condition} is therefore the
hypothesis $d_K(b)+p_*^{(q)}>r$ of
Corollary~\ref{cor:rank-loss}. That corollary gives
$\ell_{K,p}<\phi_1$ using finite-time rank $p$ and stationary
residual inertia. It allows the terminal predictor to have smaller
rank. Theorem~\ref{thm:separation} now supplies positive constants
$d_{*,K}$ and $\Delta_{*,K}$ determined by the selected continuation,
and a Gaussian-full event on which a seed-dependent small
initialization cutoff yields both inequalities in
\eqref{eq:eventual-actual}. Its proof compares a fixed finite
continuation segment, and then uses loss monotonicity for the full
tail, so the second inequality in \eqref{eq:eventual-actual} holds
at a finite physical time for each positive initialization.

Perform this argument separately for each integer
$K\ge K_{\rm cap}(b)$. The complement of the intersection of the
resulting events is a countable union of Gaussian-null sets and is
therefore null. On the intersection, keep the constants and cutoff
already obtained for each $K$. The constants $d_{*,K}$ and
$\Delta_{*,K}$ are independent of the seed, whereas
the initialization cutoff may depend on that seed and on $K$. This proves the asserted simultaneous statement for the countable
family of fixed finite depths.
\end{proof}

%% file: main.bbl
\begin{thebibliography}{31}
\providecommand{\natexlab}[1]{#1}
\providecommand{\url}[1]{\texttt{#1}}
\expandafter\ifx\csname urlstyle\endcsname\relax
  \providecommand{\doi}[1]{doi: #1}\else
  \providecommand{\doi}{doi: \begingroup \urlstyle{rm}\Url}\fi

\bibitem[Absil et~al.(2005)Absil, Mahony, and Andrews]{absil2005convergence}
P.-A. Absil, R.~Mahony, and B.~Andrews.
\newblock Convergence of the iterates of descent methods for analytic cost functions.
\newblock \emph{SIAM Journal on Optimization}, 16\penalty0 (2):\penalty0 531--547, 2005.
\newblock \doi{10.1137/040605266}.

\bibitem[Bernstein and Newhouse(2024)]{bernstein2024old}
J.~Bernstein and L.~Newhouse.
\newblock Old optimizer, new norm: An anthology.
\newblock In \emph{NeurIPS 2024 Workshop on Optimization for Machine Learning}, 2024.
\newblock URL \url{https://arxiv.org/abs/2409.20325}.

\bibitem[Berthier(2023)]{berthier2023incremental}
R.~Berthier.
\newblock Incremental learning in diagonal linear networks.
\newblock \emph{Journal of Machine Learning Research}, 24\penalty0 (171):\penalty0 1--26, 2023.

\bibitem[Bj{\"o}rck and Bowie(1971)]{bjorck1971orthogonal}
{\AA}.~Bj{\"o}rck and C.~Bowie.
\newblock An iterative algorithm for computing the best estimate of an orthogonal matrix.
\newblock \emph{SIAM Journal on Numerical Analysis}, 8\penalty0 (2):\penalty0 358--364, 1971.
\newblock \doi{10.1137/0708036}.

\bibitem[Braun et~al.(2026)Braun, Bao, Huang, and Imaizumi]{braun2026spectral}
G.~Braun, H.~Bao, W.~Huang, and M.~Imaizumi.
\newblock Spectral gradient descent mitigates anisotropy-driven misalignment: A case study in phase retrieval.
\newblock In \emph{Proceedings of the 43rd International Conference on Machine Learning}, volume 306 of \emph{Proceedings of Machine Learning Research}, pages 9687--9733, 2026.

\bibitem[Chizat et~al.(2019)Chizat, Oyallon, and Bach]{chizat2019lazy}
L.~Chizat, E.~Oyallon, and F.~Bach.
\newblock On lazy training in differentiable programming.
\newblock In \emph{Advances in Neural Information Processing Systems}, volume~32, 2019.

\bibitem[Dandi et~al.(2024)Dandi, Krzakala, Loureiro, Pesce, and Stephan]{dandi2024giant}
Y.~Dandi, F.~Krzakala, B.~Loureiro, L.~Pesce, and L.~Stephan.
\newblock How two-layer neural networks learn, one (giant) step at a time.
\newblock \emph{Journal of Machine Learning Research}, 25\penalty0 (349):\penalty0 1--65, 2024.
\newblock URL \url{https://jmlr.org/papers/v25/23-1543.html}.

\bibitem[Higham(1986)]{higham1986polar}
N.~J. Higham.
\newblock Computing the polar decomposition---with applications.
\newblock \emph{SIAM Journal on Scientific and Statistical Computing}, 7\penalty0 (4):\penalty0 1160--1174, 1986.
\newblock \doi{10.1137/0907079}.

\bibitem[Higham and Schreiber(1990)]{higham1990fast}
N.~J. Higham and R.~S. Schreiber.
\newblock Fast polar decomposition of an arbitrary matrix.
\newblock \emph{SIAM Journal on Scientific and Statistical Computing}, 11\penalty0 (4):\penalty0 648--655, 1990.
\newblock \doi{10.1137/0911038}.

\bibitem[Jacot et~al.(2021)Jacot, Ged, {\c S}im{\c s}ek, Hongler, and Gabriel]{jacot2021saddle}
A.~Jacot, F.~Ged, B.~{\c S}im{\c s}ek, C.~Hongler, and F.~Gabriel.
\newblock Saddle-to-saddle dynamics in deep linear networks: Small initialization training, symmetry, and sparsity.
\newblock \emph{arXiv preprint arXiv:2106.15933}, 2021.

\bibitem[Jiang et~al.(2023)Jiang, Chen, and Ding]{jiang2023algorithmic}
L.~Jiang, Y.~Chen, and L.~Ding.
\newblock Algorithmic regularization in model-free overparametrized asymmetric matrix factorization.
\newblock \emph{SIAM Journal on Mathematics of Data Science}, 5\penalty0 (3):\penalty0 723--744, 2023.
\newblock \doi{10.1137/22M1519833}.

\bibitem[Jin et~al.(2023)Jin, Li, Lyu, Du, and Lee]{jin2023incremental}
J.~Jin, Z.~Li, K.~Lyu, S.~S. Du, and J.~D. Lee.
\newblock Understanding incremental learning of gradient descent: A fine-grained analysis of matrix sensing.
\newblock In \emph{Proceedings of the 40th International Conference on Machine Learning}, volume 202 of \emph{Proceedings of Machine Learning Research}, pages 15200--15238, 2023.

\bibitem[Jordan et~al.(2024)Jordan, Jin, Boza, You, Cesista, Newhouse, and Bernstein]{jordan2024muon}
K.~Jordan, Y.~Jin, V.~Boza, J.~You, F.~Cesista, L.~Newhouse, and J.~Bernstein.
\newblock Muon: An optimizer for hidden layers in neural networks.
\newblock Technical blog, 2024.
\newblock URL \url{https://kellerjordan.github.io/posts/muon/}.

\bibitem[Kim and Oh(2026)]{kim2026convergence}
G.~Y. Kim and M.-h. Oh.
\newblock Convergence of {Muon} with {Newton--Schulz}.
\newblock In \emph{International Conference on Learning Representations}, 2026.

\bibitem[{Kimi Team}(2025)]{kimi2025k2}
{Kimi Team}.
\newblock {Kimi K2}: Open agentic intelligence.
\newblock \emph{arXiv preprint arXiv:2507.20534}, 2025.

\bibitem[Kovarik(1970)]{kovarik1970methods}
Z.~Kovarik.
\newblock Some iterative methods for improving orthonormality.
\newblock \emph{SIAM Journal on Numerical Analysis}, 7\penalty0 (3):\penalty0 386--389, 1970.
\newblock \doi{10.1137/0707031}.

\bibitem[Kumar and Haupt(2024)]{kumar2024early}
A.~Kumar and J.~Haupt.
\newblock Early directional convergence in deep homogeneous neural networks for small initializations.
\newblock \emph{arXiv preprint arXiv:2403.08121}, 2024.

\bibitem[Kumar and Haupt(2025)]{kumar2025beyond}
A.~Kumar and J.~Haupt.
\newblock Towards understanding gradient flow dynamics of homogeneous neural networks beyond the origin.
\newblock \emph{Journal of Machine Learning Research}, 26\penalty0 (269):\penalty0 1--83, 2025.

\bibitem[Li and Tsuchiya(2026)]{li2026smoothing}
M.~Li and T.~Tsuchiya.
\newblock {Muon} with finite {Newton--Schulz}: The smoothing benefit in nonsmooth nonconvex optimization.
\newblock \emph{arXiv preprint arXiv:2608.26288}, 2026.

\bibitem[Li and Tan(2026)]{li2026lastiterate}
P.~Li and V.~Y.~F. Tan.
\newblock Last-iterate stationarity of {Muon} with finitely many {Newton--Schulz} iterations.
\newblock \emph{arXiv preprint arXiv:2610.34158}, 2026.

\bibitem[Li et~al.(2026)Li, Di, Cao, and Gu]{li2026facts}
X.~Li, Q.~Di, Y.~Cao, and Q.~Gu.
\newblock {Muon} learns facts better: Understanding the role of spectral orthogonalization.
\newblock \emph{arXiv preprint arXiv:2610.02798}, 2026.

\bibitem[Li et~al.(2021)Li, Luo, and Lyu]{li2021greedy}
Z.~Li, Y.~Luo, and K.~Lyu.
\newblock Towards resolving the implicit bias of gradient descent for matrix factorization: Greedy low-rank learning.
\newblock In \emph{International Conference on Learning Representations}, 2021.

\bibitem[Liu et~al.(2025)Liu, Su, Yao, Jiang, Lai, Du, Qin, Xu, Lu, Yan, Chen, Zheng, Liu, Liu, Yin, He, Zhu, Wang, Wang, Dong, Zhang, Kang, Zhang, Xu, Zhang, Wu, Zhou, and Yang]{liu2025scalable}
J.~Liu, J.~Su, X.~Yao, Z.~Jiang, G.~Lai, Y.~Du, Y.~Qin, W.~Xu, E.~Lu, J.~Yan, Y.~Chen, H.~Zheng, Y.~Liu, S.~Liu, B.~Yin, W.~He, H.~Zhu, Y.~Wang, J.~Wang, M.~Dong, Z.~Zhang, Y.~Kang, H.~Zhang, X.~Xu, Y.~Zhang, Y.~Wu, X.~Zhou, and Z.~Yang.
\newblock Muon is scalable for {LLM} training.
\newblock \emph{arXiv preprint arXiv:2502.16982}, 2025.

\bibitem[Ma et~al.(2026)Ma, Huang, Chi, and Chen]{ma2026preconditioning}
J.~Ma, Y.~Huang, Y.~Chi, and Y.~Chen.
\newblock Preconditioning benefits of spectral orthogonalization in {Muon}.
\newblock \emph{arXiv preprint arXiv:2601.13474}, 2026.

\bibitem[Martin et~al.(2024)Martin, Bach, and Biroli]{martin2024impact}
S.~Martin, F.~Bach, and G.~Biroli.
\newblock On the impact of overparameterization on the training of a shallow neural network in high dimensions.
\newblock In \emph{Proceedings of the 27th International Conference on Artificial Intelligence and Statistics}, volume 238 of \emph{Proceedings of Machine Learning Research}, pages 3655--3663, 2024.
\newblock URL \url{https://proceedings.mlr.press/v238/martin24a.html}.

\bibitem[Martin et~al.(2026)Martin, Biroli, and Bach]{martin2026quadratic}
S.~Martin, G.~Biroli, and F.~Bach.
\newblock High-dimensional analysis of gradient flow for extensive-width quadratic neural networks.
\newblock \emph{Journal of Machine Learning Research}, 27\penalty0 (136):\penalty0 1--182, 2026.

\bibitem[Mityagin(2020)]{mityagin2020zero}
B.~S. Mityagin.
\newblock The zero set of a real analytic function.
\newblock \emph{Mathematical Notes}, 107\penalty0 (3):\penalty0 529--530, 2020.
\newblock Preprint arXiv:1512.07276.

\bibitem[Peng et~al.(2026)Peng, Wang, and Yu]{peng2026practical}
H.~Peng, H.~Wang, and Y.~Yu.
\newblock Convergence of practical {Muon} with finite {Newton--Schulz} iterations and {Nesterov} momentum.
\newblock \emph{arXiv preprint arXiv:2609.39595}, 2026.

\bibitem[Rhee et~al.(2026)Rhee, Simon, and Karkada]{rhee2026balanced}
M.~Rhee, J.~Simon, and D.~Karkada.
\newblock {Muon} learns balanced solutions in matrix factorization without slow saddle-to-saddle dynamics.
\newblock \emph{arXiv preprint arXiv:2606.30509}, 2026.

\bibitem[Saxe et~al.(2014)Saxe, McClelland, and Ganguli]{saxe2014exact}
A.~M. Saxe, J.~L. McClelland, and S.~Ganguli.
\newblock Exact solutions to the nonlinear dynamics of learning in deep linear neural networks.
\newblock In \emph{International Conference on Learning Representations}, 2014.
\newblock URL \url{https://arxiv.org/abs/1312.6120}.

\bibitem[Teschl(2012)]{teschl2012ordinary}
G.~Teschl.
\newblock \emph{Ordinary Differential Equations and Dynamical Systems}, volume 140 of \emph{Graduate Studies in Mathematics}.
\newblock American Mathematical Society, 2012.
\newblock URL \url{https://www.mat.univie.ac.at/~gerald/ftp/book-ode/ode.pdf}.

\end{thebibliography}
